%% file: main.tex
\documentclass[10pt,twocolumn,letterpaper]{article}

\usepackage[pagenumbers]{cvpr} 

\input{preamble}
\definecolor{cvprblue}{rgb}{0.21,0.49,0.74}
\usepackage[pagebackref,breaklinks,colorlinks,citecolor=cvprblue]{hyperref}

\def\paperID{455} 
\def\confName{CVPR}
\def\confYear{2024}

\title{Visual Concept Connectome (VCC): Open World Concept Discovery and their Interlayer Connections in Deep Models} 

\author{Matthew Kowal$^{1,3}$ \hspace{0.4cm}
Richard P. Wildes$^{1,2}$ \hspace{0.4cm}
Konstantinos G. Derpanis$^{1,2,3}$ \\
{\normalsize $^{1}$York University, $^{2}$Samsung AI Centre Toronto, $^{3}$Vector Institute} \\ \vspace{0.1cm}
Project Page: \href{https://yorkucvil.github.io/VCC}{yorkucvil.github.io/VCC}
}

\begin{document}
\maketitle
\input{sec/0_abstract}    
\input{sec/1_intro}
\input{sec/2_related}
\input{sec/3_method}
\input{sec/4_experiments_final}
\input{sec/5_application}

\input{sec/6_conclusion}
\input{sec/X_suppl} 

\clearpage
\newpage 

{
    \small
    \bibliographystyle{ieeenat_fullname}
    \bibliography{main}
}


\end{document}

%% file: preamble.tex
\usepackage{times}
\usepackage{epsfig}
\usepackage{amsmath}
\usepackage{amssymb}
\usepackage{booktabs}
\usepackage{array}
\usepackage{algpseudocode}
\usepackage{algorithm}
\usepackage{url}
\usepackage{color,soul}
\usepackage{multirow,bigdelim}
\usepackage{colortbl}
\usepackage{tabu}
\usepackage{tabularx}
\usepackage{pifont}
\usepackage{bbm}
\usepackage{mathtools}
\usepackage{float}
\usepackage{caption}
\usepackage{tikz}
\usepackage{tikz-network}
\usepackage{pgfplots}
\usepackage{enumitem}
\newcommand{\vrulesep}{\unskip\ \vrule\ }

\usetikzlibrary{arrows.meta,bending,calc}

\input{math_commands.tex}

\usepgfplotslibrary{colormaps}

\pgfplotsset{
colormap={mycolormap2}{
rgb255=(255,255,255)
rgb255=(0,0,0) 
}}

\pgfplotsset{
colormap={mycolormap}{
rgb255=(0,0,0) 
rgb255=(40,40,40) 
rgb255=(150.5,150.5,150.5)
rgb255=(170,170,170)
rgb255=(223.125,223.125,223.125)
rgb255=(240,240,240)
rgb255=(250,250,250)
}}

%% file: math_commands.tex
\usepackage{amsmath,amsfonts,bm}

\def\app#1#2{%
  \mathrel{%
    \setbox0=\hbox{$#1\sim$}%
    \setbox2=\hbox{%
      \rlap{\hbox{$#1\propto$}}%
      \lower1.1\ht0\box0%
    }%
    \raise0.25\ht2\box2%
  }%
}

\def\eqref#1{equation~\ref{#1}}

\def\1{\bm{1}}

\DeclareMathAlphabet{\mathsfit}{\encodingdefault}{\sfdefault}{m}{sl}
\SetMathAlphabet{\mathsfit}{bold}{\encodingdefault}{\sfdefault}{bx}{n}



%% file: sec/0_abstract.tex
\begin{abstract}
Understanding what deep network models capture in their learned representations is a fundamental challenge in computer vision. We present a new methodology to understanding such vision models, the Visual Concept Connectome (VCC), which discovers human interpretable concepts and their interlayer connections in a fully unsupervised manner. Our approach simultaneously reveals fine-grained concepts at a layer, connection weightings across all layers and is amendable to global analysis of network structure (e.g.\ branching pattern of hierarchical concept assemblies). Previous work yielded ways to extract interpretable concepts from single layers and examine their impact on classification, but did not afford multilayer concept analysis across an entire network architecture.
Quantitative and qualitative empirical results show the effectiveness of VCCs in the domain of image classification. 
Also, we leverage VCCs for the application of failure mode debugging to reveal where mistakes arise in deep networks.
\end{abstract}

%% file: sec/1_intro.tex
\vspace{-0.5cm}
\section{Introduction}
\vspace{-0.2cm}
This paper focuses on interpreting the intermediate representations of deep networks for computer vision. The goal is understanding how various encoded concepts impact a model's prediction as well as concepts in other layers. 
We define concepts as abstractions that generalize from particular instances, including those defined locally (\eg color and orientation), regionally (\eg texture and shading) and from higher-level considerations (\eg object parts, wholes and groupings).
Human-interpretable concepts are of particular interest in increasing human understanding of models; however,
extracting these concepts encoded in deep networks remains an open challenge in computer vision due to the complexity and opaque nature of these models. 

\begin{figure*}
   \centering   
\includegraphics[width=0.92\textwidth]{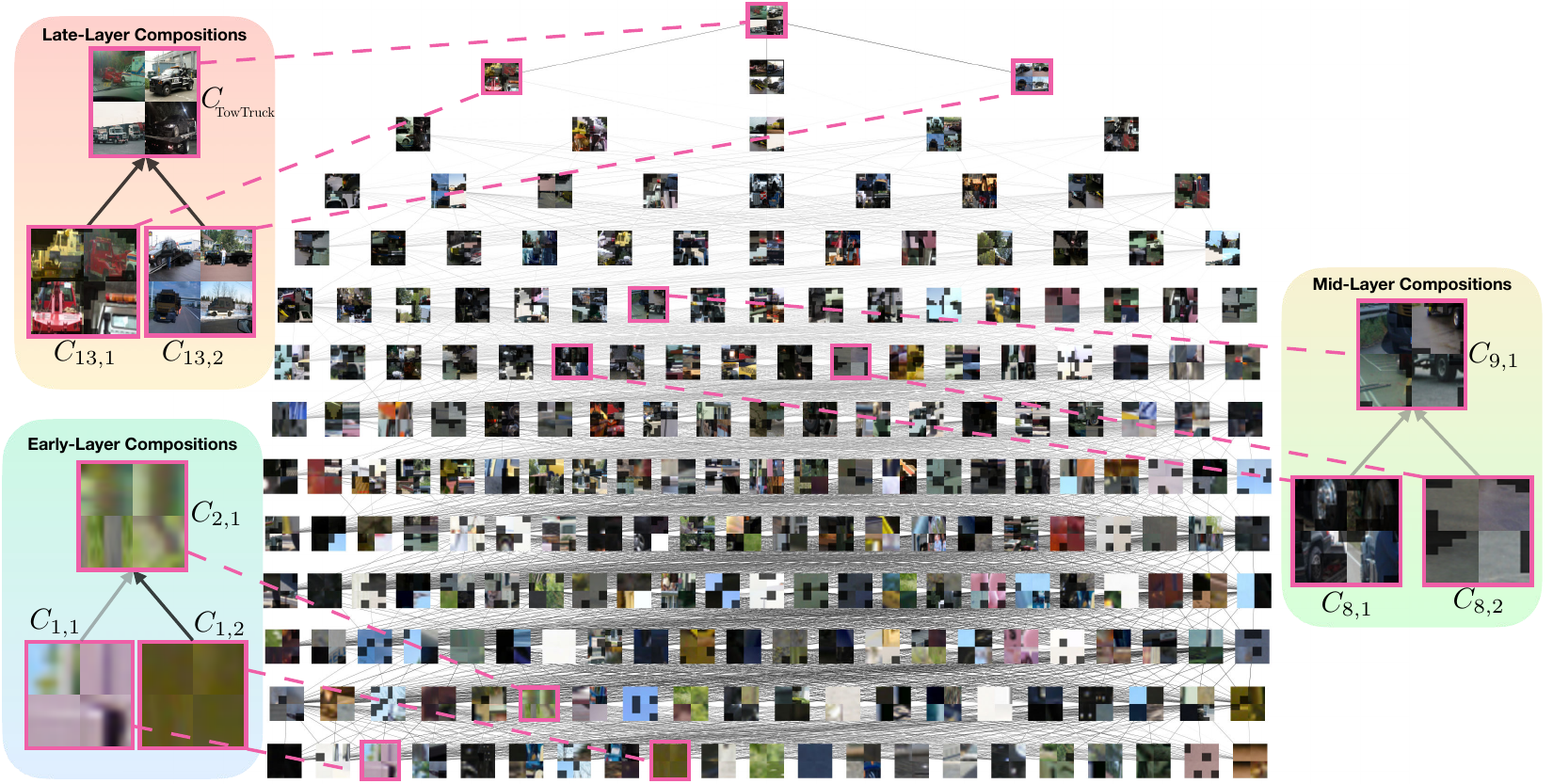}\vspace{-0.3cm}
    \captionof{figure}{A Visual Concept Connectome (VCC). At each layer the visual concepts learned by a deep model for a given class are revealed as are the learned interlayer concept connections. For each concept, up to four exemplars are shown as unmasked regions in a  $2\times 2$ image. Interlayer concept connections are shown as lines with darker lines indicating larger contributions. Shown is a VCC for every convolutional layer of a VGG16 model~\cite{szegedy2015going} trained on ImageNet~\cite{deng2009imagenet} targeting recognition of class ``Tow Truck". A closer visualization of VCC subgraphs reveals interesting compositions occurring at different levels of abstraction corresponding at different depths of the model. At early layers (bottom left), we observe oriented patterns ($C_{1,1}$) and brown color ($C_{1,2}$) composing the concept of green and brown orientation ($C_{2,1}$). Middle layers (right) show the concept of `wheel on the road' ($C_{9,1}$) being composed of wheels ($C_{8,1}$) and regions of asphalt ($C_{8,2}$). The final layer concepts (top left) show that both foreground objects, \eg tow trucks ($C_{13,1}$), and background regions, \eg road, trees, humans, or car being towed ($C_{13,2}$), concepts highly influence the final category ($C_{TowTruck}$). 
    }\label{fig:intro_fig2}\vspace{-0.4cm}
\end{figure*}

Understanding what concepts are learned by deep models and
how they are encoded 
is important for both science and applications. 
For science, understanding what information drives a model's encoding of different concepts may indicate directions to advance performance. 
For applications, several negative consequences of deploying opaque vision models have been documented, \eg~\cite{buolamwini2018gender,hansson2021self}. 

Previous work has focused on interpreting models via feature attribution, which measures the contribution of individual inputs to a model's output~\cite{zhou2016learning,selvaraju2017grad,chefer2021transformer};
so, explanations are of single pixels and may be difficult to interpret. Other work generates images that maximize activation of a model's features~\cite{zeiler2014visualizing, olah2017feature, feichtenhofer2020deep}. Like feature attribution, these approaches are qualitative and place most of the burden on the user to determine the concepts revealed. 
Concept-based interpretability, which identifies human-interpretable abstractions in a model's latent space~\cite{kim2018interpretability,ghorbani2019towards,fel2023craft,kowal2024understanding}, yield quantitative contributions of a concept to the model's output and explanations on the class-level (vs.~pixel-level). These approaches are easier to understand and validate; however, they have not been used to explore \textit{interlayer} relationships.

As it stands, no approaches can quantify the interlayer affect of a given distributed concept at one layer, $l$, to another concept at a different layer, $l'$ (rather than the model \textit{output}). Even though it is well established that deep networks learn to build concepts hierarchically as information flows through the network~\cite{zeiler2014visualizing,olah2018the}, understanding the hierarchical representations has been under-researched.
Indeed, little is known about the characteristics of this concept hierarchy for today's models. Questions abound: \textit{How many concepts exist in a network?} \textit{What are the connections and weights between concepts?} \textit{Does the model architecture impact the hierarchical structure of concept abstractions?}  

In response to these questions, we take inspiration from the biological notion of a \textit{connectome}~\cite{seung2012connectome}, defined as ``a comprehensive structural description of the network of elements and connections forming a brain.'' 
Analogously, we present the \textit{Visual Concept Connectome} (VCC), a comprehensive structural description of a deep pretrained network model in terms of human-interpretable concepts and their relationships that form the internal representation maintained by the model; Fig.~\ref{fig:intro_fig2} shows an example. Notably, VCC generation works in the \textit{open-world} setting, \ie the concepts and interlayer connections are discovered without the need for any predefined concept dictionary. 

%% file: sec/2_related.tex
\section{Related research}\label{sec:related}
\vspace{-4pt}
While a number of intrinsically interpretable networks have appeared (\eg~\cite{chen2020concept,bohle2022b,zhang2018interpretable,zhang2019interpreting,koh2020concept,sarkar2022framework}), we focus on work that, like ours, endeavours to interpret black-box models.

\textbf{Concept-based interpretability.} 
Closed-world concept interpretability considers cases where a labelled dataset defines the concepts of interest for post-hoc analysis; approaches include Network Dissection~\cite{bau2017network} and Testing with Concept Activation Vectors (TCAV)~\cite{kim2018interpretability}.
However, a desirable property for concept-based interpretability is not to be restricted to a set of predefined classes (\ie closed-world), but to support discovery of new and previously unlabelled concepts (\ie open-world). Approaches to open-world concept interpretabilty include Automatic Concept Explanations (ACE)~\cite{ghorbani2019towards}, SegDiscover~\cite{huang2022segdiscover}, ConceptSHAP~\cite{yeh2020completeness}, Invertible Concept Explanations~\cite{zhang2021invertible} and CRAFT~\cite{fel2023craft}.
These approaches are limited to single layer analysis and do not explain interlayer relationships. 



Activation maximization~\cite{zeiler2014visualizing, mahendran2015understanding, olah2017feature, feichtenhofer2020deep} visualizes the input that most activates a model component (\eg filter, layer or logit) or 
how same layer units combine information~\cite{olah2020zoom}.
These approaches do not capture interlayer relations and place a heavy burden on the user to interpret what concepts appear in their output where there may be no clear resemblance to natural concepts. 
Class Activation Maps (CAMs) and extensions visualize an input image's local region that contributes to a model's output~\cite{zhou2016learning,selvaraju2017grad,chattopadhay2018grad}. Layer-wise Relevance Propagation~\cite{binder2016layer} (LRP) generates pixel contribution heatmaps 
by assuming conservation of information is propagated through 
each neuron. These approaches interpret single images (not classes), are purely qualitative, and are sensitive to small input perturbations~\cite{kindermans2019reliability,ghorbani2019interpretation}, yet insensitive to model changes~\cite{adebayo2018sanity}.
Less related are directions interpreting generative models, \eg~\cite{liu2017analyzing,bau2020rewriting}.


\textbf{Interpretability via decomposition.}
Most closely related to our work are those that decompose a model's prediction into interpretable components. CRAFT~\cite{fel2023craft} extracts concepts from the last layer of a convolutional neural network (CNN), but also searches earlier layers for sub-concepts for the maximally activating images of a given concept, but do not provide any way to quantify the contribution of sub-concepts to concepts.
Another approach combined CAMs of multiple concepts to explain the final prediction~\cite{zhou2018interpretable}; however, it is limited to labelled data (so cannot discover concepts outside training) and only explains a final output, not intermediate relations. 
Other work interprets CNNs by distilling them into a  graph~\cite{zhang2018interpreting}; however, it applies only to individual filters conceived as capturing object parts (\ie not concepts other than objects and their parts, \eg textures, colors or grouping), does not consider the fact that representations are encoded via superposition~\cite{elhage2022superposition,olah2020zoom,bricken2022monosemanticity} (\ie more than one concept can be captured per-channel), does not yield meaningful edge weights and is only for CNNs.
Yet other work extends Grad-CAM or LRP to perform intra-layer visualizations~\cite{achtibat2022towards,cheng2023deeply}. While these approaches capture concepts other than object parts, they still ignore the core problem of superposition (and distributed representations) and do not yield interpretation for an entire class, just single images. 

\textbf{Contributions.} In the light of previous work, our contributions are  fourfold. (i) We introduce and formalize the notion of a Visual Concept Connectome (VCC) for deep network models. The VCC reveals concepts represented at any given layer of a network as well as their interlayer connectivity. (ii) We present a method for extracting a VCC from any pretrained deep network in an unsupervised, open world setting, with a focus on classification. (iii) We validate our approach with quantitative and qualitative experiments wherein we examine various standard models to yield insights into model architectures and training tasks. 
(iv) We apply VCCs to explaining failure modes in deep models.

%% file: sec/3_method.tex
\section{Visual Concept Connectomes (VCCs)}
A VCC is a directed acyclic graph, $\mathcal{\textbf{G}}(\textbf{Q},\textbf{E})$, created by distilling a pretrained deep network.
The graph is topologically sorted based on the layer order found in the original network and has $n+1$ layers, consisting of $n$ network layers to be analyzed (\ie any subset of layers that a user selects, including all layers at the extreme) plus the final prediction (\eg the object category for object recognition). The graph's nodes, $\textbf{Q}$, are vectors (cluster centroids) representing interpretable concepts and its edges, $\textbf{E}$, are scalars representing the contribution of one concept to the existence of another. 

To construct a VCC, three inputs are required: (i) a set of $I$ images representative of a given task (\eg exemplar images of an object category for object recognition), 
$\bm{\mathcal{I}} = \{ \mathcal{I}^1, \dotsc, \mathcal{I}^I\}, \mathcal{I}^i \in \mathbb{R}^{h \times w \times c}$, with $h, w$ and $c$ the image height, width and channel dimension (\eg three for RGB), respectively, (ii) an $N$ layer network, $F(\cdot) = \{f_1(\cdot),\dotsc,f_N(\cdot)\}$, with $f_j$ denoting feature extraction at layer $j$, and (iii) a set of $n$ selected layers (a subset of $F$, possibly improper), which are to be studied.

\begin{figure*}
    \centering   
\includegraphics[width=0.98\textwidth]{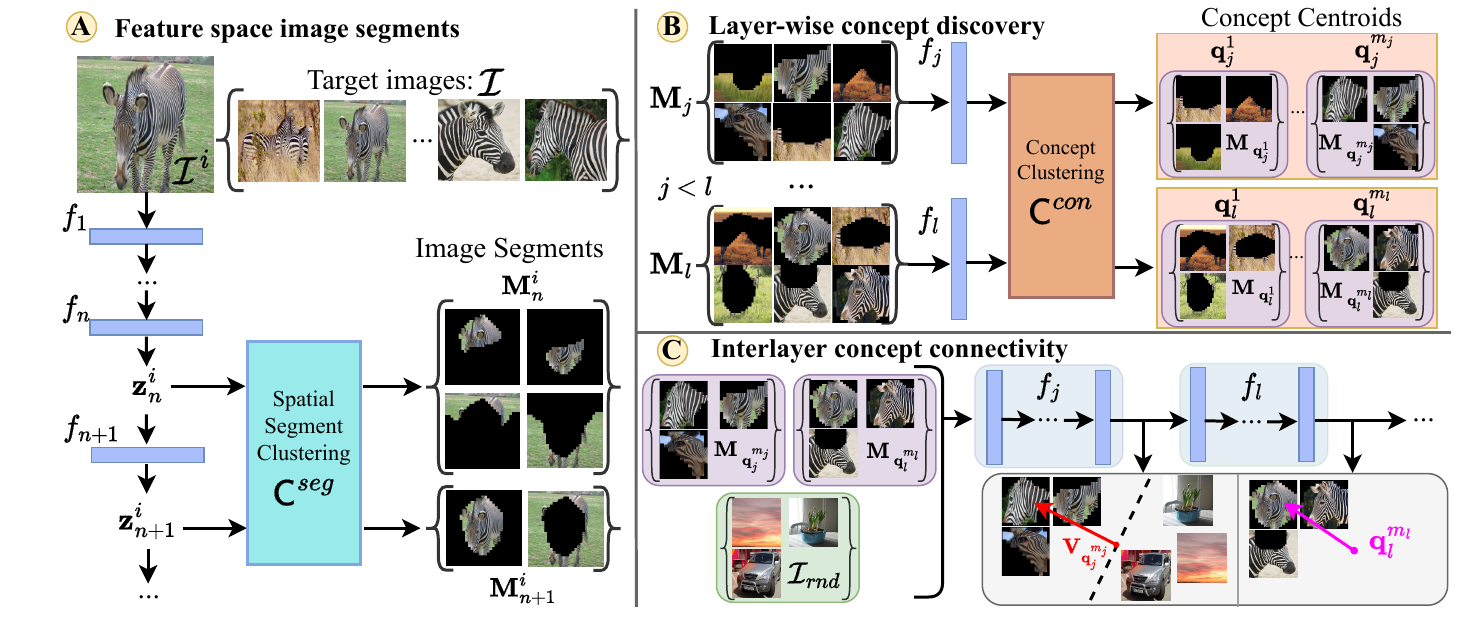}\vspace{-0.45cm}
    \caption{The three steps in building a Visual Concept Connectome (VCC). (\textbf{A}) For a given image, $\mathcal{I}^i \in \mathcal{I}$, model, $F$, and layer, $n$, we produce a set of image segments, $\mathbf{M}^i_n \in \mathbf{M}_n$, based on a recursive spatial clustering, $\mathsf{C}^{seg}$ (\ref{eq:cluster}), of the features $\textbf{z}_n^i$ conditioned on the clusters from the layer above, $n+1$. We then use (\ref{eq:rgb_mask}) to generate a set of masked RGB image segments for each layer, $\mathbf{M}_j$. (\textbf{B}) For a given layer, $j$, we pass the image segments from all images, $\mathbf{M}_j$, through $f_j$ and cluster, $\mathsf{C}^{con}$ (\ref{eq:concept_cluster}), these features across the dataset to produce $m_j$ concept centroids, $\{\textbf{q}^1_j,\dotsc,\textbf{q}^{m_j}_j\}$. (\textbf{C}) 
    To measure the contribution of an earlier layer concept, $\textbf{q}^{m_j}_j$, to a later layer concept, $\textbf{q}^{m_l}_l$, we employ
    our Interlayer Testing with CAV (ITCAV) approach (Sec.~\ref{sec:itcav}), which uses the Concept Activation Vector (CAV)~\cite{kim2018interpretability} of the earlier concept, $\textbf{V}_{\textbf{q}^{m_j}_j}$ (that points away from random examples, $\mathcal{I}_{rnd}$, but toward concept exemplars, $\mathbf{M}_{\textbf{q}^{m_j}_j}$),  and the deeper layer concept, $\textbf{q}^{m_l}_l$. 
    }
    \label{fig:main_method} \vspace{-18pt}
\end{figure*}

A VCC is constructed in three main steps. (i) Image segments are extracted in the model's feature space via divisive clustering to produce semantically meaningful image regions for each selected layer (Sec.~\ref{sec:divisive_cluster}). (ii) Layer-wise concepts, \ie the nodes of the graph, are discovered in an open world fashion (\ie no labelled data is required) via a second round of clustering over the dataset of image regions, independently for each layer (Sec.~\ref{sec:concept_discovery_clustering}). (iii) Edges are calculated that indicate the contribution of concepts from earlier to deeper layers via an approach we introduce, Interlayer Testing with Concept Activation Vectors (ITCAVs) (Sec.~\ref{sec:itcav}). We use object recognition for the explanation of the approach; nevertheless, it could be extended to other tasks in a straightforward way (\eg semantic segmentation). An overview of these steps is provided in Fig.~\ref{fig:main_method}.

\subsection{Feature space image segments}
\label{sec:divisive_cluster}

\input{sec/3_method_New31}

\subsection{Layer-wise concept discovery}\label{sec:concept_discovery_clustering}
Given the dataset of masked image segments for each layer, $\mathbf{M} = \{\mathbf{M}_1,\dotsc,\mathbf{M}_n\}$, we follow~\cite{ghorbani2019towards} and bilinearly interpolate the segments to the input image size and then pass the segments through the model to get pre-segmented activations $\textbf{Z}_{\mathbf{M}_j} = f_j(\mathbf{M}_j)$ from the layer where the segments were found in the divisive clustering step (Sec.~\ref{sec:divisive_cluster}). 
We pass the image segments (as opposed to using the image activations) through the model to remove any contextual information before we discover concepts (the segments are set to zero everywhere outside of the segment boundary). Otherwise, $\mathbf{M}_j$ could mix information from outside the segment. 

The layer-wise concept discovery step is presented visually in Fig.~\ref{fig:main_method} (\textbf{B}). With the pre-segmented activations, $\textbf{Z}_{\mathbf{M}_j}$,  calculated for all layers, $j$, we cluster over the dataset of segment features for each individual layer 
to produce cluster centroids (\ie concepts) according to
\begin{equation}\label{eq:concept_cluster}
 \textbf{Q}_j = \{\textbf{q}^1_j,\dotsc,\textbf{q}^{m_j}_j\} =  \mathsf{C}^{con}_{m_j}(\mathsf{GAP}(\textbf{Z}_{\mathbf{M}_j})), 
\end{equation}
where $m_j$ is the number of concepts discovered at layer $j$, $\mathsf{GAP}$ is spatial global average pooling and $\mathsf{C}^{con}_{m_j}$ is a clustering algorithm that returns $m_j$ cluster centroids. Following previous work~\cite{ghorbani2019towards}, we instantiate $\mathsf{C}^{con}_{m_j}$ in terms of standard k-means, using the $l_2$ norm, and then use a large number of clusters with subsequent pruning to remove noisy clusters; see the appendix for details. The output of this stage results in discovered concepts (\ie cluster centroids and associated segments) for all layers as $\textbf{Q} = \{\textbf{Q}_1,\dotsc,\textbf{Q}_n\}$ for the centroids and $\mathbf{M}_{\textbf{q}^{m_j}_j}$ their associated RGB segments.


\begin{figure*}[t]
    \centering
\begin{subfigure}{0.24\textwidth}
    \begin{tikzpicture}
	\begin{axis}[
            title style={at={(axis description cs:0.5,0.95)},anchor=north,font=\footnotesize}, 
		xlabel=Receptive Field (pixels),
		ylabel=Rel. Concept Size (\%),
            height=3.87cm,
            width=4.7cm,
            xmin=-10, xmax=450,
            ymin=-3, ymax=80,
            xtick pos=left,
            ytick pos=left,
            x tick label style={{yshift=0.08cm},font=\tiny},
            y tick label style={{xshift=0.09cm},font=\tiny},
            x label style={at={(axis description cs:0.5,0.18)},anchor=north,font=\scriptsize}, 
            y label style={at={(axis description cs:0.35,.5)},anchor=south,font=\scriptsize},
            legend style={
             nodes={scale=0.5, transform shape},
             cells={anchor=west},
             legend style={at={(0.5,1.0)},anchor=north}, font =\footnotesize},
             legend image post style={scale=0.45},
             legend columns=3]

	\addplot[color=magenta,mark=triangle] coordinates {
		(43,2.37)
            (99,2.04)
            (211,1.92)
            (435,2.13)
	}; 
        \addlegendentry{ACE-R50}
        
        \addplot[color=cyan,mark=square] coordinates {
		(58,1.3)
            (85,1.7)
            (109,2.4)
            (118,2.4)
	}; 
        \addlegendentry{ACE-ViT}

        \addplot[color=olive,mark=o] coordinates {
		(10,2.84)
            (32,2.44)
            (80,2.18)
            (176,2.19)
	}; 
        \addlegendentry{ACE-VGG16}


        
        \addplot[color=magenta,mark=triangle*] coordinates {
		(43,4.16)
            (99,11.89)
            (211,23.9)
            (435,47.9)
	}; 
        \addlegendentry{Ours-R50}

        \addplot[color=cyan,mark=square*] coordinates {
		(58,10.0)
            (85,24.4)
            (109,35.8)
            (118,54.1)
	}; 
        \addlegendentry{Ours-ViT}
        
        \addplot[color=olive,mark=*] coordinates {
		(10,3.46)
            (32,6.77)
            (80,15.45)
            (176,46.8)
	}; 
        \addlegendentry{Ours-VGG16}


        
	\end{axis}
\end{tikzpicture}
\caption{} \label{fig:1a}
\end{subfigure}
\hspace*{\fill}
\begin{subfigure}{0.24\textwidth}
    \begin{tikzpicture}
	\begin{axis}[
            title style={at={(axis description cs:0.5,0.95)},anchor=north,font=\footnotesize}, 
		xlabel=$\epsilon$,
		ylabel=Acc. (\%),
            height=4cm,
            width=4.8cm,
            xmin=-0.01, xmax=1.1,
            ymin=-0.02, ymax=1.1,
            x tick label style={{yshift=0.08cm},font=\tiny},
            y tick label style={{xshift=0.09cm},font=\tiny},
            x label style={at={(axis description cs:0.5,0.18)},anchor=north,font=\footnotesize}, 
            y label style={at={(axis description cs:0.3,.5)},anchor=south,font=\footnotesize},
            xtick pos=left,
            ytick pos=left,
            legend style={mark size=5pt},
            legend style={
             nodes={scale=0.36, transform shape},
             cells={anchor=west},
             legend style={at={(1,0.826)},anchor=east}, font=\large},
             legend image post style={scale=0.5},
             legend columns=2]

    	\addplot[color=magenta,densely dotted,mark=star,mark options={scale=0.45,solid}] coordinates {
		(0.000,	0.782)
            (0.007,	0.777)
            (0.020,	0.767)
            (0.033,	0.752)
            (0.047,	0.735)
            (0.060,	0.718)
            (0.073,	0.705)
            (0.087,	0.691)
            (0.100,	0.673)
            (0.113,	0.654)
            (0.127,	0.640)
            (0.140,	0.623)
            (0.153,	0.607)
            (0.167,	0.590)
            (0.180,	0.561)
            (0.193,	0.538)
            (0.267,	0.358)
            (0.333,	0.159)
            (0.667,	0.000)
            (1.000,	0.000)
	}; 
        \addlegendentry{R50-Ours}

        \addplot[color=magenta,mark=star,mark options={scale=0.45,solid}] coordinates {
		(0.000,	0.782)
            (0.007,	0.781)
            (0.020,	0.783)
            (0.033,	0.785)
            (0.047,	0.781)
            (0.060,	0.782)
            (0.073,	0.782)
            (0.087,	0.781)
            (0.100,	0.773)
            (0.113,	0.775)
            (0.127,	0.775)
            (0.140,	0.772)
            (0.153,	0.776)
            (0.167,	0.773)
            (0.180,	0.772)
            (0.193,	0.765)
            (0.267,	0.747)
            (0.333,	0.713)
            (0.667,	0.205)
            (1.000,	0.006)
	}; 
        \addlegendentry{R50-Rand}

	\addplot[color=blue,densely dotted,mark=triangle,mark options={scale=0.45,solid}] coordinates {
            (0.000,	0.673)
            (0.025,	0.652)
            (0.050,	0.631)
            (0.075,	0.601)
            (0.100,	0.563)
            (0.125,	0.530)
            (0.150,	0.485)
            (0.175,	0.431)
            (0.200,	0.380)
            (0.225,	0.324)
            (0.250,	0.284)
            (0.275,	0.232)
            (0.300,	0.183)
            (0.325,	0.146)
            (0.350,	0.111)
            (0.375,	0.081)
            (0.500,	0.015)
            (0.625,	0.001)
            (0.750,	0.000)
            (0.875,	0.000)
            (1.000,	0.000)
	}; 
        \addlegendentry{VGG16-Ours}
        
        \addplot[color=blue,mark=triangle,mark options={scale=0.45,solid}] coordinates {
            (0.000,	0.673)
            (0.025,	0.676)
            (0.050,	0.676)
            (0.075,	0.668)
            (0.100,	0.654)
            (0.125,	0.635)
            (0.150,	0.610)
            (0.175,	0.561)
            (0.200,	0.517)
            (0.225,	0.466)
            (0.250,	0.405)
            (0.275,	0.339)
            (0.300,	0.267)
            (0.325,	0.216)
            (0.350,	0.166)
            (0.375,	0.127)
            (0.500,	0.022)
            (0.625,	0.003)
            (0.750,	0.000)
            (0.875,	0.000)
            (1.000,	0.000)
	}; 
        \addlegendentry{VGG16-Rand}

        \addplot[color=green,densely dotted,mark=square,mark options={scale=0.45,solid}] coordinates {
            (0.000,	0.819)
            (0.075,	0.804)
            (0.150,	0.711)
            (0.225,	0.351)
            (0.300,	0.042)
            (0.375,	0.011)
            (0.450,	0.015)
            (0.525,	0.014)
            (0.600,	0.005)
            (0.675,	0.001)
            (0.750,	0.000)
            (0.825,	0.000)
            (1.000,	0.000)
	}; 
        \addlegendentry{MViT-Ours}
        	\addplot[color=green,mark=square,mark options={scale=0.45,solid}] coordinates {
            (0.000,	0.819)
            (0.075,	0.822)
            (0.150,	0.814)
            (0.225,	0.804)
            (0.300,	0.788)
            (0.375,	0.742)
            (0.450,	0.700)
            (0.525,	0.615)
            (0.600,	0.488)
            (0.675,	0.345)
            (0.750,	0.226)
            (0.825,	0.135)
            (1.000,	0.029)
	}; 
        \addlegendentry{MViT-Rand}

            \addplot[color=brown,densely dotted,mark=o,mark options={scale=0.45,solid}] coordinates {
            (0.000,	0.676)
            (0.050,	0.642)
            (0.100,	0.545)
            (0.150,	0.304)
            (0.200,	0.071)
            (0.250,	0.008)
            (0.300,	0.001)
            (0.400,	0.000)
            (0.500,	0.000)
            (0.600,	0.000)
            (0.700,	0.000)
            (0.800,	0.000)
            (0.900,	0.000)
            (1.000,	0.000)
	}; 
        \addlegendentry{ViT-Ours}
        	\addplot[color=brown,mark=o,mark options={scale=0.45,solid}] coordinates {
            (0.000,	0.676)
            (0.050,	0.677)
            (0.100,	0.669)
            (0.150,	0.659)
            (0.200,	0.635)
            (0.250,	0.611)
            (0.300,	0.585)
            (0.400,	0.501)
            (0.500,	0.380)
            (0.600,	0.239)
            (0.700,	0.141)
            (0.800,	0.064)
            (0.900,	0.030)
            (1.00,	0.014)
	}; 
        \addlegendentry{ViT-Rand}
 
	\end{axis}
\end{tikzpicture}
\caption{} \label{fig:1b}
\end{subfigure}
\begin{subfigure}{0.24\textwidth}
\includegraphics[width=1.0\textwidth]{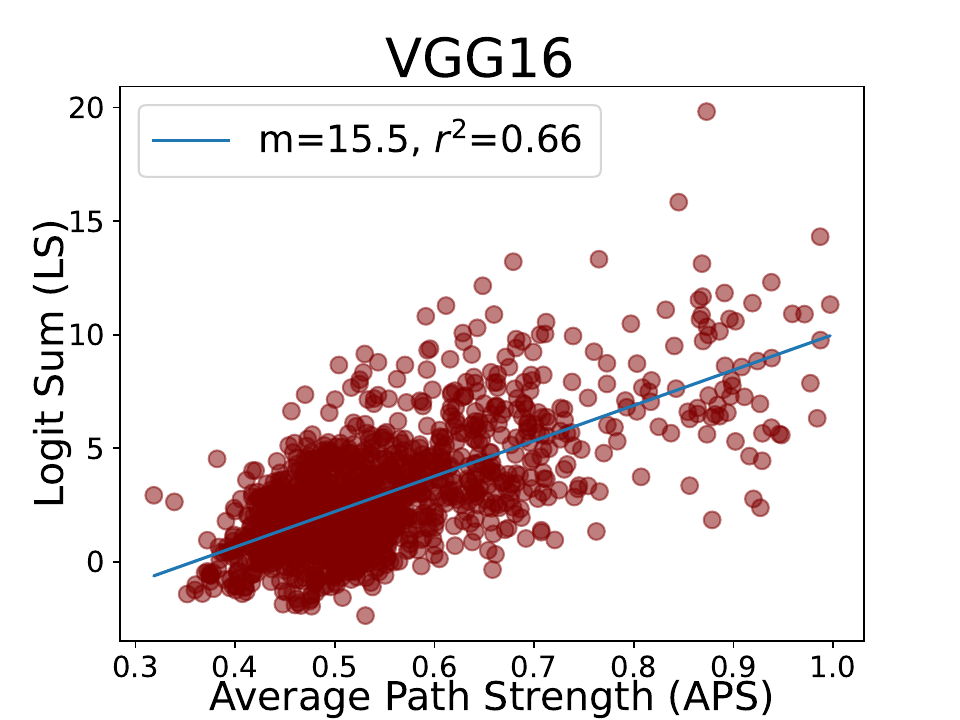}
\caption{} \label{fig:1c}
\end{subfigure}
\begin{subfigure}{0.24\textwidth}
\includegraphics[width=1.0\textwidth]{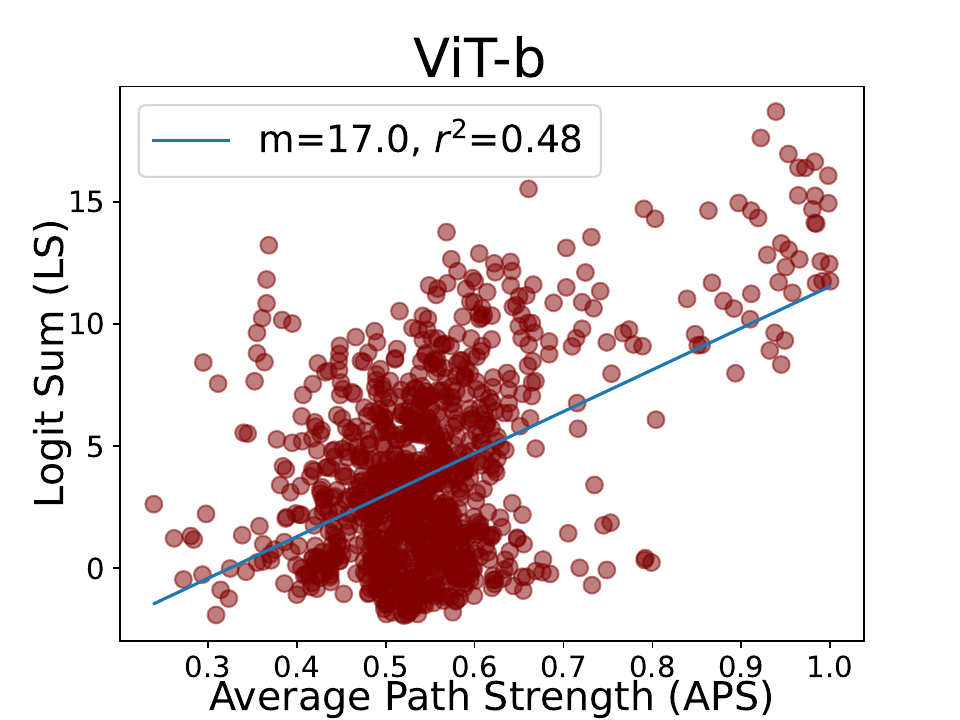}
\caption{} \label{fig:1d}
\end{subfigure}
\vspace{-0.4 cm}
\caption{Validation of the three VCC method components. 
(a) Validation of segment proposals. The relative (Rel.) concept segment size compared to entire image for a given layer is plotted against the receptive field (RF) width/height of the same layer. 
(b) Validation of discovered concepts. For 50 randomly selected ImageNet classes, we discover concepts in four layers of the model. During inference, one randomly selected concept at each layer is suppressed by a factor of $\epsilon$. (c, d) Validation of interlayer concept weights. The unnormalized logit sum (LS) scores, (\ref{eq:LS}), for the target class are plotted against the average path strength (APS) scores, (\ref{eq:APS}). A positive correlation implies that the ITCAV edge weights connecting a concept to the class are predictive of the model output having a higher probability for that class.
}
\label{fig:validation} \vspace{-0.4cm}
\end{figure*}
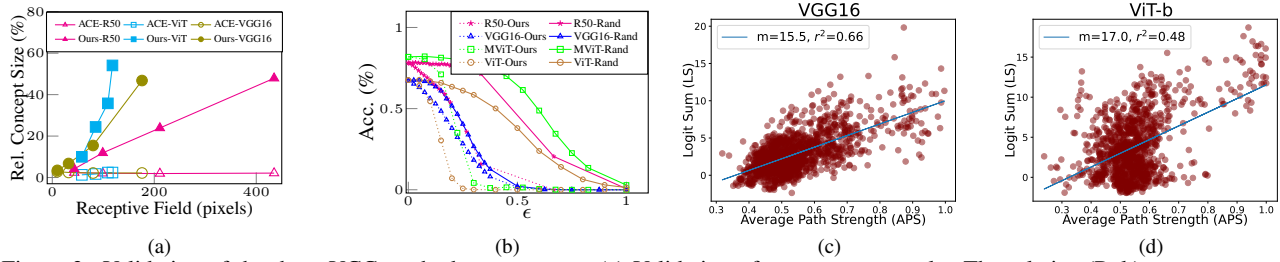

\subsection{Interlayer concept connectivity}
\label{sec:itcav}
Our approach to quantifying the contribution of discovered concepts between two layers, Interlayer Testing with Concept Activation Vectors (ITCAV), generalizes TCAV~\cite{kim2018interpretability}. Specifically, while TCAV calculates the contribution of a single mid-layer concept to the class logit using a dataset of labelled concept images, we generalize to operate between any two layers and without a dataset of labelled concept images. 
Without loss of generality, we consider a single pair of concepts at two layers, which need not be adjacent. 

The ITCAV method is presented visually in Fig.~\ref{fig:main_method} (\textbf{C}). We seek to measure the degree to which a concept at an earlier layer contributes to a deeper layer. To do so, we construct two vectors: (i) a vector that represents the concept at the earlier layer, $j$, and (ii) a vector that represents a positive contribution (\ie gradient) of a concept at a later layer, $l$. The closer the alignment of these two vectors, the larger the contribution of the concept at layer $j$ to the concept at layer $l$.
Let $\textbf{q}^{m_j}_j$ and $\mathbf{M}_{\textbf{q}^{m_j}_j}$ be the cluster centroid and associated RGB image segments in layer $j$, and let $\textbf{q}^{m_l}_l$ and $\mathbf{M}_{\textbf{q}^{m_l}_l}$ be the cluster centroid and associated RGB image segments in a deeper layer $l$, \ie $l > j$. 

We proceed by constructing the vector that represents the concept in the feature space of the earlier layer, $j$. To do so, we employ the Concept Activation Vector (CAV)~\cite{kim2018interpretability} for the earlier concept, $\textbf{q}^{m_j}_j$, by training a linear classifier, $h(\cdot)$, on the features of layer $j$ between positive concept inputs, $f_{j}(\mathbf{M}_{\textbf{q}^{m_j}_j})$, and random images, $f_{j}(\bm{\mathcal{I}}_{rnd})$. 
The orthogonal vector to the hyperplane defined by $h(\cdot)$, denoted as $ \textbf{V}_{\textbf{q}^{m_j}_j}$, is the CAV for concept $\textbf{q}^{m_j}_j$ and points in the direction of concept $\textbf{q}^{m_j}_j$ in the feature space of layer $j$. 

Next, we construct the vector in the feature space of layer $j$ that points in the direction of positive contribution of 
concept $\textbf{q}^{m_l}_l$. To do so, we calculate $\nabla f_{l}(f_{j}(\cdot))$, \ie the gradient of the deeper concept, $\textbf{q}^{m_l}_l$, at layer $l$, with respect to the earlier layer, $j$. Our approach to measuring the sensitivity between the two concepts is then
\vspace{-4pt}
\begin{equation}\label{eq:itav_sensitivity}
    S_{\textbf{q}^{m_j}_j, \textbf{q}^{m_l}_l}(x) = \nabla\Bigl(||f_{l}(f_{j}(x)) - \textbf{q}^{m_l}_l||_2\Bigr) \cdot \textbf{V}_{\textbf{q}^{m_j}_j}
    ,
\end{equation}
where $x \in \mathbf{M}_{\textbf{q}^{m_l}_l}$. 
The sensitivity score should be positive because
when the feature at layer $j$ is perturbed in the direction of concept $\textbf{q}^{m_j}_j$, \ie in the direction $\textbf{V}_{\textbf{q}^{m_j}_j}$, then the feature vector in deeper layer, $l$, will be pushed closer to concept $\textbf{q}^{m_l}_l$. Following TCAV~\cite{kim2018interpretability}, the final ITCAV edge weight, $e_{\textbf{q}^{m_j}_j, \textbf{q}^{m_l}_l} \in \textbf{E}$, 
is the ratio of positive sensitivities, 
\vspace{-4pt}
\begin{equation}\label{eq:itcav_edge_final}
   e_{\textbf{q}^{m_j}_j, \textbf{q}^{m_l}_l} = \big|x \in \mathbf{M}_{\textbf{q}^{m_l}_l} : S_{\textbf{q}^{m_j}_j, \textbf{q}^{m_l}_l}(x) > 0 \big| \big/ \big|\mathbf{M}_{\textbf{q}^{m_l}_l}\big|.
\end{equation}

Key to our sensitivity score, (\ref{eq:itav_sensitivity}), is the
$l_2$ distance between
the output, $f_{l}(f_{j}(x))$, and cluster centroid, $\textbf{q}^{m_l}_l$, \textit{before} taking the gradient, \textit{cf.}\ CAV~\cite{kim2018interpretability}. This choice serves two goals: (i) The subset of dimensions of individual image features, $f_{j}(x)$, that are not well aligned with the cluster centroid, $\textbf{q}^{m_l}_l$, will be penalized. This penalty will suppress the impact of noisy dimensions from individual samples on ITCAV.
We use the $l_2$ norm 
because the employed clustering approach used during concept discovery, k-means, uses $l_2$; see Sec.~\ref{sec:concept_discovery_clustering}.
(ii) Collapsing the tensor to a scalar (\ie a concept similarity score) 
makes the gradient calculation computationally feasible compared to the full Jacobian. 

%% file: sec/3_method_New31.tex
For each selected layer, we produce a set of image regions that plausibly belong to a concept encoded by the model at that layer. 
Previous work produces concept segments in RGB space, \ie superpixels or random crops~\cite{ghorbani2019towards,fel2023craft}; however, segmentation based on features different from those over which concepts are to be discovered, \ie the model's deep features, results in the support for the discovery process being divorced from the process that generated the support. Therefore, our segmentation uses the same deep features to be used subsequently for concept discovery. 
Our divisive clustering approach is presented visually in Fig.~\ref{fig:main_method} (\textbf{A}). 
We pass the entire set of $I$ input images, $\bm{\mathcal{I}}$, for a given class through the model, $F$, to get features,  $\textbf{z}^{i}_{n}  \in \mathbb{R}^{h_n \times w_n \times c_n}$, at each selected layer, $n$ according to
\begin{equation}\label{eq:feature_activations}
\mathbf{Z}_{n} = \{f_{n}(\mathcal{I}^{1}),...,f_{n}(\mathcal{I}^{I})\} = \{\textbf{z}^{1}_{n},...,\textbf{z}^{I}_{n}\}. 
\end{equation}

To extract image segments at layer $n$ for concept discovery (Sec.~\ref{sec:concept_discovery_clustering}), we cluster the image activations, $\textbf{Z}_{n}$, conditioned on the clusters (\ie masks) generated at the next higher layer, $n+1$. 
Let $\mathbf{p}=(x,y)$ index spatial coordinates and $\mathbf{B}^i_n(\mathbf{p};\gamma)$ be a binary mask for cluster $\gamma$ such that $\textbf{B}^i_n(\mathbf{p};\gamma)=1\iff \mathbf{z}^i_n(\mathbf{p}) \in\gamma$; otherwise, $\mathbf{B}^i_n(\mathbf{p};\gamma) = 0$.

We calculate a set of such masks
by applying a clustering algorithm, $\mathsf{C}_{\Gamma}^{seg}(\cdot)$, that returns binary support masks for $\Gamma$ clusters on its argument. Let $\Gamma_{n+1}$ be the number of segments at layer $n+1$ and $1\leq g \leq \Gamma_{n+1}$ index a particular segment, $g$. For each $g$ we calculate segments at layer $n$ as
\begin{equation}\label{eq:cluster}
    \bigl\{\mathbf{B}^{i}_{n}(\mathbf{p};\gamma)\bigr\}^{\Gamma}_{\gamma=1} = \mathsf{C}_{\Gamma}^{seg}\big(\textbf{z}^{i}_{n}(\mathbf{p}) \odot \widetilde{\mathbf{B}}^{i}_{n+1}(\mathbf{p};g)\big),
\end{equation}
where 
$\widetilde{\textbf{B}}^{i}_{n+1}$ is $\textbf{B}^{i}_{n+1}$ upsampled to the resolution of layer $n$ and $\odot$ indicates spatial element-wise masking applied to the individual image activations. 
The element-wise masking ensures clustering is done on the masked area of $\textbf{z}^i_{n}$. Thus, the support of the concept discovery at layer $n$ comes from the support of concepts at layer $n+1$. Notably, we let $\Gamma$ vary with $g$; so, different segments, $g$, at layer $n+1$ can yield a different number of segments, $\Gamma$, at layer $n$.

The top-down conditional clustering, (\ref{eq:cluster}), is repeated recursively for all image segments in all selected layers. To initiate the recursion, (\ref{eq:cluster}), at the top layer, $n_{top}$,
all the features are used by setting the number of clusters to be one, and having $\widetilde{\textbf{B}}^{i}_{n_{top}+1}(\mathbf{p};1) = 1$
for all images.

At each layer, $n$, we construct a set of RGB segmentations, $\mathbf{M}$, for use in concept discovery (Sec.~\ref{sec:concept_discovery_clustering}) by applying upsampled masks, $\textbf{B}^{i}_{n}(\mathbf{p};\gamma)$, to a given image, $\mathcal{I}^i$, via
 \begin{equation}\label{eq:rgb_mask}
     \mathbf{M}^{i}_{n}(\mathbf{p};\gamma) = \mathcal{I}^i(\mathbf{p}) \odot \textbf{B}^{i}_{n}(\mathbf{p};\gamma), 
 \end{equation}
 where $\mathbf{M}^{i}_{n}(\mathbf{p};\gamma) \in \{0,...,255\}^{h_n \times w_n}$, with $\{0,...,255\}$ specifying RGB value. 
 We follow by defining $\mathbf{M}_n$ as the set of all RGB image segments at layer $n$ (with each segment given in terms of (\ref{eq:rgb_mask})), 
 and letting $\mathbf{M} = \{\mathbf{M}_1,...,\mathbf{M}_n\}$.
 
We instantiate the clustering, $\mathsf{C}^{seg}$, via maskSLIC~\cite{irving2016maskslic}, an extension of k-means, with an $l_2$ norm, that clusters features while respecting a mask and automate  selection of the number of clusters via the silhouette method~\cite{rousseeuw1987silhouettes}.

%% file: sec/4_experiments_final.tex
\vspace{-0.1cm}
\section{Experiments}\label{sec:experiments}

\subsection{Implementation details}\label{sec:implementation_details}

For evaluation we use ResNet50~\cite{he2016deep}, VGG16~\cite{simonyan2014very}, MobileNetv3~\cite{howard2019searching}, ViT-b~\cite{vaswani2017attention} and MViT~\cite{fan2021multiscale} to sample a broad range of popular architectures. The exact layers selected are in the appendix, but always include the last feature layer before the fully connected layer to probe concepts closest to the model output. Following previous work~\cite{kim2018interpretability,ghorbani2019towards}, we perform a two-sided t-test on ITCAV scores for multiple runs of the same concept vs.\ different random sets of images to remove statistically insignificant scores and
randomly sample from the Broden dataset~\cite{bau2017network} to get the images, $\bm{\mathcal{I}}_{rnd}$ (Sec.~\ref{sec:itcav}), used in statistical testing and CAV training.
On average, four and all-layer VCCs take 15 minutes and 36 hours, resp., to generate
on an NVIDIA Quadro RTX 6000 GPU. The appendix has more details and VCCs for other layers, models and datasets~\cite{WahCUB_200_2011}.



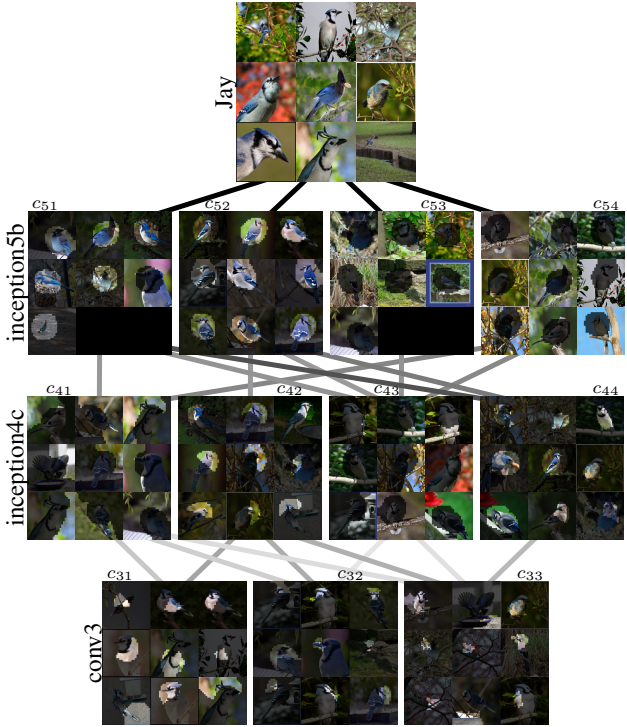
\begin{figure}[t]
    \centering
    \resizebox{0.48\textwidth}{!}{    
    \begin{tikzpicture}[
        every node/.style = {inner sep=0pt},
every label/.append style = {label distance=2pt, align = center},
         sibling distance = 6em,
           level 1/.style = {level distance=4.7em,anchor=north},
           level 2/.style = {level distance=4.5em,anchor=north},
           level 3/.style = {level distance=4.5em,anchor=north},
           ]  
\node[anchor=south,label={[label distance=-0.03cm]left:\rotatebox{90}{Jay}}](class jay)
{\includegraphics[width=25mm]{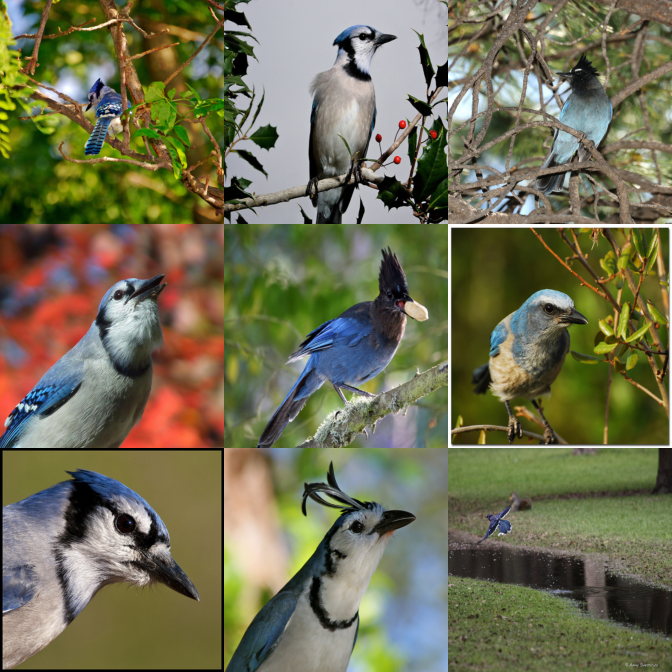}}
    child {node[label={[label distance=-0.03cm]left:\rotatebox{90}{inception5b}},label={[xshift=0.5cm,yshift=-0.05cm]above left:{\scriptsize $c_{51}$}}](inception5b_jay_concept5){\includegraphics[width=20mm]{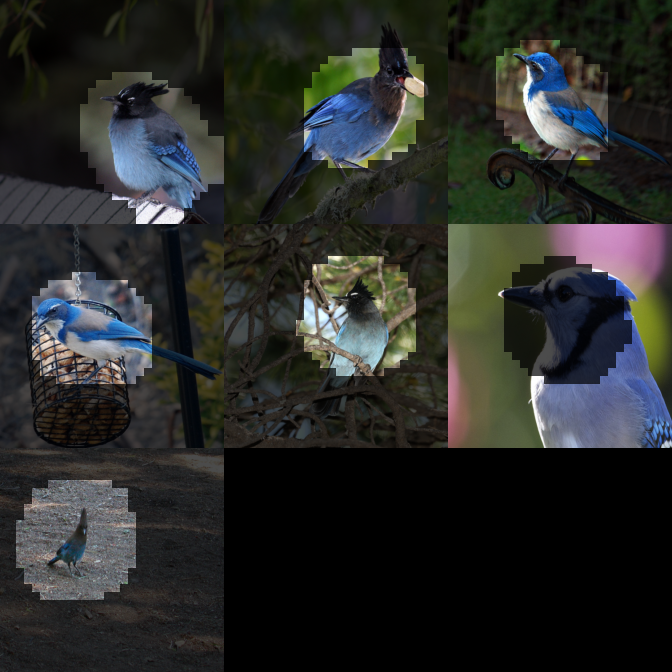}} 
    edge from parent[draw=none]} 
    child {node[label={[xshift=0.8cm,yshift=-0.05cm]above left:{\scriptsize $c_{52}$}}](inception5b_jay_concept4){\includegraphics[width=20mm]{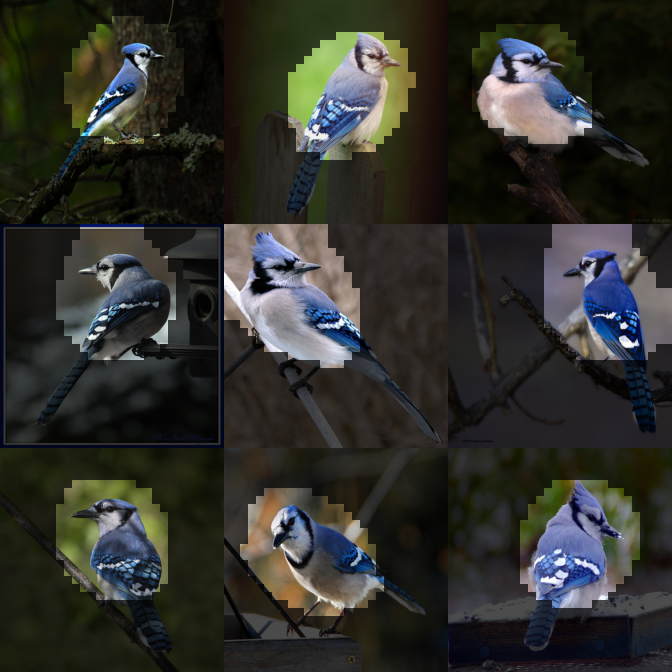}} 
        edge from parent[draw=none]
        child {node[xshift=1.05cm,label={[label distance=-0.03cm]left:\rotatebox{90}{inception4c}},label={[xshift=0.7cm,yshift=-0.05cm]above left:{\scriptsize $c_{41}$}}](inception4c_jay_concept2){\includegraphics[width=20mm]{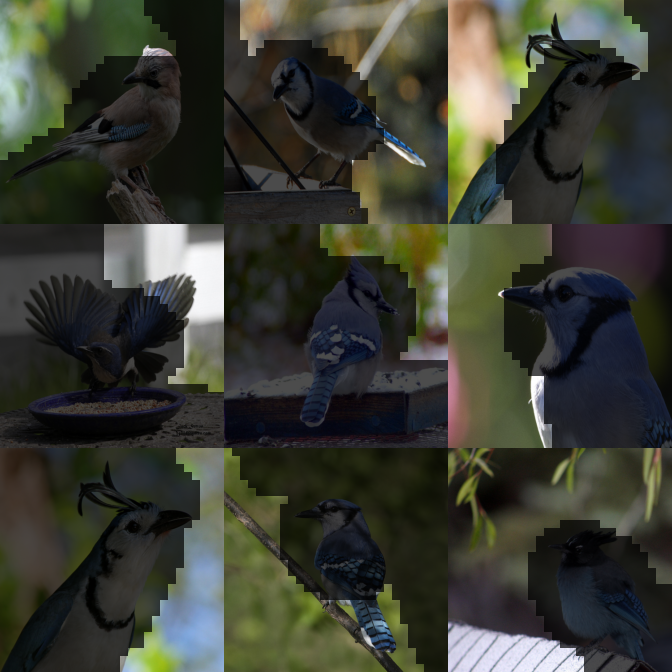}} 
        edge from parent[draw=none]}
        child {node[xshift=1.05cm,label={[xshift=-0.7cm,yshift=-0.05cm]above right:{\scriptsize $c_{42}$}}](inception4c_jay_concept5) {\includegraphics[width=20mm]{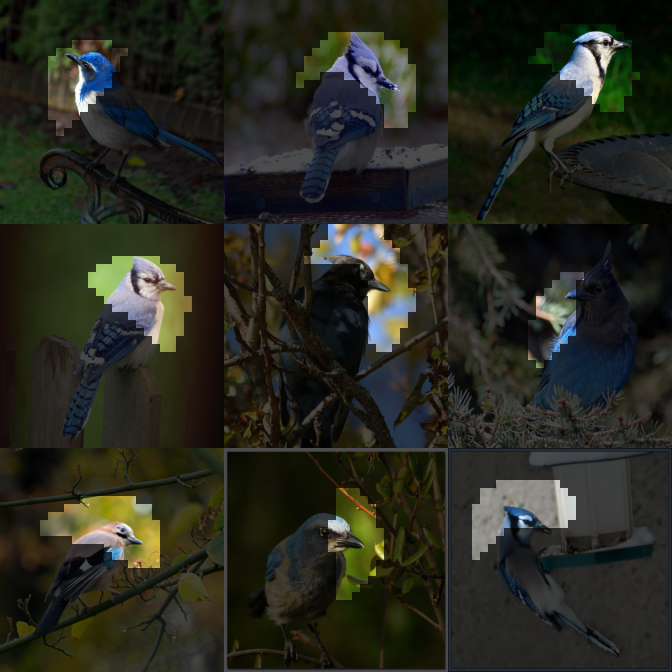}} 
        edge from parent[draw=none]
            child {node[xshift=1.05cm,label={[label distance=-0.0cm]left:\rotatebox{90}{conv3}},label={[xshift=0.5cm,yshift=-0.05cm]above left:{\scriptsize $c_{31}$}}](conv3_jay_concept3){\includegraphics[width=20mm]{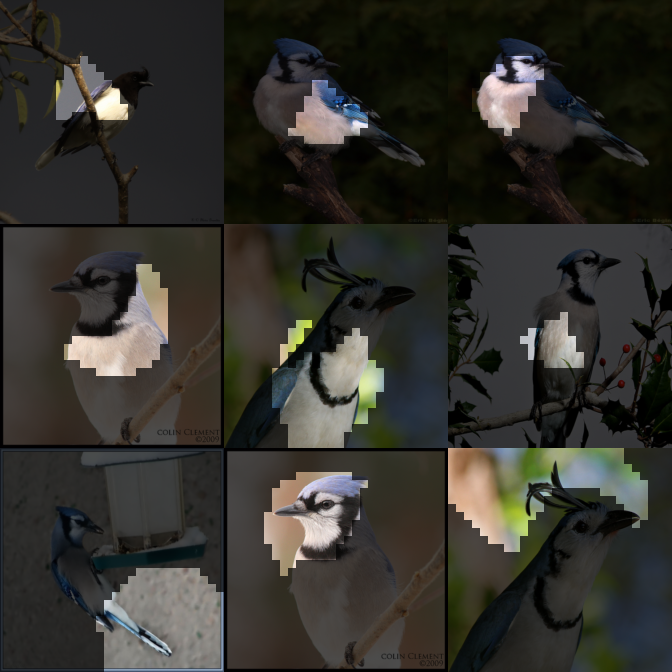}} 
            edge from parent[draw=none]}
            child {node[xshift=1.05cm,label={[xshift=-0.9cm,yshift=-0.05cm]above right:{\scriptsize $c_{32}$}}](conv3_jay_concept2){\includegraphics[width=20mm]{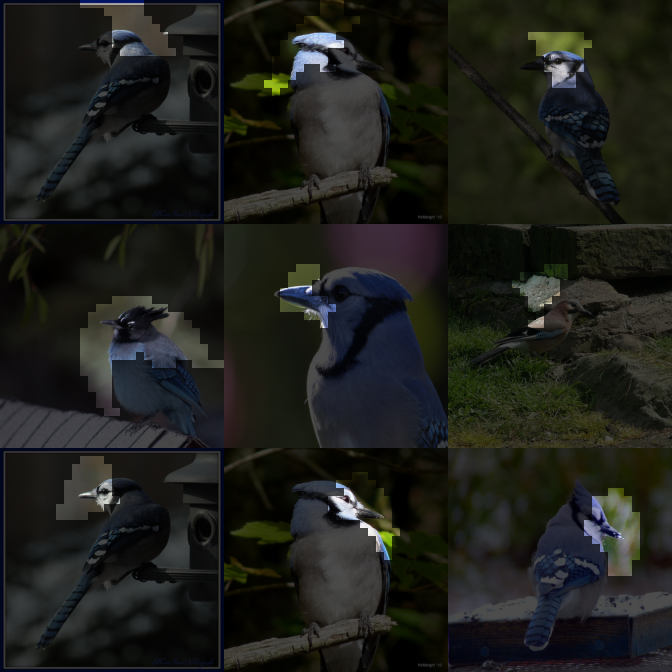}} 
            edge from parent[draw=none]}
            child {node[xshift=1.05cm,label={[xshift=-0.5cm,yshift=-0.05cm]above right:{\scriptsize $c_{33}$}}](conv3_jay_concept1){\includegraphics[width=20mm]{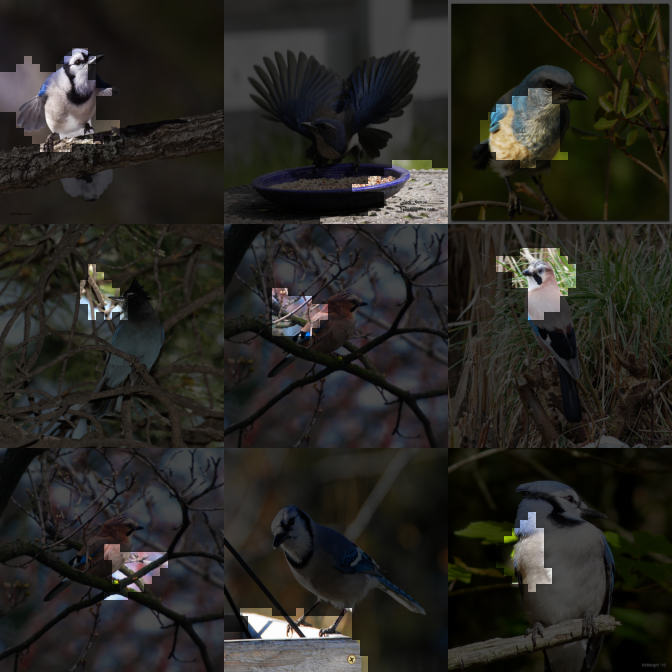}} 
            edge from parent[draw=none]}
            }
        child {node[xshift=1.05cm,label={[xshift=1.06cm,yshift=-0.05cm]above left:{\scriptsize $c_{43}$}}](inception4c_jay_concept1){\includegraphics[width=20mm]{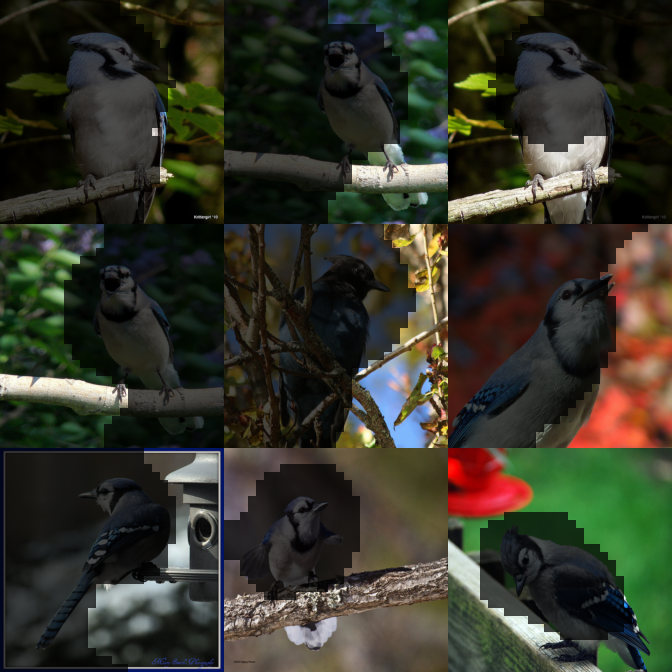}} 
            edge from parent[draw=none]} 
        child {node[xshift=1.05cm,label={[xshift=-0.5cm,yshift=-0.05cm]above right:{\scriptsize $c_{44}$}}](inception4c_jay_concept3){\includegraphics[width=20mm]{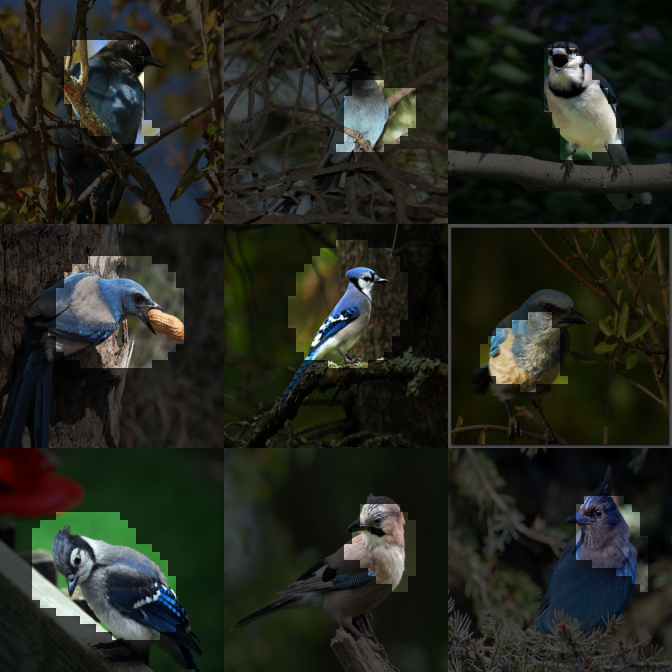}} 
            edge from parent[draw=none]}
            } 
    child {node[label={[xshift=-0.8cm,yshift=-0.05cm]above right:{\scriptsize $c_{53}$}}](inception5b_jay_concept3){\includegraphics[width=20mm]{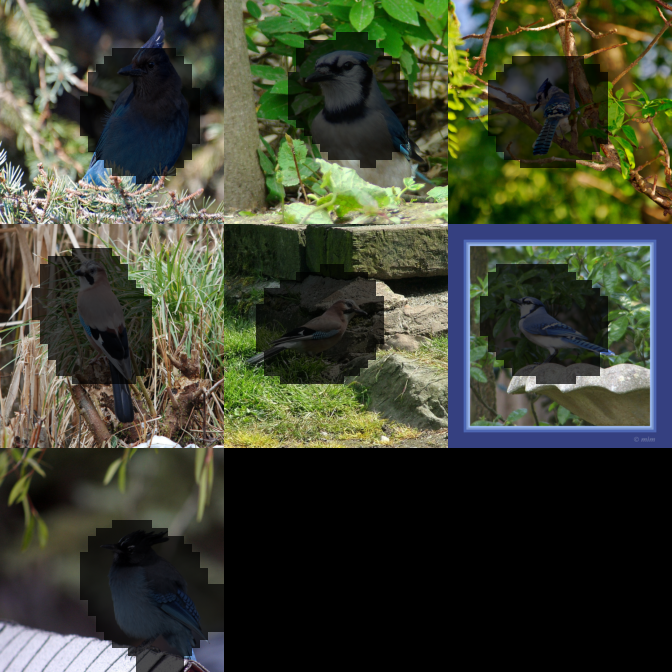}} edge from parent[draw=none]} 
    child {node[label={[xshift=-0.5cm,yshift=-0.05cm]above right:{\scriptsize $c_{54}$}}](inception5b_jay_concept2){\includegraphics[width=20mm]{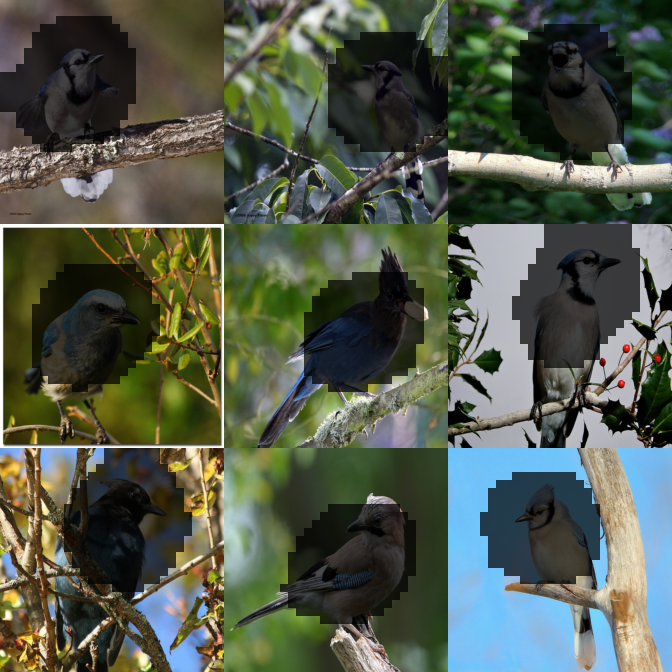}} edge from parent[draw=none]
            };
\begin{scope}[on background layer]
\draw[line width=0.7mm,,-,/pgfplots/color of colormap=(590)] ($ (conv3_jay_concept3.north) - (0,0.2) $) -- ( $ (inception4c_jay_concept2.south) + (0,0.25) $);
\draw[line width=0.7mm,,-,/pgfplots/color of colormap=(457)] ($ (conv3_jay_concept3.north) - (0,0.2) $) -- ( $ (inception4c_jay_concept5.south) + (0,0.25) $);
\draw[line width=0.7mm,,-,/pgfplots/color of colormap=(343)] ($ (inception4c_jay_concept2.north) - (0,0.2) $) -- ( $ (inception5b_jay_concept5.south) + (0,0.25) $);
\draw[line width=0.7mm,,-,/pgfplots/color of colormap=(308)] ($ (inception4c_jay_concept2.north) - (0,0.2) $) -- ( $ (inception5b_jay_concept2.south) + (0,0.25) $);
\draw[line width=0.7mm,,-,/pgfplots/color of colormap=(357)] ($ (inception4c_jay_concept5.north) - (0,0.2) $) -- ( $ (inception5b_jay_concept4.south) + (0,0.25) $);
\draw[line width=0.7mm,,-,/pgfplots/color of colormap=(734)] ($ (conv3_jay_concept2.north) - (0,0.2) $) -- ( $ (inception4c_jay_concept1.south) + (0,0.25) $);
\draw[line width=0.7mm,,-,/pgfplots/color of colormap=(617)] ($ (conv3_jay_concept2.north) - (0,0.2) $) -- ( $ (inception4c_jay_concept2.south) + (0,0.25) $);
\draw[line width=0.7mm,,-,/pgfplots/color of colormap=(510)] ($ (conv3_jay_concept2.north) - (0,0.2) $) -- ( $ (inception4c_jay_concept5.south) + (0,0.25) $);
\draw[line width=0.7mm,,-,/pgfplots/color of colormap=(329)] ($ (inception4c_jay_concept1.north) - (0,0.2) $) -- ( $ (inception5b_jay_concept5.south) + (0,0.25) $);
\draw[line width=0.7mm,,-,/pgfplots/color of colormap=(361)] ($ (inception4c_jay_concept1.north) - (0,0.2) $) -- ( $ (inception5b_jay_concept4.south) + (0,0.25) $);
\draw[line width=0.7mm,,-,/pgfplots/color of colormap=(329)] ($ (inception4c_jay_concept1.north) - (0,0.2) $) -- ( $ (inception5b_jay_concept3.south) + (0,0.25) $);
\draw[line width=0.7mm,,-,/pgfplots/color of colormap=(312)] ($ (inception4c_jay_concept1.north) - (0,0.2) $) -- ( $ (inception5b_jay_concept2.south) + (0,0.25) $);
\draw[line width=0.7mm,,-,/pgfplots/color of colormap=(547)] ($ (conv3_jay_concept1.north) - (0,0.2) $) -- ( $ (inception4c_jay_concept3.south) + (0,0.25) $);
\draw[line width=0.7mm,,-,/pgfplots/color of colormap=(675)] ($ (conv3_jay_concept1.north) - (0,0.2) $) -- ( $ (inception4c_jay_concept1.south) + (0,0.25) $);
\draw[line width=0.7mm,,-,/pgfplots/color of colormap=(657)] ($ (conv3_jay_concept1.north) - (0,0.2) $) -- ( $ (inception4c_jay_concept2.south) + (0,0.25) $);
\draw[line width=0.7mm,,-,/pgfplots/color of colormap=(520)] ($ (conv3_jay_concept1.north) - (0,0.2) $) -- ( $ (inception4c_jay_concept5.south) + (0,0.25) $);
\draw[line width=0.7mm,,-,/pgfplots/color of colormap=(229)] ($ (inception4c_jay_concept3.north) - (0,0.2) $) -- ( $ (inception5b_jay_concept5.south) + (0,0.25) $);
\draw[line width=0.7mm,,-,/pgfplots/color of colormap=(291)] ($ (inception4c_jay_concept3.north) - (0,0.2) $) -- ( $ (inception5b_jay_concept4.south) + (0,0.25) $);
\draw[line width=0.7mm,,-,/pgfplots/color of colormap=(0)] ($(inception5b_jay_concept5.north) - (0,0.3) $) -- ( $ (class jay.south) + (0,0.3) $);
\draw[line width=0.7mm,,-,/pgfplots/color of colormap=(0)] ($(inception5b_jay_concept4.north) - (0,0.3) $) -- ( $ (class jay.south) + (0,0.3) $);
\draw[line width=0.7mm,,-,/pgfplots/color of colormap=(0)] ($(inception5b_jay_concept3.north) - (0,0.3) $) -- ( $ (class jay.south) + (0,0.3) $);
\draw[line width=0.7mm,,-,/pgfplots/color of colormap=(0)] ($(inception5b_jay_concept2.north) - (0,0.3) $) -- ( $ (class jay.south) + (0,0.3) $);
\end{scope}
    \end{tikzpicture}
}\vspace{-0.3cm}
\caption{A VCC for three selected layers of a GoogLeNet model~\cite{szegedy2015going} targeting recognition of class `Jay'. Darker lines denote stronger connection weights. \label{fig:goog_jay} 
} \vspace{-0.78cm}
\end{figure}

\subsection{VCC component validation}\label{sec:validation}

\textbf{Segment proposal validation.} Our approach segments concepts based on the feature space of a given layer to allow us to capture concepts across multiple layers (Sec.~\ref{sec:divisive_cluster}). So, the spatial support of valid segments at a layer should follow the receptive field (RF) of filters at that layer, as the segments are constrained by the filters from which they are derived. For comparison, we consider ACE~\cite{ghorbani2019towards}, which does segmentation with a layer independent feature (color), limited to the input image to discover concepts at a single layer for a single model. Being directly related to the features at a layer, our approach should produce concepts that better respect the RF at that layer compared to the baseline, whose features are layer independent (\eg at CNN early layers the RF is small; so, intuitively the patch proposals should be smaller as well and conversely for the later layers).



Figure~\ref{fig:validation} (a) shows the average concept size (in terms of the ratio of segment pixels to the entire image) vs. the RF at the corresponding layer for three different models: (i) ResNet50~\cite{he2016deep}, (ii) VGG16~\cite{simonyan2014very} and (iii) ViT-b~\cite{dosovitskiy2020image} (note we use mean attention distance instead of receptive field for ViT-b). While the baseline~\cite{ghorbani2019towards} produces segments with sizes invariant to the layer analyzed, our approach produces segments with image sizes that scale with the RF.
Additional results with other models (including more transformers) are given in the appendix, which further support this observation. Overall, it is expected by design and we find that our segment proposals follow RFs at each layer.

\textbf{Concept fidelity.} Concept fidelity measures the meaningfulness of discovered concepts with respect to the target model. 
While other work focuses on single layer concept fidelity~\cite{kim2018interpretability,ghorbani2019towards,fel2023craft}, we measure the effect on a model's output when suppressing the encoding of concepts as the information propagates through the model. The expectation is that suppressing the discovered concepts should result in a quicker decrease in performance compared to a non-concept direction (\eg a randomly selected direction) as the amount of suppression gets larger. For a given model and category, we compute the concepts at various layers using our method (Sec.~\ref{sec:concept_discovery_clustering}). We perform concept suppression for a given image, $\bm{\mathcal{I}}^i$, at a given layer, $j$, for concept $\textbf{q}_j$, according to $\textbf{z}^i_j = \textbf{z}^i_j -\epsilon \cdot \textbf{q}_j \big/ ||\textbf{q}_j||_2$,
where $\epsilon$ controls the degree of perturbation. If the concepts discovered are meaningful, then the accuracy for the model's target class should decrease faster compared to random perturbations, $\mathbf{q}_{rnd}$, as $\epsilon$ increases.  

Figure~\ref{fig:validation} (b) has concept suppression results averaged over 50 randomly selected ImageNet~\cite{deng2009imagenet} classes (documented in the appendix), where a concept is randomly chosen to be suppressed at each layer. 
Note that the $\epsilon$ values required to reduce the model accuracy to zero differs by model; so, we scale the $\epsilon$ values to $[0,1]$ for visualization purposes. For all models, it is seen that perturbing in the opposite direction of a recovered concept more severely impacts the accuracy than a random perturbation. 

\begin{figure} [t]
	\begin{center}
     \centering 
\resizebox{0.48\textwidth}{!}{
\hspace{0.1cm}
\begin{tikzpicture} 
                 \begin{axis}[
                 line width=1.0,
                 title style={at={(axis description cs:0.5,1.2)},anchor=north,font=\normalsize},
                 ylabel={Branching Factor},
                 xmin=0.5, xmax=4.5,
                 ymin=0, ymax=8,
                 xtick={1,2,3,4},
                 x tick label style={font=\tiny, rotate=0, anchor=north},
                 y tick label style={font=\tiny},
                 x label style={at={(axis description cs:0.5,0.09)},anchor=north,font=\small},
                 y label style={at={(axis description cs:0.2,.5)},anchor=south,font=\small},
                 width=6.5cm,
                 height=4.0cm, 
                 ymajorgrids=false,
                 xmajorgrids=false,
                 major grid style={dotted,green!20!black},
             ]

            \addplot[line width=0.9pt, mark size=1.8pt, color=cyan, mark=o,error bars/.cd, y dir=both, y explicit,]
                     coordinates {(1, 5.484)
                                  (2, 4.945)
                                  (3, 2.799)
                                  (4, 2.915)};
            \addplot[line width=0.9pt, mark size=1.8pt, color=cyan, mark=diamond,error bars/.cd, y dir=both, y explicit,]
                     coordinates {(1, 6.765)
                                  (2, 6.249)
                                  (3, 4.599)
                                  (4, 1.696)};
            \addplot[line width=0.9pt, mark size=1.8pt, color=cyan, mark=triangle,error bars/.cd, y dir=both, y explicit,]
                     coordinates {(1, 7.102)
                                  (2, 5.175)
                                  (3, 4.3)
                                  (4, 1.149)};
            \addplot[line width=0.9pt, mark size=2pt, color=red, mark=x,error bars/.cd, y dir=both, y explicit,]
                     coordinates {(1, 5.659)
                                  (2, 5.886)
                                  (3, 6.585)
                                  (4, 2.043)};
            \addplot[line width=0.9pt, mark size=2pt, color=red, mark=+,error bars/.cd, y dir=both, y explicit,]
                     coordinates {(1, 2.967)
                                  (2, 2.044)
                                  (3, 3.337)
                                  (4, 1.66)};   
            \addplot[line width=0.9pt, mark size=2pt, color=blue, mark=o,error bars/.cd, y dir=both, y explicit,]
                     coordinates {(1, 3.748)
                                  (2, 3.837)
                                  (3, 4.57)
                                  (4, 1.042)};   
            \addplot[line width=0.9pt, mark size=2pt, color=green, mark=o,error bars/.cd, y dir=both, y explicit,]
                     coordinates {(1, 4.463)
                                  (2, 4.992)
                                  (3, 1.972)
                                  (4, 0.854)};
            \addplot[line width=0.9pt, mark size=1.8pt, color=blue, mark=diamon,error bars/.cd, y dir=both, y explicit,]
                     coordinates {(1, 4.247)
                                  (2, 4.42)
                                  (3, 2.927)
                                  (4, 1.458)};    
              \end{axis}
\end{tikzpicture}
\hspace{0.1cm}
\begin{tikzpicture} 
                  \begin{axis}[
                 line width=1.0,
                 title style={at={(axis description cs:0.5,1.2)},anchor=north,font=\normalsize},
                 ylabel={Number of Concepts},
                 xmin=0.5, xmax=4.5,
                 ymin=1.5, ymax=12.5,
                 xtick={1,2,3,4},
                 x tick label style={font=\tiny, rotate=0, anchor=north},
                 y tick label style={font=\tiny},
                 x label style={at={(axis description cs:0.5,0.09)},anchor=north,font=\small},
                 y label style={at={(axis description cs:0.2,.5)},anchor=south,font=\small},
                 width=6.5cm,
                 height=4.0cm, 
                 ymajorgrids=false,
                 xmajorgrids=false,
                 major grid style={dotted,green!20!black},
                 legend columns=2,
                 legend style={
                  nodes={scale=0.7, transform shape},
                  cells={anchor=west},
                  legend style={at={(0.5,1)},anchor=north,row sep=0.01pt}, font =\small}
             ]

            \addplot[line width=0.9pt, mark size=1.8pt, color=cyan, mark=o,error bars/.cd, y dir=both, y explicit,]
                     coordinates {(1,10.447)
                                  (2, 8)
                                  (3, 5.106)
                                  (4, 2.957)};
            \addplot[line width=0.9pt, mark size=1.8pt, color=cyan, mark=diamond,error bars/.cd, y dir=both, y explicit,]
                     coordinates {(1, 11.174)
                                  (2, 9.348)
                                  (3, 6.348)
                                  (4, 3.087)};
            \addplot[line width=0.9pt, mark size=1.8pt, color=cyan, mark=triangle,error bars/.cd, y dir=both, y explicit,]
                     coordinates {(1, 12.255)
                                  (2, 9.532)
                                  (3, 6.936)
                                  (4, 2.404)};
            \addplot[line width=0.9pt, mark size=2pt, color=red, mark=x,error bars/.cd, y dir=both, y explicit,]
                     coordinates {(1, 11.872)
                                  (2, 9.957)
                                  (3, 6.83)
                                  (4, 2.191)};
            \addplot[line width=0.9pt, mark size=2pt, color=red, mark=+,error bars/.cd, y dir=both, y explicit,]
                     coordinates {(1, 7.957)
                                  (2, 5.596)
                                  (3, 5.766)
                                  (4, 2.298)};   

            \addplot[line width=0.9pt, mark size=2pt, color=blue, mark=o,error bars/.cd, y dir=both, y explicit,]
                     coordinates {(1, 7.083)
                                  (2, 7.188)
                                  (3, 4.979)
                                  (4, 2.875)};
            \addplot[line width=0.9pt, mark size=2pt, color=green, mark=o, mark=square,error bars/.cd, y dir=both, y explicit,]
                     coordinates {(1, 7.833)
                                  (2, 7.375)
                                  (3, 4.021)
                                  (4, 2.292)};
            \addplot[line width=0.9pt, mark size=1.8pt, color=blue, mark=diamond,error bars/.cd, y dir=both, y explicit,]
                     coordinates {(1, 7.104)
                                  (2, 6.5)
                                  (3, 5.333)
                                  (4, 3.396)}; 
              \end{axis}
\end{tikzpicture}}
\resizebox{0.48\textwidth}{!}{
\hspace{0.1cm}
\begin{tikzpicture}
                 \begin{axis}[
                 line width=1.0,
                 title style={at={(axis description cs:0.5,1.2)},anchor=north,font=\normalsize},
                 xlabel={Layer},
                 ylabel={Edge Weight Average},
                 xmin=0.5, xmax=4.5,
                 ymin=0.2, ymax=1.2,
                 xtick={1,2,3,4},
                 x tick label style={font=\tiny, rotate=0, anchor=north},
                 y tick label style={font=\tiny},
                 x label style={at={(axis description cs:0.5,0.15)},anchor=north,font=\small},
                 y label style={at={(axis description cs:0.2,.5)},anchor=south,font=\small},
                 width=6.5cm,
                 height=4.0cm, 
                 ymajorgrids=false,
                 xmajorgrids=false,
                 major grid style={dotted,green!20!black},
                 legend columns=2,
                 legend style={
                  nodes={scale=0.42, transform shape},
                  cells={anchor=west},
                  legend style={at={(0,1)},anchor=north west,row sep=0.01pt}, font =\small}
             ]
            \addlegendentry{ResNet50}
            \addplot[line width=0.9pt, mark size=1.8pt, color=cyan, mark=o,error bars/.cd, y dir=both, y explicit,]
                     coordinates {(1, 0.414)
                                  (2, 0.554)
                                  (3, 0.476)
                                  (4, 0.917)};
             \addlegendentry{MViT}
            \addplot[line width=0.9pt, mark size=2pt, color=red, mark=x,error bars/.cd, y dir=both, y explicit,]
                     coordinates {(1, 0.552)
                                  (2, 0.384)
                                  (3, 0.579)
                                  (4, 0.908)};
            \addlegendentry{VGG16}
            \addplot[line width=0.9pt, mark size=1.8pt, color=cyan, mark=diamond,error bars/.cd, y dir=both, y explicit,]
                     coordinates {(1, 0.329)
                                  (2, 0.272)
                                  (3, 0.429)
                                  (4, 0.801)};
            \addlegendentry{ViT}
            \addplot[line width=0.9pt, mark size=2pt, color=red, mark=+,error bars/.cd, y dir=both, y explicit,]
                     coordinates {(1, 0.471)
                                  (2, 0.529)
                                  (3, 0.395)
                                  (4, 0.755)};     
            \addlegendentry{MobileNet-v2}
            \addplot[line width=0.9pt, mark size=1.8pt, color=cyan, mark=triangle,error bars/.cd, y dir=both, y explicit,]
                     coordinates {(1, 0.500)
                                  (2, 0.498)
                                  (3, 0.623)
                                  (4, 0.75)};                      
            \addlegendentry{ResNet50 (ADV)}
            \addplot[line width=0.9pt, mark size=2pt, color=blue, mark=o,error bars/.cd, y dir=both, y explicit,]
                     coordinates {(1, 0.531)
                                  (2, 0.367)
                                  (3, 0.801)
                                  (4, 0.547)};
            \addlegendentry{ResNet50 (SimCLR)}
            \addplot[line width=0.9pt, mark size=2pt, color=green, mark=o,error bars/.cd, y dir=both, y explicit,]
                     coordinates {(1, 0.342)
                                  (2, 0.38)
                                  (3, 0.584)
                                  (4, 0.447)};
            \addlegendentry{VGG16 (ADV)}                             
            \addplot[line width=0.9pt, mark size=1.8pt, color=blue, mark=diamond,error bars/.cd, y dir=both, y explicit,]
                     coordinates {(1, 0.517)			
                                  (2, 0.247)
                                  (3, 0.474)
                                  (4, 0.716)};  
              \end{axis}
\end{tikzpicture}
\hspace{0.1cm}
\begin{tikzpicture} 
                  \begin{axis}[
                 line width=1.0,
                 title style={at={(axis description cs:0.5,1.2)},anchor=north,font=\normalsize},
                 xlabel={Layer},
                 ylabel={Edge Weight Variance},
                 xmin=0.5, xmax=4.5,
                 ymin=0, ymax=0.03,
                 xtick={1,2,3,4},
                 x tick label style={font=\tiny, rotate=0, anchor=north},
                 y tick label style={font=\tiny},
                 x label style={at={(axis description cs:0.5,0.15)},anchor=north,font=\small},
                 y label style={at={(axis description cs:0.2,.5)},anchor=south,font=\small},
                 width=6.5cm,
                 height=4.0cm, 
                 ymajorgrids=false,
                 xmajorgrids=false,
                 major grid style={dotted,green!20!black},
                    legend columns=2,
                 legend style={
                  nodes={scale=0.45, transform shape},
                  cells={anchor=west},
                  legend style={at={(0.45,1)},anchor=north,row sep=0.01pt}, font =\small}
             ]

            \addplot[line width=0.9pt, mark size=1.8pt, color=cyan, mark=o,error bars/.cd, y dir=both, y explicit,]
                     coordinates {(1, 0.003)
                                  (2, 0.007)
                                  (3, 0.01)
                                  (4, 0.007)};
            \addplot[line width=0.9pt, mark size=2pt, color=red, mark=x,error bars/.cd, y dir=both, y explicit,]
                     coordinates {(1, 0.003)
                                  (2, 0.003)
                                  (3, 0.005)
                                  (4, 0.016)};
            \addplot[line width=0.9pt, mark size=1.8pt, color=cyan, mark=diamond,error bars/.cd, y dir=both, y explicit,]
                     coordinates {(1, 0.008)
                                  (2, 0.013)
                                  (3, 0.007)
                                  (4, 0.003)};
            \addplot[line width=0.9pt, mark size=2pt, color=red, mark=+,error bars/.cd, y dir=both, y explicit,]
                     coordinates {(1, 0.011)
                                  (2, 0.007)
                                  (3, 0.008)
                                  (4, 0.028)};          
            \addplot[line width=0.9pt, mark size=1.8pt, color=cyan, mark=triangle,error bars/.cd, y dir=both, y explicit,]
                     coordinates {(1, 0.008)
                                  (2, 0.008)
                                  (3, 0.006)
                                  (4, 0.004)};
            \addplot[line width=0.9pt, mark size=2pt, color=blue, mark=o,error bars/.cd, y dir=both, y explicit,]
                     coordinates {(1, 0.003)
                                  (2, 0.006)
                                  (3, 0.008)
                                  (4, 0.003)};
            \addplot[line width=0.9pt, mark size=2pt, color=green, mark=o,error bars/.cd, y dir=both, y explicit,]
                     coordinates {(1, 0.005)
                                  (2, 0.004)
                                  (3, 0.009)
                                  (4, 0.002)};
            \addplot[line width=0.9pt, mark size=1.8pt, color=blue, mark=diamond,error bars/.cd, y dir=both, y explicit,]
                     coordinates {(1, 0.006)
                                  (2, 0.008)
                                  (3, 0.023)
                                  (4, 0.004)};    
                             
              \end{axis}
\end{tikzpicture}
}
\end{center}
\vspace{-0.8cm}
\caption{Graph metrics on four layer VCCs comparing CNN vs. transformer architectures and training objectives.
}\label{fig:cnn_trans}
\vspace{-0.6cm}
\end{figure}
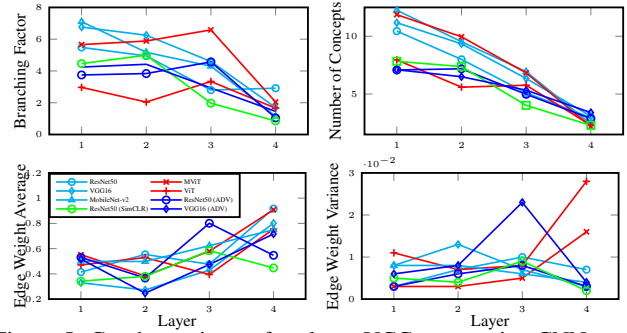

\textbf{ITCAV validation.} If the interlayer concept connection weights are meaningful, then the accumulated weights from earlier to later layer concepts should be correlated with their probability of predicting the target class.
We use this intuition to validate our ITCAV approach to recovering interlayer concept weights as follows. 
We define the average path strength (APS), a scalar value between zero and one representing the average strength of the connection between a concept and the class logit. Given concept $\textbf{q}_j$ at layer $j$, let there be $L$ layers in the VCC between the concept and the logit layer; thus, each path from the concept to the logit consists of $L$ edge weights. We consider all possible paths in the VCC from this concept to the class logit. Let $e_{\textbf{q}_j}(l,p)$ be the edge weight (\ie ITCAV score) of the $p^{\text{th}}$ path (from concept $\textbf{q}_j$ to the class concept) for VCC layer $l$. To calculate the APS score for a given concept, we average over all paths, $P$, and edge weights in each path to get
\begin{equation}\label{eq:APS}
    \text{APS}(\textbf{q}_j) = \frac{1}{P}\sum_{p=1}^P \big[\frac{1}{L}\sum_{l=1}^L e_{\textbf{q}_j}(l,p)\big].
\end{equation}
Next, we calculate the logit sum (LS) score for a given concept, $\textbf{q}_j$, by passing all of the corresponding masked segments, $\mathbf{M}_{\textbf{q}_j}$, through the model, $F(\cdot)$. We then sum up the logit scores for the target class, $c$, to yield
\vspace{-4pt}
\begin{equation}\label{eq:LS}
    \text{LS}(\textbf{q}_j) = \sum_{i=1}^{|\mathbf{M}_{\textbf{q}_j}|} F(\mathbf{M}^i_{\textbf{q}_j})|_c,
\end{equation}
where $\mathbf{M}^i_{\textbf{q}_j}$ denotes the $i^{\text{th}}$ segment mask, and $F(\cdot)|_c$ denotes the $c^{\text{th}}$ logit score. If the ITCAV edge weights are meaningful, then there should be positive correlation between the APS and LS scores, \ie $ \text{APS(\textbf{q})} \propto \text{LS(\textbf{q})}$.

Figure~\ref{fig:validation} (c, d) shows the LS scores plotted vs. the APS scores for 
VGG16~\cite{he2016deep} and ViT-b~\cite{vaswani2017attention}.
A positive correlation is found between the APS and LS scores for both architectures. These results suggest that the combination of ITCAV scores is predictive of whether a concept is representative of the target class. The appendix has more LS vs.\ APS plots for more models that support these findings.


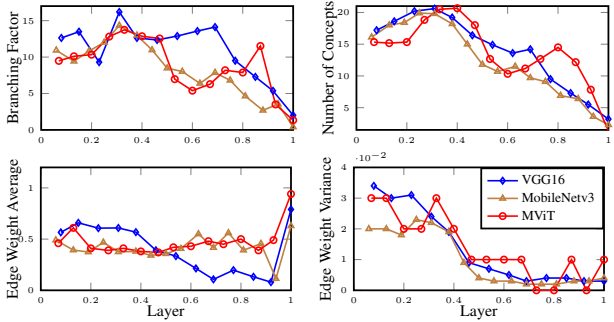
\begin{figure} [t]
	\begin{center}
     \centering 
\resizebox{0.48\textwidth}{!}{
\hspace{0.1cm}
\begin{tikzpicture} 
                 \begin{axis}[
                 line width=1.0,
                 title style={at={(axis description cs:0.5,1.2)},anchor=north,font=\normalsize},
                 ylabel={Branching Factor},
                 xmin=0, xmax=1,
                 ymin=0, ymax=17,
                 xtick={0,0.2,0.4,0.6,0.8,1},
                 x tick label style={font=\tiny, rotate=0, anchor=north},
                 y tick label style={font=\tiny},
                 x label style={at={(axis description cs:0.5,0.09)},anchor=north,font=\small},
                 y label style={at={(axis description cs:0.2,.5)},anchor=south,font=\small},
                 width=6.5cm,
                 height=4.0cm, 
                 ymajorgrids=false,
                 xmajorgrids=false,
                 major grid style={dotted,green!20!black},
             ]
            \addplot[line width=0.9pt, mark size=1.8pt, color=blue, mark=diamond,error bars/.cd, y dir=both, y explicit,]
                     coordinates {(0.08,12.627)
                                    (0.15,13.484)
                                    (0.23,9.301)
                                    (0.31,16.194)
                                    (0.38,12.628)
                                    (0.46,12.349)
                                    (0.54,12.888)
                                    (0.62,13.575)
                                    (0.69,14.122)
                                    (0.77,9.5)
                                    (0.85,7.28)
                                    (0.92,5.367)
                                    (1.00,2)};

            \addplot[line width=0.9pt, mark size=1.8pt, color=brown, mark=triangle,error bars/.cd, y dir=both, y explicit,]
                     coordinates {
                                (0.06,10.927)
                                (0.13,9.423)
                                (0.19,10.839)
                                (0.25,12.01)
                                (0.31,14.332)
                                (0.38,12.99)
                                (0.44,10.975)
                                (0.50,8.478)
                                (0.56,8.016)
                                (0.63,6.384)
                                (0.69,7.857)
                                (0.75,6.81)
                                (0.81,4.642)
                                (0.88,2.685)
                                (0.94,3.55)
                                (1.00,0.4)};     

            \addplot[line width=0.9pt, mark size=1.8pt, color=red, mark=o,error bars/.cd, y dir=both, y explicit,]
                     coordinates {
(0.07,	9.48)
(0.13,	10.11)
(0.20,	10.33)
(0.27,	12.81)
(0.33,	13.74)
(0.40,	12.86)
(0.47,	12.55)
(0.53,	6.97)
(0.60,	5.39)
(0.67,	6.28)
(0.73,	8.19)
(0.80,	7.88)
(0.87,	11.52)
(0.93,	3.50)
(1.00,	1.33)
};   
              \end{axis}
\end{tikzpicture}
\hspace{0.1cm}
\begin{tikzpicture} 
                  \begin{axis}[
                 line width=1.0,
                 title style={at={(axis description cs:0.5,1.2)},anchor=north,font=\normalsize},
                 ylabel={Number of Concepts},
                 xmin=0, xmax=1,
                 ymin=1.5, ymax=21,
                 xtick={0,0.2,0.4,0.6,0.8,1},
                 x tick label style={font=\tiny, rotate=0, anchor=north},
                 y tick label style={font=\tiny},
                 x label style={at={(axis description cs:0.5,0.09)},anchor=north,font=\small},
                 y label style={at={(axis description cs:0.2,.5)},anchor=south,font=\small},
                 width=6.5cm,
                 height=4.0cm, 
                 ymajorgrids=false,
                 xmajorgrids=false,
                 major grid style={dotted,green!20!black},
                 legend columns=2,
                 legend style={
                  nodes={scale=0.7, transform shape},
                  cells={anchor=west},
                  legend style={at={(0.5,1)},anchor=north,row sep=0.01pt}, font =\small}
             ]
            \addplot[line width=0.9pt, mark size=1.8pt, color=blue, mark=diamond,error bars/.cd, y dir=both, y explicit,]
                     coordinates {(0.08,17.2)
                                (0.15,18.6)
                                (0.23,20.2)
                                (0.31,20.6)
                                (0.38,19.2)
                                (0.46,16.4)
                                (0.54,14.9)
                                (0.62,13.6)
                                (0.69,14.2)
                                (0.77,9.5)
                                (0.85,7.3)
                                (0.92,5.5)
                                (1.00,3.2)};

            \addplot[line width=0.9pt, mark size=1.8pt, color=brown, mark=triangle,error bars/.cd, y dir=both, y explicit,]
                     coordinates {
                                (0.06,16.1)
                                (0.13,18)
                                (0.19,18.4)
                                (0.25,19.9)
                                (0.31,19.8)
                                (0.38,18.2)
                                (0.44,15)
                                (0.50,11.8)
                                (0.56,10.7)
                                (0.63,11.5)
                                (0.69,9.7)
                                (0.75,9.1)
                                (0.81,6.9)
                                (0.88,6.4)
                                (0.94,3.6)
                                (1.00,2.3)};   
            \addplot[line width=0.9pt, mark size=1.8pt, color=red, mark=o,error bars/.cd, y dir=both, y explicit,]
                     coordinates {
(0.07,	15.33)
(0.13,	15.17)
(0.20,	15.33)
(0.27,	18.83)
(0.33,	20.50)
(0.40,	20.67)
(0.47,	18.00)
(0.53,	12.67)
(0.60,	10.33)
(0.67,	11.17)
(0.73,	12.67)
(0.80,	14.50)
(0.87,	12.17)
(0.93,	7.83)
(1.00,	1.33)
};   
              \end{axis}
\end{tikzpicture}}
\resizebox{0.48\textwidth}{!}{
\hspace{0.1cm}
\begin{tikzpicture}
                 \begin{axis}[
                 line width=1.0,
                 title style={at={(axis description cs:0.5,1.2)},anchor=north,font=\normalsize},
                 xlabel={Layer},
                 ylabel={Edge Weight Average},
                 xmin=0, xmax=1,
                 ymin=0, ymax=1.2,
                 xtick={0,0.2,0.4,0.6,0.8,1},
                 x tick label style={font=\tiny, rotate=0, anchor=north},
                 y tick label style={font=\tiny},
                 x label style={at={(axis description cs:0.5,0.15)},anchor=north,font=\small},
                 y label style={at={(axis description cs:0.2,.5)},anchor=south,font=\small},
                 width=6.5cm,
                 height=4.0cm, 
                 ymajorgrids=false,
                 xmajorgrids=false,
                 major grid style={dotted,green!20!black},
                 legend columns=2,
                 legend style={
                  nodes={scale=0.6, transform shape},
                  cells={anchor=west},
                  legend style={at={(0.5,1)},anchor=north,row sep=0.01pt}, font =\small}
             ]

            \addplot[line width=0.9pt, mark size=1.8pt, color=blue, mark=diamond,error bars/.cd, y dir=both, y explicit,]
                     coordinates {
                                (0.08,0.566)
                                (0.15,0.659)
                                (0.23,0.607)
                                (0.31,0.609)
                                (0.38,0.568)
                                (0.46,0.391)
                                (0.54,0.334)
                                (0.62,0.214)
                                (0.69,0.107)
                                (0.77,0.197)
                                (0.85,0.132)
                                (0.92,0.08)
                                (1.00,0.79)};

            \addplot[line width=0.9pt, mark size=1.8pt, color=brown, mark=triangle,error bars/.cd, y dir=both, y explicit,]
                     coordinates {
                                (0.06,0.49)
                                (0.13,0.393)
                                (0.19,0.374)
                                (0.25,0.468)
                                (0.31,0.375)
                                (0.38,0.38)
                                (0.44,0.341)
                                (0.50,0.357)
                                (0.56,0.407)
                                (0.63,0.55)
                                (0.69,0.417)
                                (0.75,0.56)
                                (0.81,0.392)
                                (0.88,0.455)
                                (0.94,0.114)
                                (1.00,0.629)};  
            \addplot[line width=0.9pt, mark size=1.8pt, color=red, mark=o,error bars/.cd, y dir=both, y explicit,]
                     coordinates {
(0.07,	0.46)
(0.13,	0.61)
(0.20,	0.41)
(0.27,	0.39)
(0.33,	0.41)
(0.40,	0.38)
(0.47,	0.37)
(0.53,	0.42)
(0.60,	0.43)
(0.67,	0.48)
(0.73,	0.45)
(0.80,	0.50)
(0.87,	0.39)
(0.93,	0.49)
(1.00,	0.94)
};   
              \end{axis}
\end{tikzpicture}
\hspace{0.1cm}
\begin{tikzpicture} 
                  \begin{axis}[
                 line width=1.0,
                 title style={at={(axis description cs:0.5,1.2)},anchor=north,font=\normalsize},
                 xlabel={Layer},
                 ylabel={Edge Weight Variance},
                 xmin=0, xmax=1,
                 ymin=0, ymax=0.04,
                 xtick={0,0.2,0.4,0.6,0.8,1},
                 x tick label style={font=\tiny, rotate=0, anchor=north},
                 y tick label style={font=\tiny},
                 x label style={at={(axis description cs:0.5,0.15)},anchor=north,font=\small},
                 y label style={at={(axis description cs:0.2,.5)},anchor=south,font=\small},
                 width=6.5cm,
                 height=4.0cm, 
                 ymajorgrids=false,
                 xmajorgrids=false,
                 major grid style={dotted,green!20!black},
                    legend columns=1,
                 legend style={
                  nodes={scale=0.85, transform shape},
                  cells={anchor=west},
                  legend style={at={(1,1)},anchor=north east,row sep=0.01pt}, font =\small}
             ]
            \addlegendentry{VGG16}
            \addplot[line width=0.9pt, mark size=1.8pt, color=blue, mark=diamond,error bars/.cd, y dir=both, y explicit,]
                     coordinates {
                                (0.08,0.034)
                                (0.15,0.03)
                                (0.23,0.031)
                                (0.31,0.024)
                                (0.38,0.019)
                                (0.46,0.009)
                                (0.54,0.007)
                                (0.62,0.005)
                                (0.69,0.003)
                                (0.77,0.004)
                                (0.85,0.004)
                                (0.92,0.003)
                                (1.00,0.003)}; 
                                  
            \addlegendentry{MobileNetv3}
            \addplot[line width=0.9pt, mark size=1.8pt, color=brown, mark=triangle,error bars/.cd, y dir=both, y explicit,]
                     coordinates {
                                (0.06,0.02)
                                (0.13,0.02)
                                (0.19,0.018)
                                (0.25,0.023)
                                (0.31,0.022)
                                (0.38,0.019)
                                (0.44,0.009)
                                (0.50,0.004)
                                (0.56,0.003)
                                (0.63,0.003)
                                (0.69,0.002)
                                (0.75,0.002)
                                (0.81,0.002)
                                (0.88,0.003)
                                (0.94,0.003)
                                (1.00,0.004)};
            \addlegendentry{MViT}
            \addplot[line width=0.9pt, mark size=1.8pt, color=red, mark=o,error bars/.cd, y dir=both, y explicit,]
                     coordinates {
(0.07,	0.03)
(0.13,	0.03)
(0.20,	0.02)
(0.27,	0.02)
(0.33,	0.03)
(0.40,	0.02)
(0.47,	0.01)
(0.53,	0.01)
(0.60,	0.01)
(0.67,	0.01)
(0.73,	0.00)
(0.80,	0.00)
(0.87,	0.01)
(0.93,	0.00)
(1.00,	0.01)
}; 
                             
              \end{axis}
\end{tikzpicture}
}
\end{center}
\vspace{-0.75cm}
\caption{Graph metrics of all layer VCCs 
for three diverse architectures.
Layer number normalized to allow for comparison of models with different numbers of layers.
}\label{fig:local_global}
\vspace{-0.6cm}
\end{figure}

\subsection{Understanding models with VCCs}\label{sec:architectures}
To demonstrate VCC's unique ability to interpret models at different resolutions, we present  model analyses in three parts: (i) qualitative analyses of VCCs generated for a subset of layers as well as at all layers, (ii)  quantitative analysis of VCCs generated for a subset of four layers, (iii) a broadened quantitative analysis to encompass \textit{all} model layers.

\textbf{Visualizing concept hierarchies}. Figure~\ref{fig:goog_jay} shows a three layer VCC for  GoogLeNet~\cite{szegedy2015going} targeting the class ``Jay''. Notice how the inception5b `bird' concepts ($c_{51},c_{52}$) form as selective weighting of inception4c concepts background ($c_{41}$), bird part ($c_{42},c_{44}$) and tree branch ($c_{43}$), while the inception5b background concepts ($c_{53},c_{54}$) form differently from weighting of solely inception4c background part concepts (\eg tree branch ($c_{43}$) and green leaves ($c_{41}$)). Notably, the network separates subspecies of Jay in the final layer (\eg Blue Jay ($c_{52}$) and other types ($c_{51}$)). The concepts found in inception4c are composed from varying combinations of colors and parts found in conv3 (\eg various bird parts ($c_{31},c_{33}$) contribute to the bird concepts at inception4c). In the end, both scene and object contribute with strong weights to the final category.

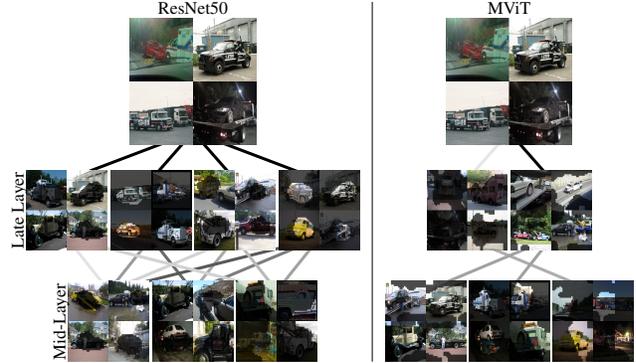
\begin{figure}[t]
    \centering
    \resizebox{0.48\textwidth}{!}{    
    \begin{tikzpicture}[
        every node/.style = {inner sep=0pt},
every label/.append style = {label distance=2pt, align = center},
         sibling distance = 4.75em,
           level 1/.style = {level distance=5em,anchor=north},
           level 2/.style = {level distance=4em,anchor=north},
           level 3/.style = {level distance=3em,anchor=north},
           ]  
\node[anchor=south, label=ResNet50](class_tow_truck)
{\includegraphics[width=25mm]{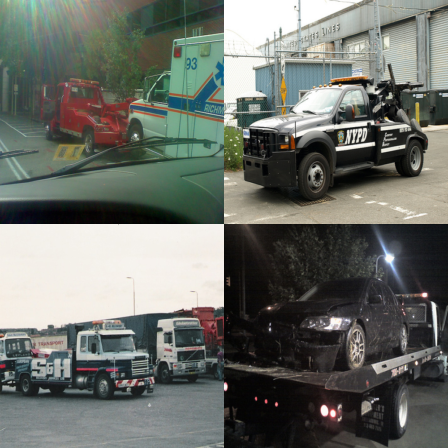}}
    child {node[label={[label distance=-0.03cm]left:\rotatebox{90}{Late Layer}}](layer4_tow_truck_concept1){\includegraphics[width=16mm]{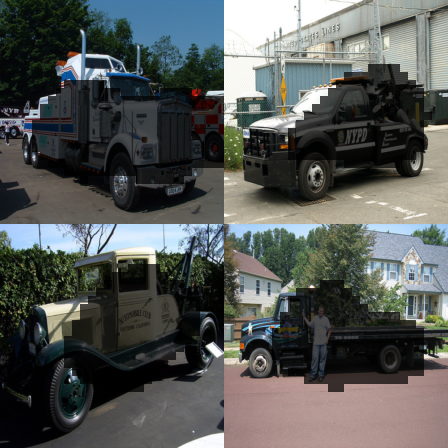}} 
        edge from parent[draw=none]} 
    child {node[](layer4_tow_truck_concept2){\includegraphics[width=16mm]{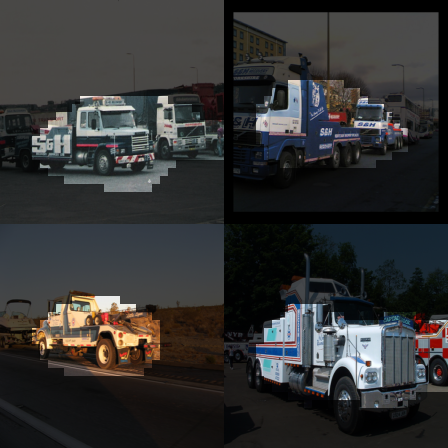}} 
        edge from parent[draw=none]} 
    child {node[](layer4_tow_truck_concept3){\includegraphics[width=16mm]{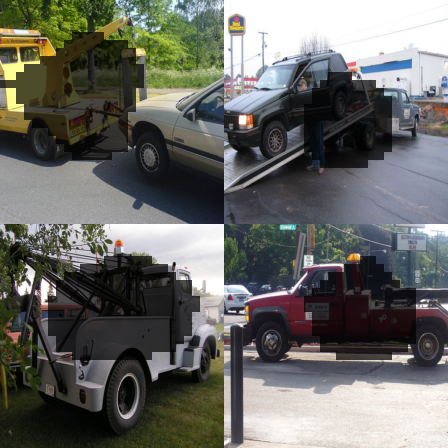}} 
        edge from parent[draw=none]
        child {node[xshift=-0.84cm,label={[label distance=-0.03cm]left:\rotatebox{90}{Mid-Layer}}](layer3_tow_truck_concept1){\includegraphics[width=16mm]{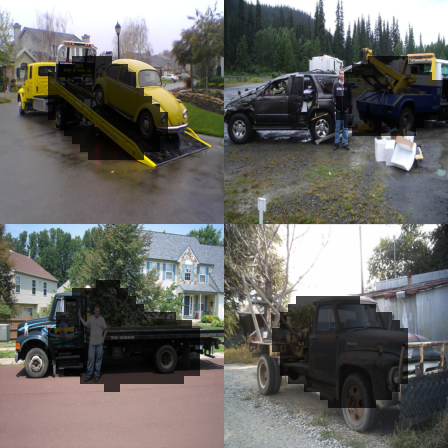}} 
            edge from parent[draw=none]}
        child {node[xshift=-0.84cm](layer3_tow_truck_concept4){\includegraphics[width=16mm]{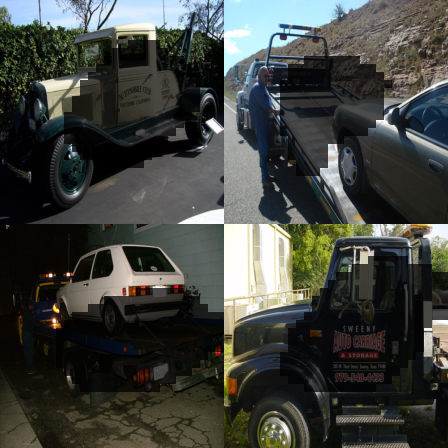}} 
            edge from parent[draw=none]} 
        child {node[xshift=-0.84cm](layer3_tow_truck_concept5) {\includegraphics[width=16mm]{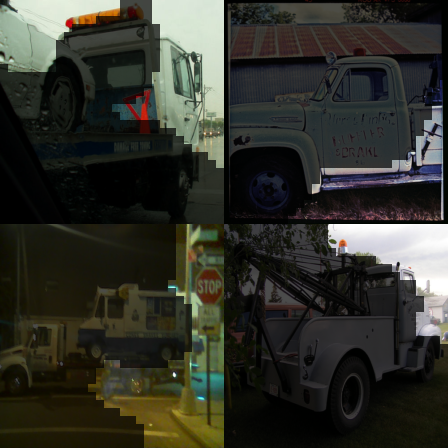}}
            edge from parent[draw=none]} 
            } 
    child {node[](layer4_tow_truck_concept4){\includegraphics[width=16mm]{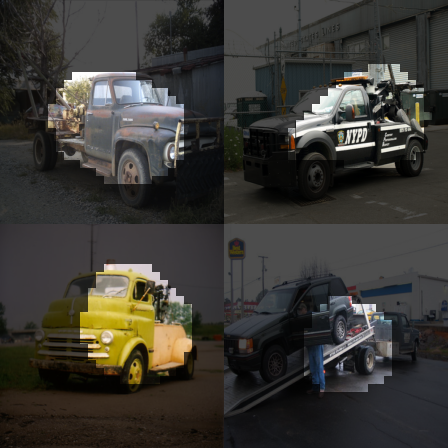}} edge from parent[draw=none]};
\begin{scope}[on background layer]
\draw[line width=0.6mm,-,/pgfplots/color of colormap=(295)] ($ (layer3_tow_truck_concept4.north) - (0, 0.2) $) -- ( $ (layer4_tow_truck_concept4.south) + (0, 0.25) $);
\draw[line width=0.6mm,-,/pgfplots/color of colormap=(491)] ($ (layer3_tow_truck_concept4.north) - (0, 0.2) $) -- ( $ (layer4_tow_truck_concept3.south) + (0, 0.25) $);
\draw[line width=0.6mm,-,/pgfplots/color of colormap=(267)] ($ (layer3_tow_truck_concept4.north) - (0, 0.2) $) -- ( $ (layer4_tow_truck_concept2.south) + (0, 0.25) $);
\draw[line width=0.6mm,-,/pgfplots/color of colormap=(241)] ($ (layer3_tow_truck_concept1.north) - (0, 0.2) $) -- ( $ (layer4_tow_truck_concept4.south) + (0, 0.25) $);
\draw[line width=0.6mm,-,/pgfplots/color of colormap=(628)] ($ (layer3_tow_truck_concept1.north) - (0, 0.2) $) -- ( $ (layer4_tow_truck_concept3.south) + (0, 0.25) $);
\draw[line width=0.6mm,-,/pgfplots/color of colormap=(373)] ($ (layer3_tow_truck_concept1.north) - (0, 0.2) $) -- ( $ (layer4_tow_truck_concept2.south) + (0, 0.25) $);
\draw[line width=0.6mm,-,/pgfplots/color of colormap=(710)] ($ (layer3_tow_truck_concept1.north) - (0, 0.2) $) -- ( $ (layer4_tow_truck_concept1.south) + (0, 0.25) $);
\draw[line width=0.6mm,-,/pgfplots/color of colormap=(420)] ($ (layer3_tow_truck_concept5.north) - (0, 0.2) $) -- ( $ (layer4_tow_truck_concept4.south) + (0, 0.25) $);
\draw[line width=0.6mm,-,/pgfplots/color of colormap=(546)] ($ (layer3_tow_truck_concept5.north) - (0, 0.2) $) -- ( $ (layer4_tow_truck_concept3.south) + (0, 0.25) $);
\draw[line width=0.6mm,-,/pgfplots/color of colormap=(626)] ($ (layer3_tow_truck_concept5.north) - (0, 0.2) $) -- ( $ (layer4_tow_truck_concept1.south) + (0, 0.25) $);
\draw[line width=0.6mm,-,/pgfplots/color of colormap=(0)] ($ (layer4_tow_truck_concept4.north) - (0, 0.2) $) -- ( $ (class_tow_truck.south) + (0, 0.25) $);
\draw[line width=0.6mm,-,/pgfplots/color of colormap=(0)] ($ (layer4_tow_truck_concept3.north) - (0, 0.2) $) -- ( $ (class_tow_truck.south) + (0, 0.25) $);
\draw[line width=0.6mm,-,/pgfplots/color of colormap=(0)] ($ (layer4_tow_truck_concept1.north) - (0, 0.2) $) -- ( $ (class_tow_truck.south) + (0, 0.25) $);
\draw[line width=0.6mm,-,/pgfplots/color of colormap=(0)] ($ (layer4_tow_truck_concept2.north) - (0, 0.2) $) -- ( $ (class_tow_truck.south) + (0, 0.25) $);
\end{scope}
    \end{tikzpicture}
\hspace{0.075cm}
\vrulesep
\hspace{0.075cm}
    \begin{tikzpicture}[
        every node/.style = {inner sep=0pt},
every label/.append style = {label distance=2pt, align = center},
         sibling distance = 4.75em,
           level 1/.style = {level distance=5em,anchor=north},
           level 2/.style = {level distance=4em,anchor=north},
           level 3/.style = {level distance=3em,anchor=north},
           ]  
\node[anchor=south, label=MViT](class_tow_truck)
{\includegraphics[width=25mm]{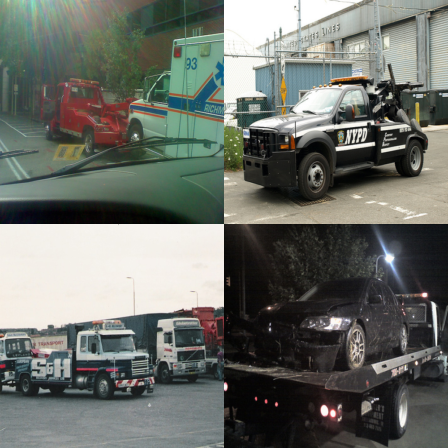}}
    child {node[](15_tow_truck_concept1){\includegraphics[width=16mm]{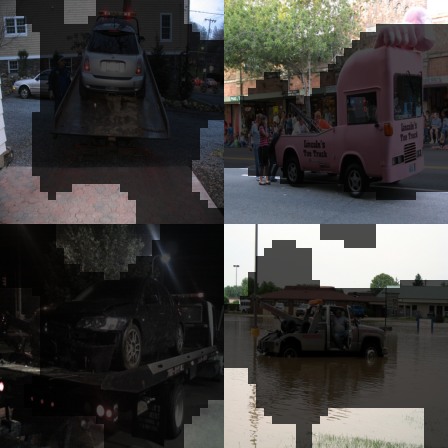}} 
        edge from parent[draw=none]
        child {node[xshift=0.84cm](9_tow_truck_concept3){\includegraphics[width=16mm]{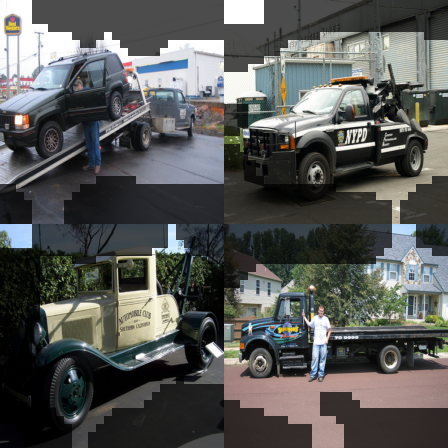}} 
            edge from parent[draw=none]}
        child {node[xshift=0.84cm](9_tow_truck_concept5){\includegraphics[width=16mm]{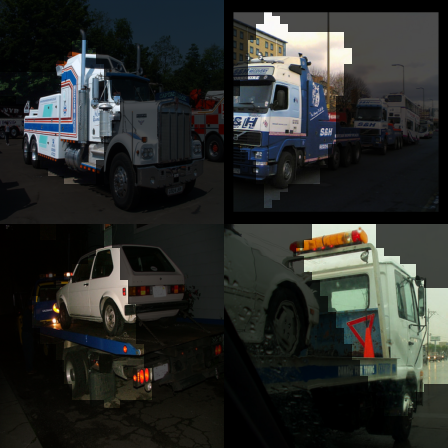}} 
            edge from parent[draw=none]} 
        child {node[xshift=0.84cm](9_tow_truck_concept2) {\includegraphics[width=16mm]{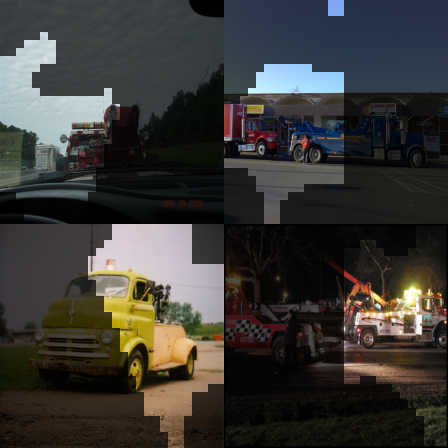}}
        edge from parent[draw=none]}
            } 
    child {node[](15_tow_truck_concept2){\includegraphics[width=16mm]{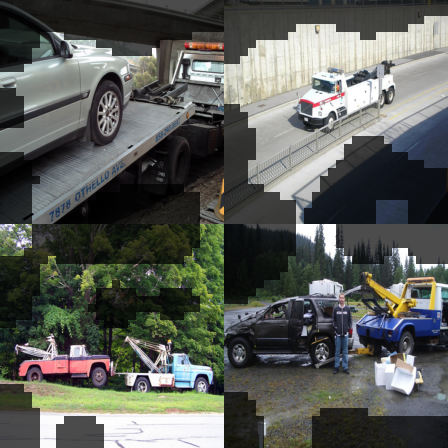}} 
    edge from parent[draw=none]};
\begin{scope}[on background layer]
\draw[line width=0.6mm,-,/pgfplots/color of colormap=(514)] ($ (9_tow_truck_concept2.north) - (0, 0.2) $) -- ( $ (15_tow_truck_concept1.south) + (0, 0.25) $);
\draw[line width=0.6mm,-,/pgfplots/color of colormap=(419)] ($ (9_tow_truck_concept5.north) - (0, 0.2) $) -- ( $ (15_tow_truck_concept2.south) + (0, 0.25) $);
\draw[line width=0.6mm,-,/pgfplots/color of colormap=(432)] ($ (9_tow_truck_concept3.north) - (0, 0.2) $) -- ( $ (15_tow_truck_concept2.south) + (0, 0.25) $);
\draw[line width=0.6mm,-,/pgfplots/color of colormap=(0)] ($ (15_tow_truck_concept2.north) - (0, 0.2) $) -- ( $ (class_tow_truck.south) + (0, 0.25) $);
\draw[line width=0.6mm,-,/pgfplots/color of colormap=(737)] ($ (15_tow_truck_concept1.north) - (0, 0.2) $) -- ( $ (class_tow_truck.south) + (0, 0.25) $);
\end{scope}
    \end{tikzpicture}
}\vspace{-0.34cm}
\caption{VCCs for a CNN (ResNet50) and transformer (MViT) for the class `Tow Truck'. Shown are only the last two layers with the class output.
}
\label{fig:cnn_trans_vcc_compare} \vspace{-0.65cm}
\end{figure}

An all layer VGG-16 VCC visualization was presented in Fig.~\ref{fig:intro_fig2}. As discussed in the caption, hierarchical concept assemblies again are revealed, with both target object as well as its background contributing to the final classification. While both the few layer and all layer visualizations reveal the concept representations of the models under analysis, they afford different levels of granularity: Few layer visualization provides a concise summary with a focus on specific layers, while all layer provides very detailed study.

\begin{figure*}
   \centering   
\includegraphics[width=0.96\textwidth]{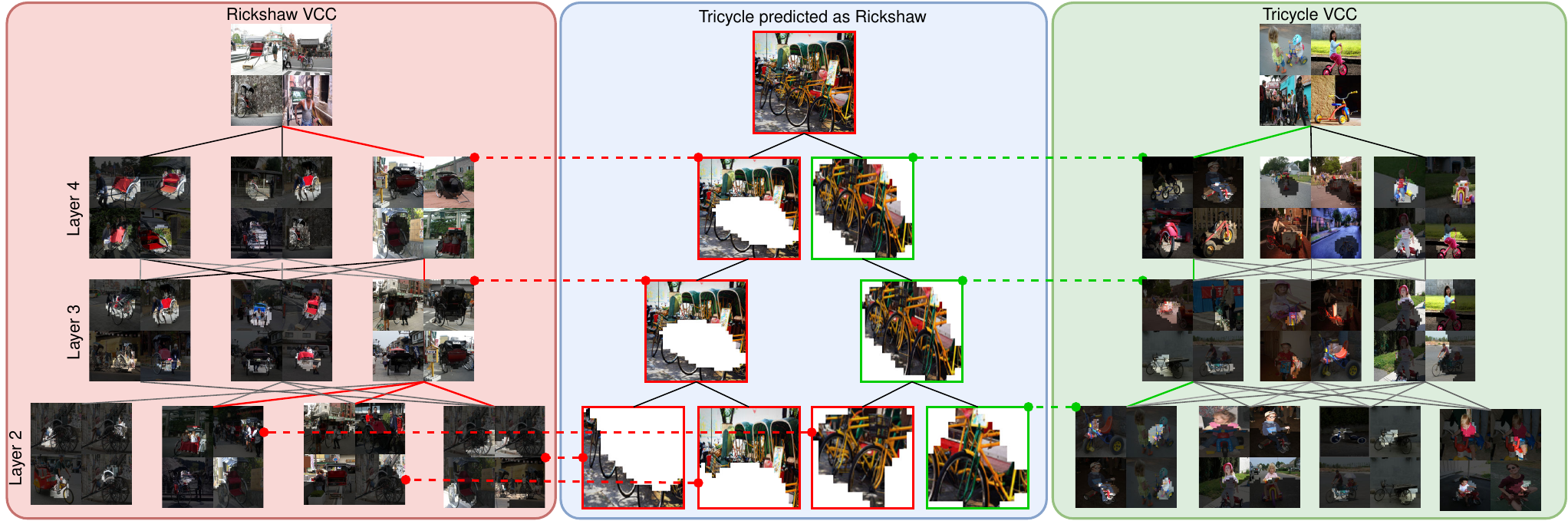}\vspace{-0.25cm}
    \caption{Debugging model failure modes with VCCs. We show an image of a tricycle incorrectly predicted by a ResNet50 as a rickshaw (middle) as well as the top-down segmentation of the image (Sec.~\ref{sec:divisive_cluster}). We also show the incorrect (left) and correct (right) VCCs. Following the hierarchy of concepts reveals that the model incorrectly focused heavily on the collapsible hood, starting at Layer 2.}\label{fig:debug}\vspace{-0.58cm}
\end{figure*}

\textbf{Subset VCC analysis.} We begin our quantitative analysis on four-layer VCCs generated for the CNNs and transformers listed in Sec.~\ref{sec:implementation_details}. Layers are selected approximately uniformly (see appendix for exact layers) across a model to capture concepts at different abstractions. We use standard metrics for analyzing tree-like data structures in a per-layer fashion: branching factor (extent to which the representation is distributed), number of nodes (how many concepts the model uses to encode a particular class), edge weights (strength of connection between concepts) and edge weight variance (variability of connection strength). We calculate averages over 50 VCCs for 50 randomly selected ImageNet classes. Results are shown in Fig.~\ref{fig:cnn_trans}.

Patterns are apparent in concept numbers and edge weights: more, but weaker, concepts in early layers vs. fewer, but stronger, concepts in later layers. These patterns reflect the shared low-level features (\eg colors, textures) across classes and more specific features near the end, which yield larger ITCAV values. Also, CNNs show a decreasing branching factor, while transformers maintain a consistent number until the final layer, where all models converge to about two concepts, typically of an ImageNet class's foreground-background structure. Transformers have higher final layer edge weight variance compared to CNNs, indicating their ability to better differentiate earlier concepts' importance in forming later concepts, potentially explaining their superior classification performance (\ie all information is not equally valuable). We also compare models (ResNet50 and VGG16) when trained with self-supervision~\cite{chen2020simple} (SimCLR) or for adversarial robustness, \ie on Stylized ImageNet~\cite{geirhos2018imagenet} (ADV). We observe that robust and self-supervised models have fewer low-level concepts and compositionality than the originals, likely as their training yields less reliance on texture (stylization perturbs texture) and color (SimCLR training jitters color).


To examine these patterns further, VCC visualizations for a CNN and transformer are shown in Fig.~\ref{fig:cnn_trans_vcc_compare}. 
Here, we limit to two (diverse) models and two later layers with class output for space.
The connection diversity in the last layer is indeed observed to be larger for the transformer vs.\ the CNN. Notably, over half of the concepts in the CNN capture background centric concepts, while the transformer has only a single background centric concept.

\textbf{All layer VCCs}. We present quantitative analyses on all layer VCCs for
three diverse models: (i) a standard CNN, VGG-16, (ii) an efficient model, MobileNetv3 and (iii) a transformer, MViT. Averages are taken over 10 VCCs for 10 random ImageNet classes. Results are shown in Fig.~\ref{fig:local_global}. 
Common trends appear in all models. Concept composition is non-linear across layers, with branching factor ranging from 5-15 and converging to approximately two near the last layers. The peak number of concepts, around 20, is consistently at 30-40\% of network depth and, as in the four-layer analysis, also converges to two in the final layer. 

Edge weights and variances are in accord with the main findings of the four-layer analysis, but also reveal other insights into the compositional nature of the models. At fine grained, all layer analysis, each model more readily displays unique edge weight characteristics: VGG16's average weights decrease in later layers, MobileNetv3's drop greatly  before the final layer and MViT maintains consistent values. Still, overall these results indicate that penultimate concepts differ between CNNs vs. transformers, as supported by our four-layer VCC analysis; see Fig.~\ref{fig:cnn_trans}.
Higher variances in initial layers suggest a diverse combination of concepts, whereas deeper layers indicate a more uniform composition. Transformers, however, show a variance increase in the final layer, indicating greater compositionality.

In summary, the four and all-layer VCC analyses consistently highlight key aspects of deep networks: (i) Early and mid-layers use more concepts compared to deeper layers. (ii) Concept interactions, \eg ITCAV values and variances, differ greatly between adjacent layers, as opposed to across multiple layers. (iii) CNNs and transformers compose concepts differently, especially at the final layers.

%% file: sec/5_application.tex
\vspace{-0.2cm}
\section{Application: Diagnosing failure predictions}
To show the VCC's practical utility, we consider application to model failure analysis. Contrasting failure analysis methods that do not explain how errors are made compositionally~\cite{selvaraju2017grad} or only offer explanations of single neuron heatmaps~\cite{cheng2023deeply},  VCCs provide insights on compositional concepts across layers and distributed representations.

Figure~\ref{fig:debug} shows a `Tricycle' incorrectly classified as a `Rickshaw' by a ResNet50 model, and the corresponding incorrect VCC (`Rickshaw', left) and correct VCC (`Tricycle', right). As the image is decomposed using our top-down segmentation (Sec.~\ref{sec:divisive_cluster}), it is revealed that the majority of pooled segments are closer, in terms of $l_2$ distance, to concepts in the Rickshaw VCC (red outlines) than the tricycle VCC (green outlines). While the model correctly encoded the wheel and handlebar portions of the images as tricycle concepts, the background and collapsible hood concepts are composed from layers two through four as rickshaw concepts, which may cause the error. We also note the lack of other tricycle-specific concepts (\eg children).



%% file: sec/6_conclusion.tex
\vspace{-0.15cm}
\section{Discussion and conclusion}
The Visual Concept Connectome (VCC) is a method for discovering human interpretable concepts and their interlayer connections in a fully unsupervised way. VCCs allow for the interpretation of deep models at multiple resolutions, revealing interesting properties via application to a range of networks. When comparing architecture types, CNNs were found to rely heavily on final layer concepts for prediction, while transformers were found to have more diverse connection strengths in the final layer and a more complex assembly of later concepts from earlier ones. 
We also applied VCCs to model debugging and showed how the hierarchical concept representation can indicate how mistakes arise. Given VCCs generality, 
our methodology should be applicable 
to additional networks, tasks and applications.

%% file: sec/X_suppl.tex
\clearpage
\setcounter{page}{1}
\maketitlesupplementary

\section{Introduction}

This document provides additional material that is supplemental to our main submission. Section~\ref{sec:algorithms} outlines the algorithms used in our technical approach. Section~\ref{sec:supp_implementation_details} describes additional implementation details for our approach, including further Visual Concept Connectome (VCC) generation settings, model details, clustering details and target classes chosen for evaluation. Section~\ref{sec:experiments} provides additional empirical results in terms of validation of the segment proposal method, validation of the concept discovery method, validation of the interlayer testing with concept activation vector (ITCAV) method, VCC visualizations comparing models, classes, and layers, as well as VCCs with a larger number of layers, including all layers. Section~\ref{sec:limitations} discusses the limitations of VCCs and our associated methodology to generate them. Section~\ref{sec:societal} discusses the societal implications, both positive and negative, of our method. Finally, Section~\ref{sec:assets} details the used assets and accompanying licenses.

\section{Algorithms}\label{sec:algorithms}


In this section, we present pseudocode for the three main algorithmic components of our method: (i) Top-down feature segmentation (Sec.~\ref{sec:divisive_cluster} in the main paper), (ii) Layer-wise concept discovery (Sec.~\ref{sec:concept_discovery_clustering} in the main paper) and (iii) Interlayer testing with concept activation vectors (ITCAV) (Sec.~\ref{sec:itcav} in the main paper). The top-down feature segmentation method is shown in Algorithm~\ref{alg:segment}. The layer-wise concept discovery method is shown in Algorithm~\ref{alg:concept_discovery}. Finally, The ITCAV method is shown in Algorithm~\ref{alg:itcav}. All references to equations in the algorithms refer to equations in the main paper.

\input{sec/X_algorithms}

\section{Implementation details}\label{sec:supp_implementation_details}

\begin{table*}[t]
    \centering
    \resizebox{0.75\textwidth}{!}{
        \begin{tabu}[t]{c | ccc | ccc | ccc | ccc}
        \tabucline[1pt]{-}
              & \multicolumn{3}{c}{\textbf{ResNet50}}  & \multicolumn{3}{c}{\textbf{VGG16}}  & \multicolumn{3}{c}{\textbf{MViT}}  & \multicolumn{3}{c}{\textbf{ViT-b}} \\
               &  RF & ACE & Ours & RF & ACE & Ours & RF & ACE & Ours & RF & ACE & Ours \\ 
               \hline
            Layer1 & 43 & 2.4 & 4.2    & 10 & 2.8& 3.5 & 224 & 1.7  & 7.5 & 224 & 1.3& 10.0 \\
            Layer2 & 99 & 2.0 & 11.9  & 32 & 2.4& 6.8  & 224 & 1.0 & 15.2 & 224 & 1.7& 24.4 \\
            Layer3 & 211 & 1.92 & 23.9& 80 & 2.2& 15.5 & 224 & 2.5 & 28.1 & 224 & 2.4& 35.8 \\
            Layer4 & 435 & 2.1 & 47.9 & 176& 2.2& 46.8 & 224 & 2.4 & 50.1 & 224 & 2.4& 54.1 \\
             \tabucline[1pt]{-}
        \end{tabu}
        }
    \caption{Validation of segment proposal component of our method (Sec.~\ref{sec:divisive_cluster}). The relative concept segment size compared to the entire image for a given layer, is shown with the receptive field (RF) width/height of the same layer. We compare our method (Ours) to the baseline method, ACE~\cite{ghorbani2019towards}. For all models, the relative segment size discovered using our method has a stronger correlation with the receptive field size than the concepts discovered using ACE.}
    \label{tab:segment_val}
\end{table*}

\textbf{VCC settings.} 50 target images are used to generate each VCC. The statistical testing and training of the CAVs~\cite{kim2018interpretability} use 20 unique sets of random images from the Broden dataset~\cite{bau2017network}. When computing the concept connection edge weight between the final selected layer and the class logit, the standard TCAV~\cite{kim2018interpretability} score is used. 

\textbf{Model settings.} For the CLIP-ResNet50~\cite{radford2021learning} experiments (Sec.~\ref{sec:tasks}), we follow the original paper~\cite{radford2021learning} and compute the logit for each ImageNet~\cite{deng2009imagenet} class using a single query sentence `\textit{a photo of a \{class\}}'; where `\textit{\{class\}}' is the target ImageNet class. The layer names used (according to the PyTorch~\cite{paszke2019pytorch} module nomenclature) for each model when generating all four-layer VCCs are as follows: 

\begin{enumerate}
    \item ResNet18~\cite{he2016deep},  ResNet50~\cite{he2016deep} and CLIP-ResNet50~\cite{radford2021learning}: \textit{Layer1, Layer2, Layer3, Layer4}
    \item VGG16~\cite{simonyan2014very}: \textit{8, 15, 22, 29}
    \item MobileNetv3~\cite{sandler2018mobilenetv2}: \textit{0, 2, 4, 6}
    \item MViT~\cite{fan2021multiscale}: \textit{1, 3, 9, 15}
    \item ViT-b~\cite{vaswani2017attention}: \textit{2, 5, 8, 10}
\end{enumerate}

The layer names used (according to the PyTorch~\cite{paszke2019pytorch} module nomenclature) for each model when generating the all-layer VCCs are as follows: 

\begin{enumerate}
    \item ResNet50~\cite{radford2021learning}: \textit{layer1.0, layer1.1, layer1.2, layer2.0, layer2.1, layer2.2, layer2.3, layer3.0, layer3.1, layer3.2, layer3.3, layer3.4, layer3.5, layer4.0, layer4.1, layer4.2}
    \item VGG16~\cite{simonyan2014very}: \textit{1, 3, 6, 8, 11, 13, 15, 18, 20, 22, 25, 27, 29}
    \item MobileNetv3~\cite{sandler2018mobilenetv2}: \textit{0.0, 1.0, 1.1, 2.0, 2.1, 2.2, 3.0, 3.1, 3.2, 3.3, 4.0, 4.1, 5.0, 5.1, 5.2, 6.0}
    \item MViT~\cite{fan2021multiscale}: \textit{0, 1, 2, 3, 4, 5, 6, 7, 8, 9, 10, 11, 12, 13, 14, 15}
    \item ViT-b~\cite{vaswani2017attention}: \textit{0, 1, 2, 3, 4, 5, 6, 7, 8, 9, 10}
\end{enumerate}


\textbf{Clustering details.} Following previous work~\cite{ghorbani2019towards}, during the concept discovery clustering, $\mathsf{C}^{con}$, we over-cluster and then prune to ensure that fewer concepts will be missed. We follow previous work~\cite{ghorbani2019towards} and choose the number of clusters to be $k_{m}=25$ in the concept discovery step. However, as VCCs target the discovery of concepts at potentially all layers, we select a different pruning protocol~\cite{ghorbani2019towards}, where they prune based on a single minimum value. Instead, we prune clusters that have less than $Y$ images, via the generalized logistic sigmoid
\begin{equation}
    Y = A + \frac{K - A}{(C + Qe^{-Bt})^{1/\nu}},
\end{equation}
where $A=-102$, $K=115$, $C=1$, $Q=1$, $B=0.0004$ and $v=1$. Pruning based on a sigmoid shaped function enables different levels of leniency when considering what constitutes a concept for each layer. This is crucial as different layers contain a different number of segments from the top-down segmentation algorithm (Sec.~\ref{sec:divisive_cluster} in the main paper).

For the maskSLIC~\cite{irving2016maskslic} clustering stage, we use a compactness of 0.8 and all other parameters are set to the Scikit-Image~\cite{van2014scikit} defaults. The Euclidean distance is used for all clustering steps.

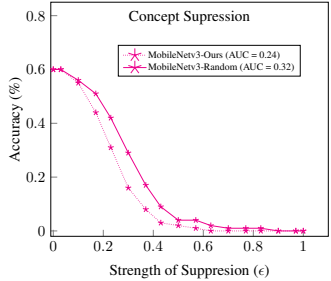
\begin{figure}
    \centering
    \resizebox{0.25\textwidth}{!}{
        \begin{tikzpicture}
	\begin{axis}[
            title=Concept Supression,
            title style={at={(axis description cs:0.5,0.95)},anchor=north}, 
		xlabel=Strength of Suppresion ($\epsilon$),
		ylabel=Accuracy (\%),
            height=7cm,
            width=8cm,
            xmin=-0.01, xmax=1.1,
            ymin=-0.02, ymax=0.85,
            x tick label style={{yshift=0.0cm}},
            y tick label style={{xshift=0.0cm}},
            x label style={at={(axis description cs:0.5,0)},anchor=north}, 
            y label style={at={(axis description cs:0.11,0.5)},anchor=south},
            xtick pos=left,
            ytick pos=left,
            legend style={mark size=5pt},
            legend style={
             nodes={scale=0.45, transform shape},
             cells={anchor=west},
             legend style={at={(0.9,0.75)},anchor=east}, font=\Large},
             legend image post style={scale=0.9},
             legend columns=1]

    	\addplot[color=magenta,densely dotted,mark=star,mark options={scale=1,solid}] coordinates {
		(0,0.60)(0.03,0.60)(0.10,0.55)(0.17,0.44)(0.23,0.31)(0.30,0.16)(0.37,0.08)(0.43,0.03)(0.5,0.02)(0.57,0.01)(0.63,0)(0.70,0.0)(0.77,0)(0.83,0)(0.90,0)(0.97,0)(1,0)
	}; 
        \addlegendentry{MobileNetv3-Ours (AUC = 0.24)}

        \addplot[color=magenta,mark=star,mark options={scale=1,solid}] coordinates {
		(0,0.60)(0.03,0.60)(0.10,0.56)(0.17,0.51)(0.23,0.42)(0.30,0.29)(0.37,0.17)(0.43,0.09)(0.5,0.04)(0.57,0.04)(0.63,0.02)(0.70,0.01)(0.77,0.01)(0.83,0.01)(0.90,0)(0.97,0)(1,0)
	}; 
        \addlegendentry{MobileNetv3-Random (AUC = 0.32)}
        
	\end{axis}
\end{tikzpicture}}
    \caption{Additional validation results for the concept discovery method (Sec.~\ref{sec:concept_discovery_clustering} in the main paper) for the MobileNetv3 model~\cite{howard2019searching}. For a set of 50 randomly selected ImageNet classes, we discover concepts in four layers of the model. During inference, one randomly selected concept at each layer is suppressed by a factor of $\epsilon$.}
    \label{fig:concept_validation}
\end{figure}

\textbf{Randomized testing.} When applying our ITCAV method to calculate the strength of connection between two concepts, we protect against the impact of spurious results by performing a statistical significance test on all ITCAV scores. More specifically, instead of simply calculating the ITCAV score with the target concept images, we calculate an additional 20 ITCAV scores using random images sampled from Broden~\cite{bau2017network}. We perform a two-sided \textit{t}-test of the ITCAV score based on the 20 random scores. We test whether the null hypothesis (\ie a ITCAV score of 0.5) can be rejected with a \textit{p}-value of $p>0.05$. All ITCAV scores shown in the main paper and appendix pass this statistical test, \ie $p\leq0.05$.

\textbf{ImageNet classes.} The 50 ImageNet classes used for the model and task analysis experiments (Sec.~\ref{sec:architectures} in the main paper) are the following: \textit{tow truck,
sturgeon,
sax,
wool,
basketball,
whiptail,
toy poodle,
acorn,
crutch,
church,
backpack,
spaghetti squash,
snowmobile,
teapot,
ant,
chain,
gorilla,
holster,
wreck,
ice lolly
schipperke,
cradle,
dowitcher,
leopard,
oystercatcher,
saltshaker,
drake,
loupe,
spotlight,
Newfoundland,
bagel,
electric fan,
ping-pong ball,
streetcar,
knot,
plate,
sea lion,
leafhopper,
tusker,
punching bag,
black widow,
traffic light,
tricycle,
paper towel,
guinea pig,
castle,
go-kart,
platypus,
badger and
bicycle-built-for-two}.

The 10 ImageNet classes used for the all-layer VCC analysis experiments (Sec.~\ref{sec:architectures} in the main paper) are the following: \textit{tow truck,
sturgeon,
sax,
wool,
basketball,
whiptail,
toy poodle,
acorn,
crutch,
church}

\section{Additional empirical results}\label{sec:experiments}

\subsection{VCC component validation}

\subsubsection{Segment proposal validation}
Table~\ref{tab:segment_val} presents additional results to validate our top-down feature segmentation approach (Sec.~\ref{sec:divisive_cluster}). In particular, we show results for four additional models. We observe findings consistent with those in the main paper: Our method produces concepts that increase in size as the information flows deeper through the model. It is interesting to observe a similar phenomenon in transformer-based architectures, \ie MViT~\cite{fan2021multiscale} and ViT-b~\cite{vaswani2017attention}. While the relative concept size of the baseline, ACE~\cite{ghorbani2019towards}, varies less than 2\% across all architectures and layers, the size of concepts produced by our method can differ up to 20\% between architectures (\ie comparing VGG16-Ours with ViT-b-Ours) and 40\% between layers. This finding is to be expected as it is unlikely for all architectures at all layers to capture concepts of the same size.

\subsubsection{Concept validation}
Figure~\ref{fig:concept_validation} presents additional results to validate our layer-wise concept discovery method (Sec.~\ref{sec:concept_discovery_clustering} in the main paper) for the MobileNetv3~\cite{howard2019searching} model. The results are consistent with those in the main paper, \ie the accuracy for the target class decreases faster when a concept is suppressed compared to a randomly chosen direction. This result implies that the concepts discovered throughout the model represent meaningful directions in the latent space.

\subsubsection{Interlayer concept weight validation}

\begin{figure*}
    \centering
    \resizebox{0.99\textwidth}{!}{
\includegraphics[width=.25\textwidth]{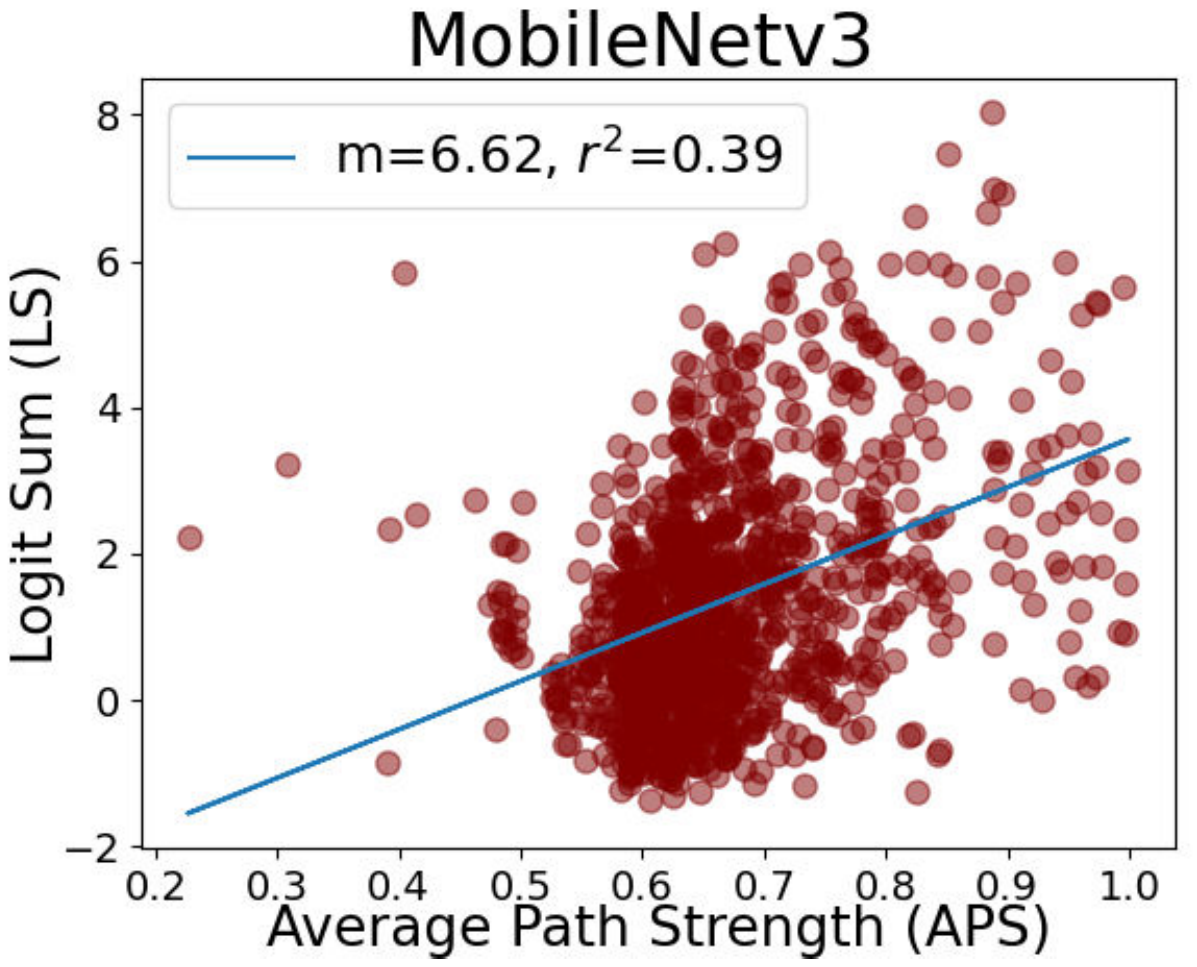}
\includegraphics[width=.24\textwidth]{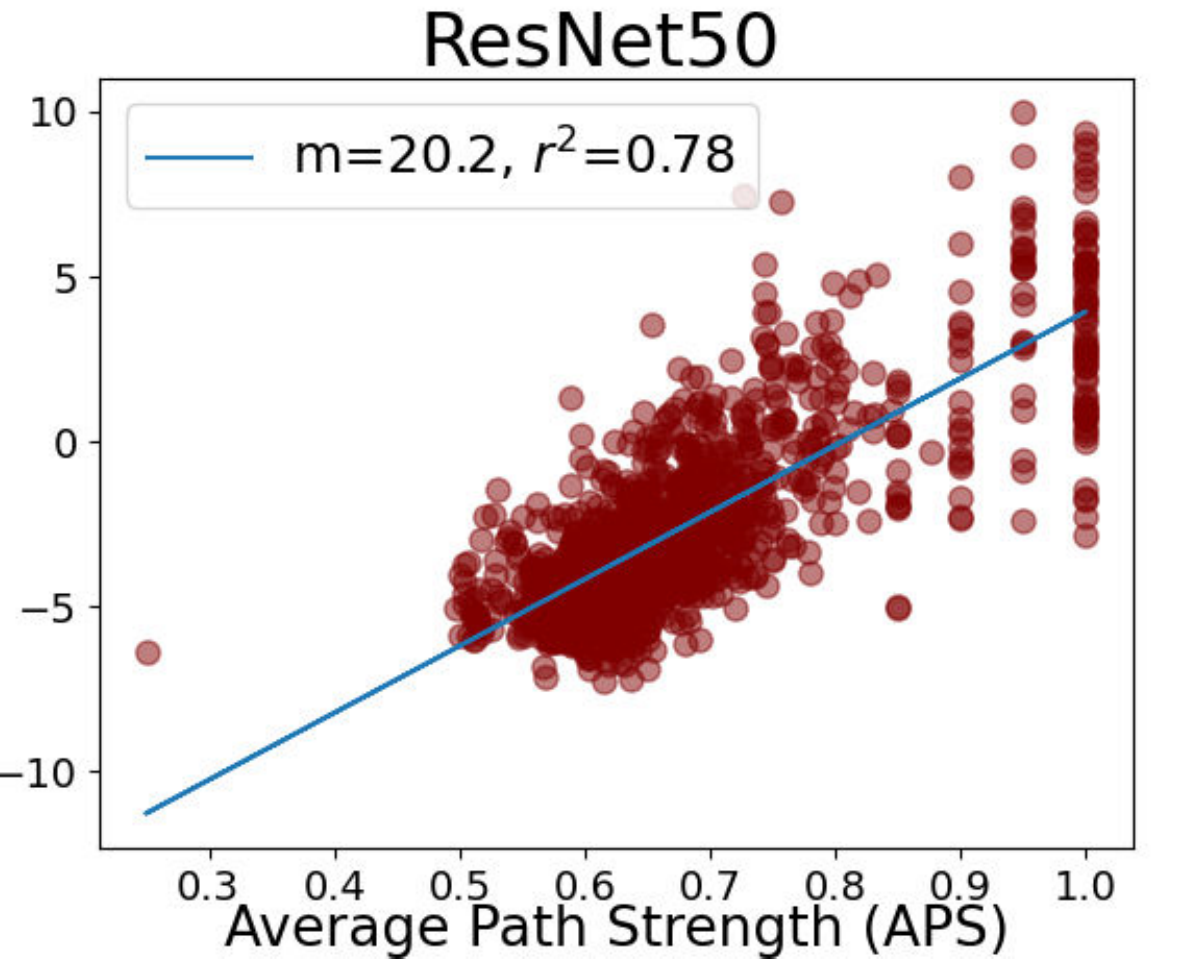}
\includegraphics[width=.23\textwidth]{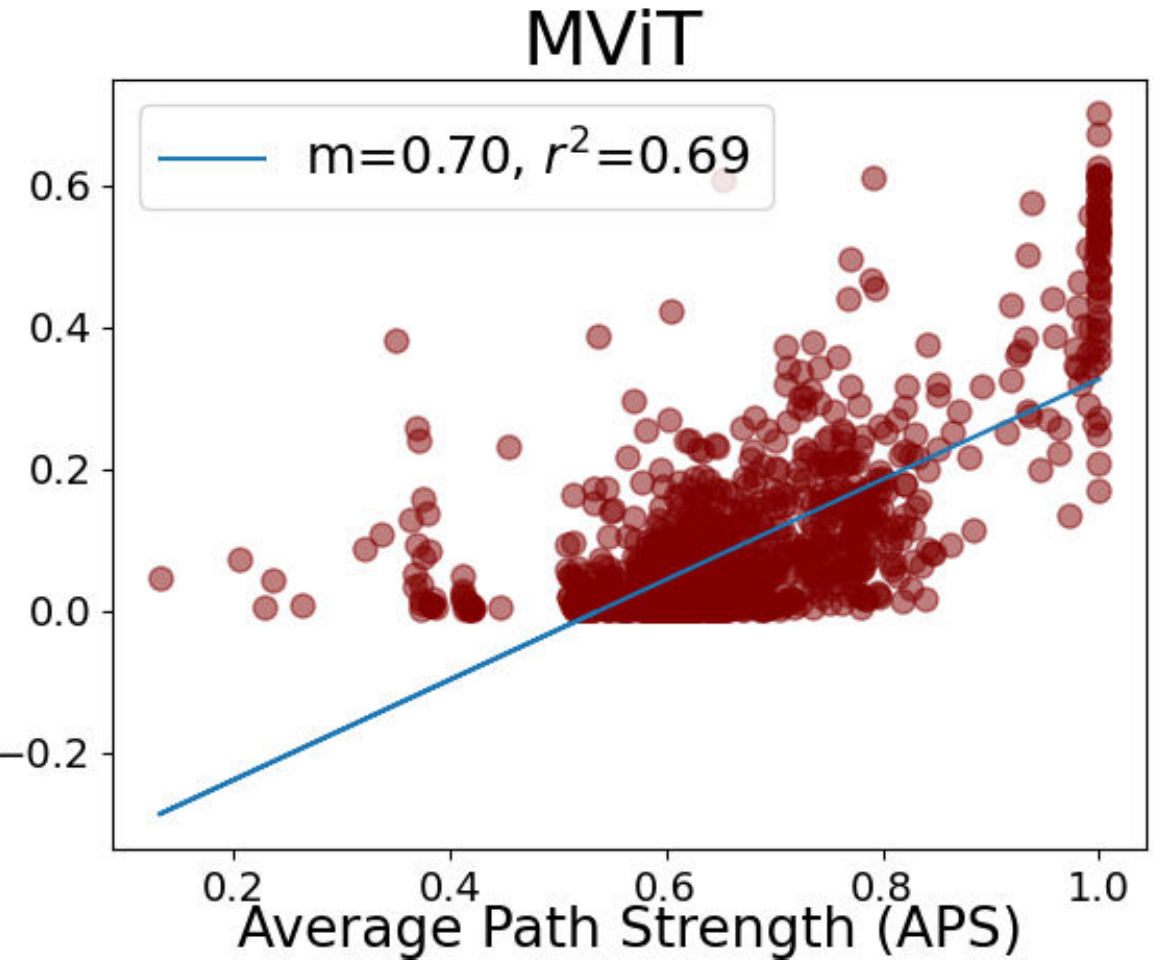}
}
\caption{Additional validation results of interlayer concept weights. The unnormalized logit sum (LS) scores, main paper (Eq.~\ref{eq:LS}), for the target class are plotted against the average path strength (APS) scores, main paper (Eq.~\ref{eq:APS}). A positive correlation implies that the ITCAV edge weights connecting a concept to the class are predictive of the model output having a higher probability for that class.}
\label{fig:ITCAV_validation} \vspace{-0.4cm}
\end{figure*}

We now extend the validation of our Interlayer Testing with Concept Activation Vectors (ITCAV) method (Sec.~\ref{sec:itcav} in the main paper). In particular, we show results for four additional models in Fig.~\ref{fig:ITCAV_validation} and observe findings consistent with those in the main paper: There is a positive correlation between the average path strength (APS) and the logit sum (LS) score. These results further suggest that the combination of ITCAV scores is predictive
of whether a concept is representative of the target class.



\subsection{Understanding models}\label{sec:supp_models}
\begin{figure} [t]
	\begin{center}
     \centering 
\resizebox{0.48\textwidth}{!}{
\hspace{0.1cm}
\begin{tikzpicture} 
                 \begin{axis}[
                 line width=1.0,
                 title style={at={(axis description cs:0.5,1.2)},anchor=north,font=\normalsize},
                 ylabel={Branching Factor},
                 xmin=0, xmax=1,
                 ymin=0, ymax=17,
                 xtick={0,0.2,0.4,0.6,0.8,1},
                 x tick label style={font=\tiny, rotate=0, anchor=north},
                 y tick label style={font=\tiny},
                 x label style={at={(axis description cs:0.5,0.09)},anchor=north,font=\small},
                 y label style={at={(axis description cs:0.2,.5)},anchor=south,font=\small},
                 width=6.5cm,
                 height=4.0cm, 
                 ymajorgrids=false,
                 xmajorgrids=false,
                 major grid style={dotted,green!20!black},
             ]

            \addplot[line width=0.9pt, mark size=1.8pt, color=blue, mark=diamond,error bars/.cd, y dir=both, y explicit,]
                     coordinates {
(0.06,	9.17)
(0.13,	12.72)
(0.19,	11.72)
(0.25,	12.52)
(0.31,	12.69)
(0.38,	11.94)
(0.44,	8.70)
(0.50,	4.87)
(0.56,	4.46)
(0.63,	4.59)
(0.69,	4.98)
(0.75,	6.23)
(0.81,	5.00)
(0.88,	5.17)
(0.94,	3.33)
(1.00,	1.89)};     

            \addplot[line width=0.9pt, mark size=1.8pt, color=red, mark=o,error bars/.cd, y dir=both, y explicit,]
                     coordinates {
(0.09,	13.50)
(0.18,	13.46)
(0.27,	3.83)
(0.36,	4.71)
(0.45,	4.20)
(0.55,	3.60)
(0.64,	4.22)
(0.73,	3.93)
(0.82,	2.72)
(0.91,	3.03)
(1.00,	1.82)
};   
              \end{axis}
\end{tikzpicture}
\hspace{0.1cm}
\begin{tikzpicture} 
                  \begin{axis}[
                 line width=1.0,
                 title style={at={(axis description cs:0.5,1.2)},anchor=north,font=\normalsize},
                 ylabel={Number of Concepts},
                 xmin=0, xmax=1,
                 ymin=1.5, ymax=21,
                 xtick={0,0.2,0.4,0.6,0.8,1},
                 x tick label style={font=\tiny, rotate=0, anchor=north},
                 y tick label style={font=\tiny},
                 x label style={at={(axis description cs:0.5,0.09)},anchor=north,font=\small},
                 y label style={at={(axis description cs:0.2,.5)},anchor=south,font=\small},
                 width=6.5cm,
                 height=4.0cm, 
                 ymajorgrids=false,
                 xmajorgrids=false,
                 major grid style={dotted,green!20!black},
                 legend columns=2,
                 legend style={
                  nodes={scale=0.7, transform shape},
                  cells={anchor=west},
                  legend style={at={(0.5,1)},anchor=north,row sep=0.01pt}, font =\small}
             ]
            \addplot[line width=0.9pt, mark size=1.8pt, color=blue, mark=diamond,error bars/.cd, y dir=both, y explicit,]
                     coordinates {
(0.06,	16.89)
(0.13,	18.11)
(0.19,	17.44)
(0.25,	18.67)
(0.31,	17.67)
(0.38,	15.44)
(0.44,	11.67)
(0.50,	6.11)
(0.56,	4.78)
(0.63,	5.89)
(0.69,	6.22)
(0.75,	6.67)
(0.81,	5.00)
(0.88,	5.78)
(0.94,	5.11)
(1.00,	1.89)    
                     };
 
            \addplot[line width=0.9pt, mark size=1.8pt, color=red, mark=o,error bars/.cd, y dir=both, y explicit,]
                     coordinates {
(0.09,	18.73)
(0.18,	17.73)
(0.27,	13.00)
(0.36,	8.09)
(0.45,	7.46)
(0.55,	8.82)
(0.64,	10.36)
(0.73,	9.64)
(0.82,	7.36)
(0.91,	5.82)
(1.00,	2.73)
};   
              \end{axis}
\end{tikzpicture}}
\resizebox{0.48\textwidth}{!}{
\hspace{0.1cm}
\begin{tikzpicture}
                 \begin{axis}[
                 line width=1.0,
                 title style={at={(axis description cs:0.5,1.2)},anchor=north,font=\normalsize},
                 xlabel={Layer},
                 ylabel={Edge Weight Average},
                 xmin=0, xmax=1,
                 ymin=0, ymax=1.2,
                 xtick={0,0.2,0.4,0.6,0.8,1},
                 x tick label style={font=\tiny, rotate=0, anchor=north},
                 y tick label style={font=\tiny},
                 x label style={at={(axis description cs:0.5,0.15)},anchor=north,font=\small},
                 y label style={at={(axis description cs:0.2,.5)},anchor=south,font=\small},
                 width=6.5cm,
                 height=4.0cm, 
                 ymajorgrids=false,
                 xmajorgrids=false,
                 major grid style={dotted,green!20!black},
                 legend columns=2,
                 legend style={
                  nodes={scale=0.6, transform shape},
                  cells={anchor=west},
                  legend style={at={(0.5,1)},anchor=north,row sep=0.01pt}, font =\small}
             ]
            \addplot[line width=0.9pt, mark size=1.8pt, color=blue, mark=diamond,error bars/.cd, y dir=both, y explicit,]
                     coordinates {
(0.06,	0.43)
(0.13,	0.50)
(0.19,	0.45)
(0.25,	0.44)
(0.31,	0.46)
(0.38,	0.39)
(0.44,	0.33)
(0.50,	0.39)
(0.56,	0.40)
(0.63,	0.41)
(0.69,	0.40)
(0.75,	0.38)
(0.81,	0.13)
(0.88,	0.80)
(0.94,	0.66)
(1.00,	0.86)     
                                };

            \addplot[line width=0.9pt, mark size=1.8pt, color=red, mark=o,error bars/.cd, y dir=both, y explicit,]
                     coordinates {
(0.09,	0.53)
(0.18,	0.31)
(0.27,	0.46)
(0.36,	0.35)
(0.45,	0.73)
(0.55,	0.42)
(0.64,	0.55)
(0.73,	0.35)
(0.82,	0.53)
(0.91,	0.64)
(1.00,	0.69)
};   
              \end{axis}
\end{tikzpicture}
\hspace{0.1cm}
\begin{tikzpicture} 
                  \begin{axis}[
                 line width=1.0,
                 title style={at={(axis description cs:0.5,1.2)},anchor=north,font=\normalsize},
                 xlabel={Layer},
                 ylabel={Edge Weight Variance},
                 xmin=0, xmax=1,
                 ymin=0, ymax=0.062,
                 xtick={0,0.2,0.4,0.6,0.8,1},
                 x tick label style={font=\tiny, rotate=0, anchor=north},
                 y tick label style={font=\tiny},
                 x label style={at={(axis description cs:0.5,0.15)},anchor=north,font=\small},
                 y label style={at={(axis description cs:0.2,.5)},anchor=south,font=\small},
                 width=6.5cm,
                 height=4.0cm, 
                 ymajorgrids=false,
                 xmajorgrids=false,
                 major grid style={dotted,green!20!black},
                    legend columns=1,
                 legend style={
                  nodes={scale=0.85, transform shape},
                  cells={anchor=west},
                  legend style={at={(1,1)},anchor=north east,row sep=0.01pt}, font =\small}
             ]
            \addlegendentry{ResNet50}
            \addplot[line width=0.9pt, mark size=1.8pt, color=blue, mark=diamond,error bars/.cd, y dir=both, y explicit,]
                     coordinates {
(0.063,	0.029)
(0.125,	0.039)
(0.188,	0.037)
(0.250,	0.028)
(0.313,	0.042)
(0.375,	0.022)
(0.438,	0.011)
(0.500,	0.008)
(0.563,	0.003)
(0.625,	0.002)
(0.688,	0.001)
(0.750,	0.002)
(0.813,	0.004)
(0.875,	0.006)
(0.938,	0.006)
(1.000,	0.006)
}; 
            \addlegendentry{ViTb}
            \addplot[line width=0.9pt, mark size=1.8pt, color=red, mark=o,error bars/.cd, y dir=both, y explicit,]
                     coordinates {
(0.09,	0.14)
(0.18,	0.06)
(0.27,	0.03)
(0.36,	0.01)
(0.45,	0.03)
(0.55,	0.02)
(0.64,	0.01)
(0.73,	0.02)
(0.82,	0.03)
(0.91,	0.02)
(1.00,	0.03)
}; 
                             
              \end{axis}
\end{tikzpicture}
}
\end{center}
\vspace{-0.6cm}
\caption{Graph metrics of all layer VCCs 
for two architectures.
Layer number normalized to allow for comparison of models with different numbers of layers.
}\label{fig:supp_all_layer}
\vspace{-0.5cm}
\end{figure}
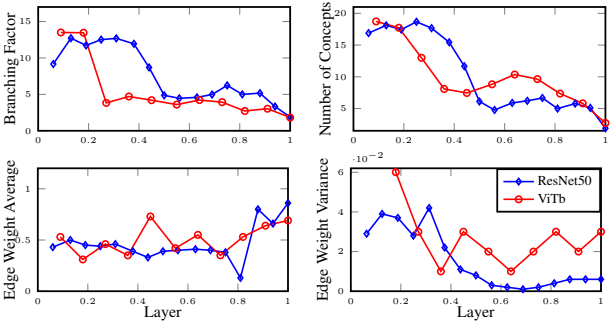

Figure~\ref{fig:supp_all_layer} extends the results from Fig.~\ref{fig:local_global} in the main paper and shows a quantitative analysis for all-layer VCCs on two additional models: ResNet50~\cite{he2016deep} and ViT-b~\cite{dosovitskiy2020image}. Consistent with the results from the main paper, we again see that the branching pattern and number of concepts start at a higher value and converges, suggesting that many concepts are shared between classes at early layers while the later layers capture ImageNet's foreground-background structure. We also observe patterns in the ITCAV values and variances that are consistent with the main paper. The edge weight values are consistent until the final layer at which point they increase, denoting the stronger contribution of the final layers to the output. In terms of the ITCAV variance, we again see that transformers (ViT-b) have a higher variance than CNNs (ResNet50) in the last layer, further suggesting that transformers have greater compositionality of concepts before the final prediction.

\subsection{Understanding tasks
}\label{sec:tasks}

\begin{table}[t]
    \centering
        \resizebox{0.48\textwidth}{!}{
        \begin{tabu}{c cccccc}
        \tabucline[1pt]{-}
             & \multicolumn{2}{c}{\textbf{Branching Factor}}  & \multicolumn{2}{c}{\textbf{Number of Concepts}} & \multicolumn{2}{c}{\textbf{Edge Weight Ave.}}\\
             \hline
              & R50 & CLIP & R50 & CLIP & R50 &  CLIP \\
             \tabucline[1pt]{-}
             Layer1 & 5.484 & 6.824 & 10.447 & 11.085 & 0.414 & 0.417\\ 
             Layer2 & 4.945 & 4.141 & 8.000  & 7.468   & 0.554 & 0.54 \\ 
             Layer3 & 2.799 & 2.754 & 5.106  & 5.702  & 0.476 & 0.563\\ 
             Layer4 & 2.915 & 1.574 & 2.957  & 2.383  & 0.917 & 0.634\\ 
        \tabucline[1pt]{-}
        \end{tabu}}
\vspace{-8pt}
    \caption{VCC metrics for ResNet50~\cite{he2016deep} trained on ImageNet~\cite{deng2009imagenet} and via contrastive image-language pretraining (CLIP)~\cite{radford2021learning}.}
    \label{tab:clip_r50_compare}
    \vspace{-12pt}
\end{table}

We now explore how VCCs can reveal the effect of the training task on learned concepts and their connections. In particular, given the recent advances of image-language training paradigms, we  compare the standard ResNet50~\cite{he2016deep} model trained on ImageNet~\cite{deng2009imagenet} with ResNet50 trained via Contrastive Language Image Pretraining (CLIP)~\cite{radford2021learning}.


Table~\ref{tab:clip_r50_compare} compares graph metrics over VCC layers between the two models at four residual blocks. 
We observe small but notable differences between the two models. First, CLIP contains a higher branching factor and number of concepts in the first layer than ResNet50, suggesting slightly more concepts are discovered and composed at the beginning of the network. The pattern is reversed at the end of the models, where CLIP has a slightly lower number of concepts and branching factor than ResNet50. When considering average edge weight values, we also observe a general consistency across models apart from the final layer, where ResNet50 has a much larger average value. This may be due to ImageNet trained CNNs having less compositionality at the end of the model as we observed both object and background classes having a large impact on the output in the main paper (Sec.~\ref{sec:architectures}).



\subsection{Additional VCCs visualizations}

\subsubsection{Four layer VCCs}\label{sec:4_lay_vccs}

We now supplement the analysis from Sec.~\ref{sec:architectures} from the main paper by generating VCCs for the entirety of the five models analyzed for different classes in the four layer setting. We specifically chose these models and layer settings as they are the same as in Sec.~\ref{sec:architectures} in the main paper. The models shown are ResNet50~\cite{he2016deep} (Fig.\ \ref{fig:vcc_r50}), VGG16~\cite{simonyan2014very} (Fig.\ \ref{fig:vcc_vgg16}), MobileNetv3~\cite{howard2019searching} (Fig.\ \ref{fig:vcc_mobilenetv3}), MViT~\cite{fan2021multiscale} (Fig.\ \ref{fig:vcc_mvit}) and ViT-b~\cite{vaswani2017attention} (Fig.\ \ref{fig:vcc_vitb}). All models are  trained on ImageNet~\cite{deng2009imagenet}. The layers selected are the same ones as detailed in Sec.~\ref{sec:supp_implementation_details}.

We observe differences in mid and late layer connection strengths between CNNs and transformers. Similar to the main paper discussion (Sec.~\ref{sec:architectures}), CNNs (Figs.~\ref{fig:vcc_r50},~\ref{fig:vcc_vgg16} and ~\ref{fig:vcc_mobilenetv3}) show stronger connections with less variance between the $4^{\text{th}}$ layer and class logit than the transformers (Figs.\ \ref{fig:vcc_mvit} and~\ref{fig:vcc_vitb}). Additionally, CNNs tend toward concepts which capture either the entire foreground or background in later layers. 
Meanwhile, the transformers produce concept shapes of varying shapes and sizes, \eg the VCC for ViT-b in Fig.~\ref{fig:vcc_vitb} contains concepts of both small patches and the entire images in the final VCC layer and concepts of varying sizes in the first VCC layer. These findings for the transformers are consistent with the ability of such models to form data associations across their input without the locality constraints that are inherent in convolutional models.

\textbf{Finegrained dataset VCC.} 
To show how VCCs generalize to other datasets, we generate a VCC for the CUB~\cite{WahCUB_200_2011} finegrained classification dataset, where the goal is to classify different types of birds. Figure~\ref{fig:vcc_r50_cub} shows a four layer VCC for the ResNet18~\cite{he2016deep} model targeting the class ``indigo bunting''. We again see interesting concepts being composed. For example, branches and the color blue occur in stage1 and stage2, while stage4 bird concepts are composed from branch, background and bird head concepts in stage3.

\begin{figure*}
   \centering   
\includegraphics[width=0.95\textwidth]{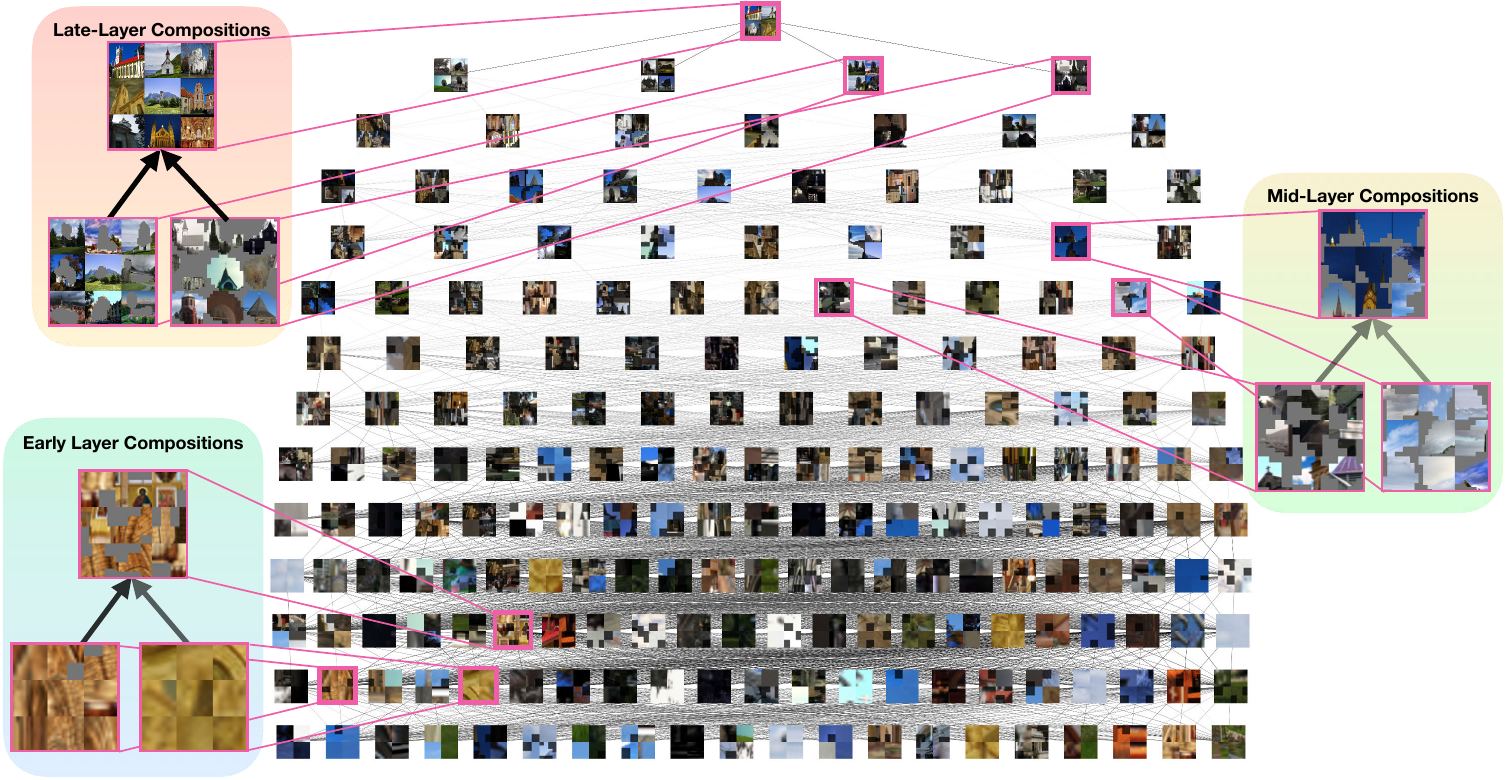}\vspace{-0.3cm}
    \captionof{figure}{An all-layer VCC of the VGG16 network targetting the class ``church''. Darker lines denote larger concept contributions. 
    }\label{fig:all_lay_vcc2}\vspace{-0.3cm}
\end{figure*}

\textbf{All layer VCC.} 
We show an additional all-layer VCC in Fig.~\ref{fig:all_lay_vcc2} of the VGG16 model~\cite{szegedy2015going} targeting recognition of class ``church". As in the visualization in the main paper, we visualize VCC subgraphs and observe interesting compositions occurring at different levels of abstraction corresponding at different depths of the model. At early layers (bottom left), we observe oriented brown patterns and yellow color  composing the concept of brown and yellow orientation. Middle layers (right) show the concept of `church roof with sky in the background' being composed of `church roof' and `sky'. The final layer concepts (top left) show that both foreground objects, \eg churches, and background regions, \eg trees or sky, concepts highly influence the final category.

\section{Application: Diagnosing failure predictions}
\begin{figure*}
   \centering   
\includegraphics[width=0.96\textwidth]{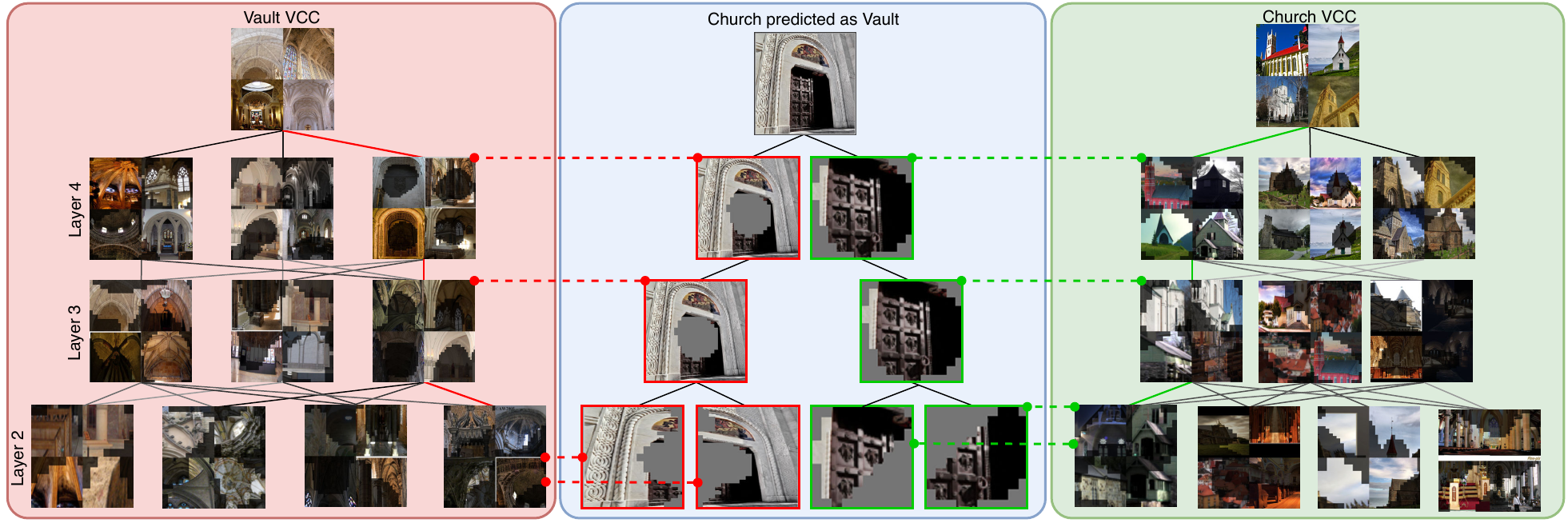}\vspace{-0.3cm}
    \caption{Debugging model failure modes with VCCs. We show an image of a church incorrectly predicted by a ResNet50 as a vault (middle) as well as the top-down segmentation of the image (Sec.~\ref{sec:divisive_cluster}). We also show the incorrect (left) and correct (right) VCCs. Following the hierarchy of concepts reveals that the model incorrectly focused heavily on the cement door frame, starting at Layer 2.}\label{fig:debug2}\vspace{-0.5cm}
\end{figure*}

To further show our VCC's practical utility, we show another example of model debugging. Figure~\ref{fig:debug2} shows a `church' incorrectly predicted as a `vault' by a ResNet50 model, and the corresponding incorrect VCC (`vault', left) and correct VCC (`church', right). As the image is decomposed using our top-down segmentation (Sec.~\ref{sec:divisive_cluster}), it is revealed that several segments are closer, in terms of the $l_2$ distance of the pooled segment activations, to concepts in the `vault' VCC (red outlines) than the `church' VCC (green outlines). While the model correctly encoded the door as a `church' concept, the regions outside the door are identified as `vault' concepts from layers two through four, which may cause the error. We also note the lack of other `church' specific concepts, such as the sky or cylindrical columns.

\section{Limitations}\label{sec:limitations}

We note some limitations of our method. We rely on the Silhouette method~\cite{rousseeuw1987silhouettes} to select the number of clusters (\ie segments) during the top-down feature segmentation stage to automate this step. However, use of a different method for selecting the number of clusters could yield different results and therefore different overall VCCs. In practice, we have found that using the Silhouette method consistently produces meaningful segments; so, sensitivity to this choice is not a serious limitation. 
Another limitation arises is that we do not provide a method for selecting the set of layers to analyze. Such a method for automatic layer selection could reveal further interesting and useful patterns, such as uncovering the set of layers, along with their connections, which impact the model output most significantly. A direction to realize such an algorithm could be to construct a large VCC and subsequently trim the least important nodes and edges (\eg based on the average path strength (APS) to the logit, as defined in Eq.~\ref{eq:APS} in the main paper).

\section{Societal implications}\label{sec:societal}

Understanding the decision making processes of deep networks is an important and open problem in computer vision. Given their potential for negative impacts when deployed, various jurisdictions are moving forward with legislation that may curtail certain applications and mandate interpretable components in deployed systems~\cite{euro2021laying}. VCCs are a step towards a holistic understanding of how concepts in deep networks are learned and in the future may provide a direction to design legally recognized interpretations of these models. 

VCCs may have implications in terms of recognizing both \textit{what} and \textit{how} biases are learned by deep networks. While the learning of various biases by deep networks is well documented~\cite{mehrabi2021survey}, it is not well understood \textit{how} these biases are constructed and learned by the model. For example, it is not sufficient to explain a model's prediction by saying it uses the background as a feature. It would be more desirable to explain what concepts are composed in earlier layers that lead the model to encode the background feature in the later layers, which we have shown that VCCs can reveal. Moreover, such information could open up new directions for model debiasing.

In terms of negative consequences, VCCs (and explainable AI in general) may give users a false sense of security and allow them to deploy models that ultimately do more harm than good. Furthermore, the contribution of additional explainable AI methods may contribute to the disagreement problem~\cite{krishna2022disagreement}, \ie when multiple explanations of a given model disagree with each other. It is an open research question on how to resolve such disagreements, when potentially dozens of possible explanations for a given model exist. 


\section{Assets and licensing}\label{sec:assets}

\noindent \textbf{Models.} We use provided code and trained weights from the MViT\footnote{\url{https://github.com/facebookresearch/mvit}} and CLIP\footnote{\url{https://github.com/openai/CLIP}} repositories. MViT is licensed under the Apache 2.0 license\footnote{\url{https://github.com/facebookresearch/mvit/blob/main/LICENSE}} and CLIP is licensed under the MIT license\footnote{\url{https://github.com/openai/CLIP/blob/main/LICENSE}}.

\noindent \textbf{Datasets.} We use the ImageNet dataset\footnote{\url{https://www.image-net.org/}} which is under the BSD 3-Clause License\footnote{\url{https://github.com/floydhub/imagenet/blob/master/LICENSE}} and the Broden dataset\footnote{\url{https://github.com/CSAILVision/NetDissect-Lite}} which is under the MIT license\footnote{\url{https://github.com/davidbau/quick-netdissect/blob/master/LICENSE}}.

\input{VCCs/r50_ant}
\input{VCCs/vgg16_curch}
\input{VCCs/mobilenetv3_guineapig}
\input{VCCs/mvit_crutch}
\input{VCCs/vitb_snowmobile}
\input{VCCs/r18_indigobunting}

\clearpage
\newpage











%% file: sec/X_algorithms.tex
\begin{algorithm}[t]
\caption{\textbf{Top-Down Feature Segmentation} }\label{alg:segment}
		\begin{algorithmic}[1]
	   {\fontsize{9pt}{9pt}\selectfont
            \Statex \textbf{Input}: Model $F$, Set of images $\bm{\mathcal{I}}$, $n$ selected layers of $F$ to study, Spatial clustering algorithm $\mathsf{C}^{seg}$
            \Statex \hspace{0cm}\textcolor{cyan}{/*\textit{$\mathsf{C}^{seg}$ is instatiated in terms of maskSLIC~\cite{irving2016maskslic}}*/}
            \Statex \textbf{Output}: Set of RGB image masks $\textbf{M}$

            \Statex \hspace{0cm}\textcolor{cyan}{/*\textit{Set lists for collection of masks and activations}*/}

            \State $\textbf{M}$ $\leftarrow$ $[]$,
    
            \For{ $j$ $\in$ $n$}
            
                    \State $\textbf{M}_j$, $\textbf{Z}_j$ $\leftarrow$ $[]$, $[]$
                    
                \EndFor

            \Statex \hspace{0cm}\textcolor{cyan}{/*\textit{Collect activations at each layer, Eq. (1)}*/}
            
            \For{ $\mathcal{I}^i$ $\in$ $\bm{\mathcal{I}}$}
                
                \For{ $j$ $\in$ $n$}
                      \vspace{0.2cm} 
    
                      \State $\textbf{z}^i_j$  $\leftarrow$ $f_j(\mathcal{I}^i)$
    
                      \State $\textbf{Z}_j$.append($\textbf{z}^i_j$)
    
                      \EndFor
                      
                \EndFor

            \Statex \hspace{0cm}\textcolor{cyan}{/*\textit{Iterate through all images}*/}
            
            \For{$i$ $\in$ $|\bm{\mathcal{I}}|$}

            \Statex \hspace{0.5cm}\textcolor{cyan}{/*\textit{Iterate through layers top-down}*/}
                    
                    \For{ $j$ $\in$ $\{n,\dotsc,1\}$}

                        \State $\textbf{B}^i_j$ $\leftarrow$ []

            
                    \Statex \hspace{1cm}\textcolor{cyan}{/*\textit{All features considered at top layer}*/}
                    \If {$j$ == $n$}

                        \State $\widetilde{\textbf{B}}^{i}_{j+1}(\textbf{p};1)$ $\leftarrow$ $\{1\}^{h_{j} \times w_{j}}$

                    
                        \State $\Gamma^i_j$ $\leftarrow$ silhouette($\textbf{z}^i_j(\textbf{p})$ $\odot$ $\widetilde{\textbf{B}}^{i}_{j+1}(\textbf{p};1)$)
                        
                        \State $\{\textbf{B}^{i}_{j}(\textbf{p};\gamma)\}_{\gamma}^{\Gamma^i_j}$ $\leftarrow$ $\mathsf{C}^{seg}_{\Gamma^i_j}(\textbf{z}^i_j(\textbf{p})$ $\odot$ $\widetilde{\textbf{B}}^{i}_{j+1}(\textbf{p};1))$

                        \State $\textbf{B}^i_j$.append($\{\textbf{B}^{i}_{j}(\textbf{p};\gamma)\}_{\gamma}^{\Gamma^i_j}$)
                        
                    \Else

                        \Statex \hspace{1cm}\textcolor{cyan}{/*\textit{Top-down masking for all other layers}*/}
            
                        \For{ $\textbf{B}^{i}_{j+1}(\textbf{p};k)$ $\in$ $\{\textbf{B}^i_{j+1}\}_{\gamma}^{\Gamma^i_{j+1}}$}

                            \Statex \hspace{2cm}\textcolor{cyan}{/*\textit{Bilinear interpolate mask to feature shape}*/}

                            \State $\widetilde{\textbf{B}}^{i}_{j+1}(\textbf{p};k)$ $\leftarrow$ BiInterp($\textbf{B}^{i}_{j+1}(\textbf{p};k)$, $\textbf{z}^i_j$.shape)

                            \Statex \hspace{1.5cm}\textcolor{cyan}{/*\textit{Mask with higher layer binary mask, Eq. (2)}*/}


                            \State $\Gamma^i_j$ $\leftarrow$ silhouette($\textbf{z}^i_j(\textbf{p})$ $\odot$ $\widetilde{\textbf{B}}^{i}_{j+1}(\textbf{p};k)$)
                        
                            \State $\{\textbf{B}^{i}_{j}(\textbf{p};\gamma)\}_{\gamma}^{\Gamma^i_j}$ $\leftarrow$ $\mathsf{C}^{seg}_{\Gamma^i_j}(\textbf{z}^i_j(\textbf{p})$ $\odot$ $\widetilde{\textbf{B}}^{i}_{j+1}(\textbf{p};k))$
  
                            \State $\textbf{B}^i_j$.append($\{\textbf{B}^{i}_{j}(\textbf{p};\gamma)\}_{\gamma}^{\Gamma^i_j}$)

                            \EndFor

                    \EndIf

                    \Statex \hspace{1cm}\textcolor{cyan}{/*\textit{Create and save RGB Masks, Eq. (3)}*/}

                    \For{$\textbf{B}^{i}_j(\textbf{p};\gamma)$ $\in$ $\textbf{B}^i_{j}$}
                    
                        \State $\textbf{M}^{i}_j(\textbf{p};\gamma)$ $\leftarrow$ $\mathcal{I}^i$ $\odot$ $\textbf{B}^{i}_j(\textbf{p};\gamma)$

                        \State $\textbf{M}_j$.append($\textbf{M}^{i}_j(\textbf{p};\gamma)$)

                        \EndFor

                    \State $\textbf{M}$.append($\textbf{M}_j$)
                        
                    \EndFor

                \EndFor     
                
			\Statex \textbf{Return} $\textbf{M}$
				\vspace{0.2cm}}
		\end{algorithmic}
\end{algorithm}

\begin{algorithm}[t]
\caption{\textbf{Concept Discovery} }\label{alg:concept_discovery}
		\begin{algorithmic}[1]
			
            \Statex \textbf{Input}: Model $F$, $n$ selected layers of $F$ to study, Set of RGB Image Masks at each Layer $\textbf{M} = \{\textbf{M}_1,...,\textbf{M}_n\}$, Clustering algorithm $\mathsf{C}^{con}$
            \Statex \hspace{0cm}\textcolor{cyan}{/*\textit{$\mathsf{C}^{con}$ is instatiated in terms of k-means~\cite{lloyd1982least}}*/}
            \Statex \textbf{Output}: Set of concept centroids $\textbf{Q} = \{\textbf{Q}_1,...,\textbf{Q}_n\}$

            \State $\textbf{Q}$ $\leftarrow$ []

            \For{ $j$ $\in$ $n$}

                \Statex \hspace{0.5cm}\textcolor{cyan}{/*\textit{Cluster segment activations, Eq. (4)}*/}

                \State $\textbf{Z}_{\textbf{M}_j}$ $\leftarrow$ $f_j(\textbf{M}_j)$
                
                \State $\textbf{Q}_j$ $\leftarrow$ $\mathsf{C}^{con}(\text{GAP}(\textbf{Z}_{\textbf{M}_j}))$

                \Statex \hspace{0.5cm}\textcolor{cyan}{/*\textit{Prune clusters (see Sec.~\ref{sec:implementation_details}) for details}*/}

                \State $\textbf{Q}_j$ $\leftarrow$ prune($\textbf{Q}_j$)

                \State $\textbf{Q}$.append($\textbf{q}_j$)
                
                \EndFor
\Statex \textbf{Return} $\textbf{Q}$
				\vspace{0.2cm}
		\end{algorithmic}
\end{algorithm}

\begin{algorithm}
\caption{\textbf{Interlayer Testing with Concept Activation Vectors}}\label{alg:itcav}
		\begin{algorithmic}[1]
			
            \Statex \textbf{Input}: Model $F$, higher layer selected to study $l$, lower layer selected to study $j$, Concept Centroid for higher layer $\textbf{q}^{m_l}_l$, Concept centroid for lower layer $\textbf{q}^{m_j}_j$, Set of RGB image masks each associated with higher layer concept centroid $\textbf{M}_{\textbf{q}^{m_l}_l}$, Set of RGB image masks each associated with lower layer concept centroid $\textbf{M}_{\textbf{q}^{m_j}_j}$, Set of random images $\bm{\mathcal{I}}_{rnd}$, Linear classifier $h$
            \Statex \textbf{Output}: Concept connection edge weight between concepts $\textbf{q}^{m_j}_j$ and $\textbf{q}^{m_l}_l$: $e_{\textbf{q}^{m_j}_j, \textbf{q}^{m_l}_l}$


            \Statex \hspace{0cm}\textcolor{cyan}{/*\textit{Get activations for lower level concept}*/}

            \State $\textbf{z}_{\textbf{M}_{\textbf{q}^{m_j}_j}}$ $\leftarrow$ $f_{j}(\textbf{M}_{\textbf{q}^{m_j}_j})$

            \Statex \hspace{0cm}\textcolor{cyan}{/*\textit{Get activations for random concept}*/}

            \State $\textbf{z}_{\bm{\mathcal{I}}_{rnd}}$ $\leftarrow$ $f_{j}(\bm{\mathcal{I}}_{rnd})$

            \Statex \hspace{0cm}\textcolor{cyan}{/*\textit{Train CAV and get orthogonal vector to hyperplane in direction of lower concept}*/}
            
            \State $\textbf{V}_{\textbf{q}^{m_j}_j}$ $\leftarrow$ h($\textbf{z}_{\textbf{M}_{\textbf{q}^{m_j}_j}}$,$\textbf{z}_{\bm{\mathcal{I}}_{rnd}}$).train()

            \State CountPositive $\leftarrow$ 0

            \Statex \hspace{0cm}\textcolor{cyan}{/*\textit{Iterate through higher concept segments}*/}
            
            \For{ $x$ $\in$ $\textbf{M}_{\textbf{q}^{m_l}_l}$}
            
                \State $\textbf{z}_j$ $\leftarrow$ $f_{j}(x)$

                \Statex \hspace{0cm}\textcolor{cyan}{/*\textit{Get gradient of segment at layer $l$ with respect to lower layer $j$}*/}

                \State $\textbf{g}_j$ $\leftarrow$ $\nabla_{f_{j}}$  $||f_{l}(\textbf{z}_j) - \textbf{q}^{m_l}_l||_2$

\Statex \hspace{0cm}\textcolor{cyan}{/*\textit{Calculate sensitivity of upper concept to lower concept, Eq. (5)}*/}

                \State $S_{\textbf{q}^{m_j}_j, \textbf{q}^{m_l}_l}$ = $\textbf{g}_j \cdot \textbf{V}_{\textbf{q}^{m_j}_j}$ 
                
                
                \If{ $S_{\textbf{q}^{m_j}_j, \textbf{q}^{m_l}_l}$ $>$ $1$}

                    \State CountPositive = CountPositive + 1
                    
                    \EndIf
                
                \EndFor

            \Statex \hspace{0cm}\textcolor{cyan}{/*\textit{Calculate fraction of positive alignments, Eq. (6)}*/}
                                
            \State $e_{\textbf{q}^{m_j}_j, \textbf{q}^{m_l}_l}$ = CountPositive \big/ $|\textbf{M}_{\textbf{q}^{m_l}_l}|$  

            \Statex \textbf{Return} $e_{\textbf{q}^{m_j}_j, \textbf{q}^{m_l}_l}$
            
				\vspace{0.2cm}
		\end{algorithmic}
\end{algorithm}

%% file: VCCs/r50_ant.tex
\begin{figure*}
    \begin{center}
    \begin{tikzpicture}[grow=left,
        every node/.style = {inner sep=0pt},
every label/.append style = {label distance=2pt, align = center},
         sibling distance = 6em,
           level 1/.style = {level distance=9em,anchor=east},
           level 2/.style = {level distance=8em,anchor=east},
           level 3/.style = {level distance=8em,anchor=east},
           level 4/.style = {level distance=8em,anchor=east},
           ]  
\node[anchor=south,label=Ant](class ant)
{\includegraphics[width=34mm]{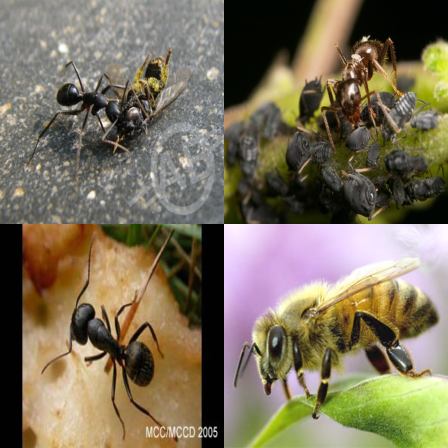}}
    child {node[label=layer4](layer4_ant_concept3){\includegraphics[width=20mm]{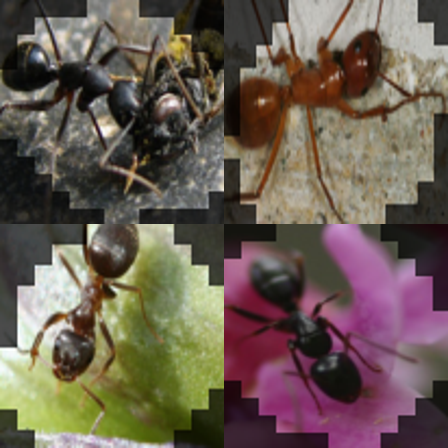}} edge from parent[draw=none]}
    child {node[](layer4_ant_concept2){\includegraphics[width=20mm]{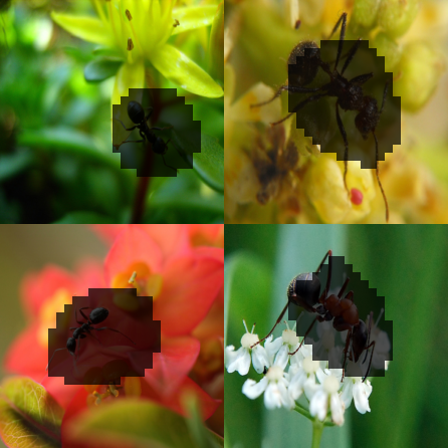}} edge from parent[draw=none]
        child {node[yshift=-0.93cm,label=layer3](layer3_ant_concept1){\includegraphics[width=20mm]{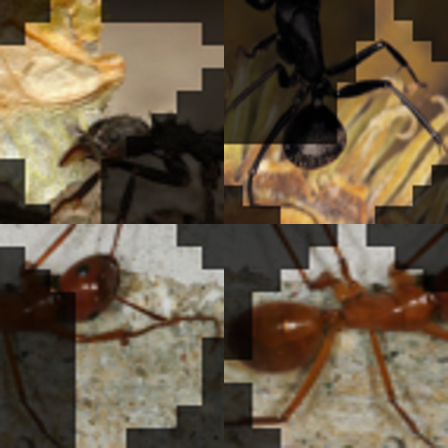}} 
        edge from parent[draw=none]}
        child {node[yshift=-0.93cm](layer3_ant_concept2) {\includegraphics[width=20mm]{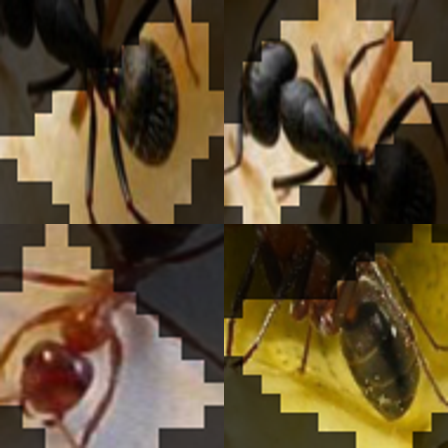}} 
        edge from parent[draw=none]
            child {node[label=layer2](layer2_ant_concept2){\includegraphics[width=20mm]{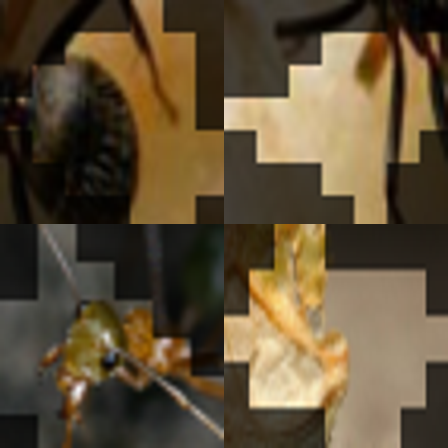}} 
            edge from parent[draw=none]}
            child {node[](layer2_ant_concept5){\includegraphics[width=20mm]{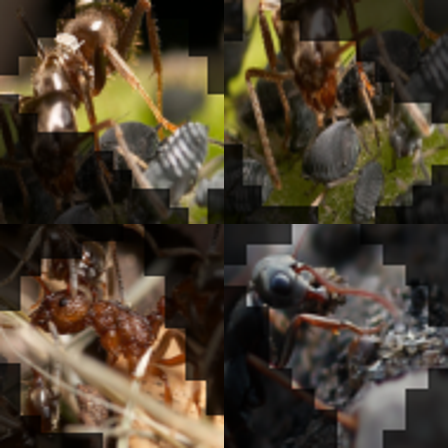}} 
            edge from parent[draw=none]
                child {node[label=layer1](layer1_ant_concept8){\includegraphics[width=20mm]{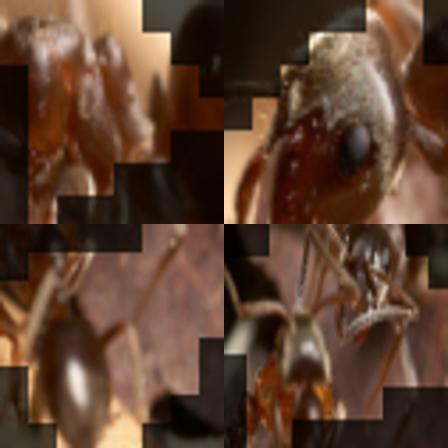}} 
                edge from parent[draw=none]}
                child {node[](layer1_ant_concept7){\includegraphics[width=20mm]{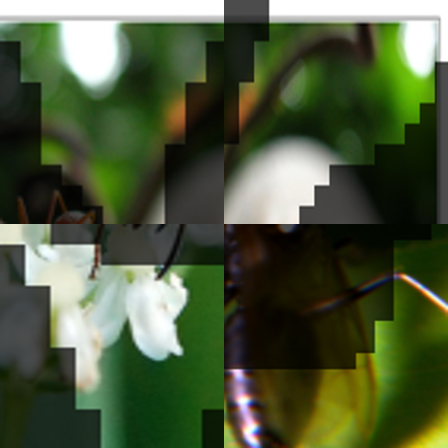}} 
                edge from parent[draw=none]}
                child {node[](layer1_ant_concept2){\includegraphics[width=20mm]{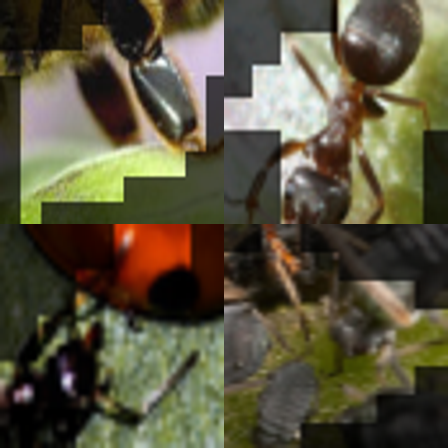}} 
                edge from parent[draw=none]}
                }
            child {node[](layer2_ant_concept3){\includegraphics[width=20mm]{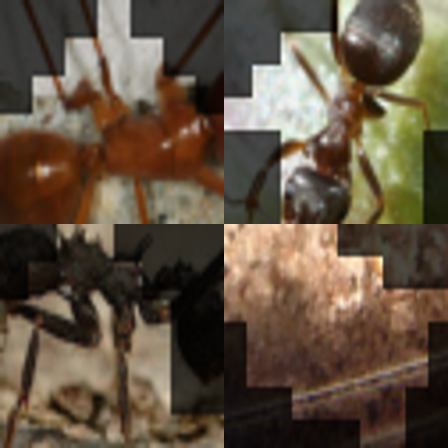}} 
            edge from parent[draw=none]}
            }
        child {node[yshift=-0.93cm](layer3_ant_concept3){\includegraphics[width=20mm]{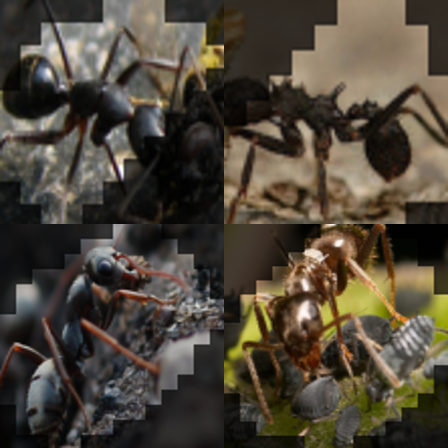}} 
        edge from parent[draw=none]} 
    }  
    child {node[](layer4_ant_concept1){\includegraphics[width=20mm]{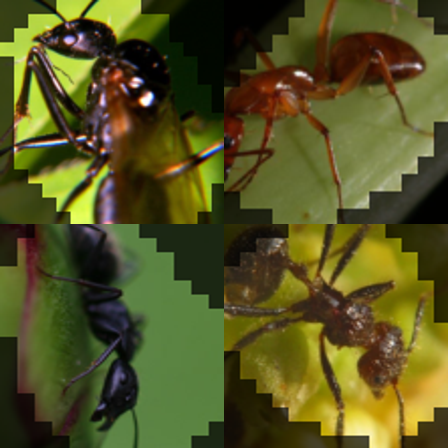}} 
    edge from parent[draw=none]} 
    child {node[](layer4_ant_concept4){\includegraphics[width=20mm]{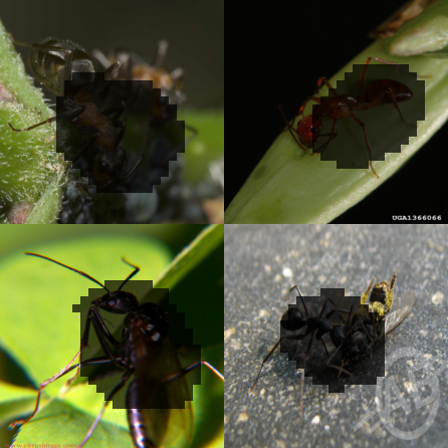}} edge from parent[draw=none]};
\begin{scope}[on background layer]
\draw[line width=0.5236842105263158mm,-,/pgfplots/color of colormap=(477)] ($ (layer1_ant_concept7.east) - (0.2, 0) $) -- ( $ (layer2_ant_concept5.west) + (0.25,0) $);
\draw[line width=0.5357142857142857mm,-,/pgfplots/color of colormap=(465)] ($ (layer2_ant_concept5.east) - (0.2, 0) $) -- ( $ (layer3_ant_concept1.west) + (0.25,0) $);
\draw[line width=0.38157894736842096mm,-,/pgfplots/color of colormap=(619)] ($ (layer1_ant_concept8.east) - (0.2, 0) $) -- ( $ (layer2_ant_concept5.west) + (0.25,0) $);
\draw[line width=0.4092105263157896mm,-,/pgfplots/color of colormap=(591)] ($ (layer1_ant_concept2.east) - (0.2, 0) $) -- ( $ (layer2_ant_concept3.west) + (0.25,0) $);
\draw[line width=0.38809523809523805mm,-,/pgfplots/color of colormap=(612)] ($ (layer2_ant_concept3.east) - (0.2, 0) $) -- ( $ (layer3_ant_concept1.west) + (0.25,0) $);
\draw[line width=0.3868421052631578mm,-,/pgfplots/color of colormap=(614)] ($ (layer2_ant_concept3.east) - (0.2, 0) $) -- ( $ (layer3_ant_concept2.west) + (0.25,0) $);
\draw[line width=0.553125mm,-,/pgfplots/color of colormap=(447)] ($ (layer2_ant_concept2.east) - (0.2, 0) $) -- ( $ (layer3_ant_concept3.west) + (0.25,0) $);
\draw[line width=0.5545454545454545mm,-,/pgfplots/color of colormap=(446)] ($ (layer3_ant_concept3.east) - (0.2, 0) $) -- ( $ (layer4_ant_concept2.west) + (0.25,0) $);
\draw[line width=0.7416666666666666mm,-,/pgfplots/color of colormap=(259)] ($ (layer3_ant_concept3.east) - (0.2, 0) $) -- ( $ (layer4_ant_concept1.west) + (0.25,0) $);
\draw[line width=0.5399999999999999mm,-,/pgfplots/color of colormap=(461)] ($ (layer3_ant_concept1.east) - (0.2, 0) $) -- ( $ (layer4_ant_concept3.west) + (0.25,0) $);
\draw[line width=0.29999999999999993mm,-,/pgfplots/color of colormap=(701)] ($ (layer3_ant_concept1.east) - (0.2, 0) $) -- ( $ (layer4_ant_concept2.west) + (0.25,0) $);
\draw[line width=0.4409090909090909mm,-,/pgfplots/color of colormap=(560)] ($ (layer3_ant_concept2.east) - (0.2, 0) $) -- ( $ (layer4_ant_concept2.west) + (0.25,0) $);
\draw[line width=0.7249999999999999mm,-,/pgfplots/color of colormap=(276)] ($ (layer3_ant_concept2.east) - (0.2, 0) $) -- ( $ (layer4_ant_concept1.west) + (0.25,0) $);
\draw[line width=1.0mm,-,/pgfplots/color of colormap=(0)] ($ (layer4_ant_concept3.east) - (0.2, 0) $) -- ( $ (class ant.west) + (0.25,0) $);
\draw[line width=1.0mm,-,/pgfplots/color of colormap=(0)] ($ (layer4_ant_concept2.east) - (0.2, 0) $) -- ( $ (class ant.west) + (0.25,0) $);
\draw[line width=1.0mm,-,/pgfplots/color of colormap=(0)] ($ (layer4_ant_concept1.east) - (0.2, 0) $) -- ( $ (class ant.west) + (0.25,0) $);
\draw[line width=0.8mm,-,/pgfplots/color of colormap=(200)] ($ (layer4_ant_concept4.east) - (0.2, 0) $) -- ( $ (class ant.west) + (0.25,0) $);

\end{scope}
    \end{tikzpicture}
    \end{center}
    \vspace{-0.5cm}
    \caption{A VCC for four layers of a ResNet50~\cite{he2016deep} model targeting the class ``ant''. Darker lines denote larger concept contributions.}
    \label{fig:vcc_r50}
\end{figure*}

%% file: VCCs/vgg16_curch.tex
\begin{figure*}
    \begin{center}
    \begin{tikzpicture}[grow=left,
        every node/.style = {inner sep=0pt},
every label/.append style = {label distance=2pt, align = center},
         sibling distance = 6em,
           level 1/.style = {level distance=9em,anchor=east},
           level 2/.style = {level distance=8em,anchor=east},
           level 3/.style = {level distance=8em,anchor=east},
           level 4/.style = {level distance=8em,anchor=east},
           ]  
\node[anchor=south,label=Church](class church)
{\includegraphics[width=34mm]{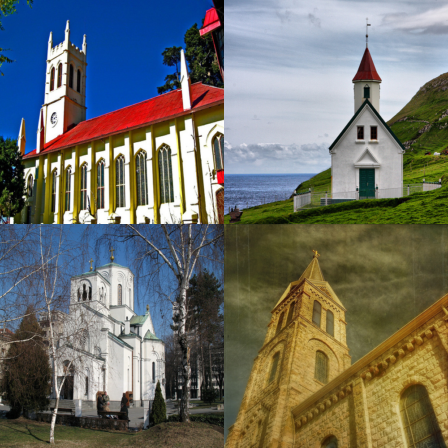}}
    child {node[label=layer4](29_church_concept3){\includegraphics[width=20mm]{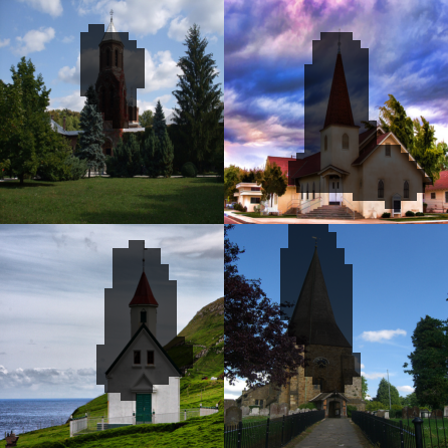}} edge from parent[draw=none]}
    child {node[](29_church_concept2){\includegraphics[width=20mm]{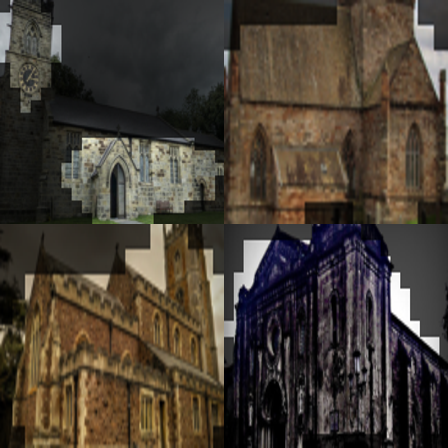}} edge from parent[draw=none]
        child {node[yshift=-0.93cm,label=layer3](22_church_concept4) {\includegraphics[width=20mm]{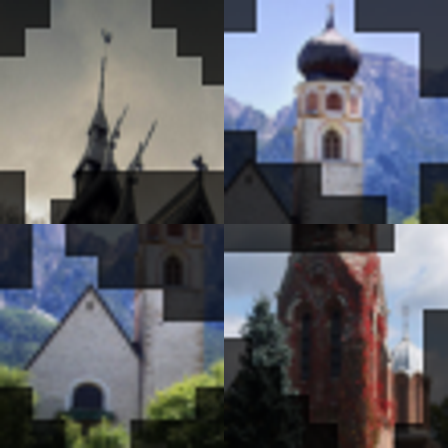}} 
        edge from parent[draw=none]}
        child {node[yshift=-0.93cm](22_church_concept6){\includegraphics[width=20mm]{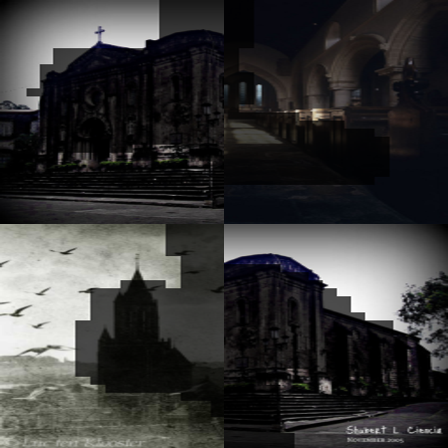}} 
        edge from parent[draw=none]            
        child {node[label=layer2](15_church_concept5){\includegraphics[width=20mm]{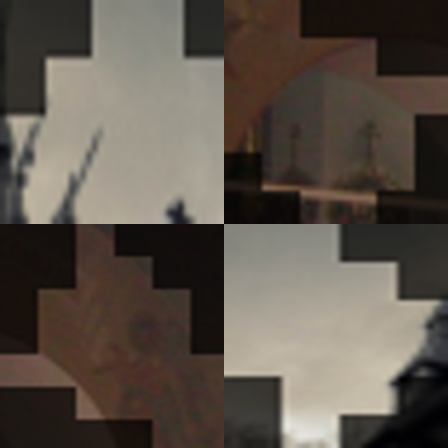}} 
            edge from parent[draw=none]}
            child {node[](15_church_concept2){\includegraphics[width=20mm]{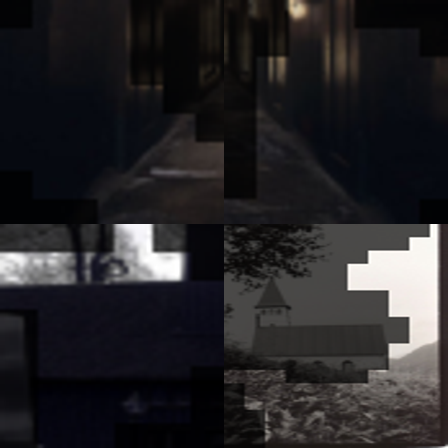}} 
            edge from parent[draw=none]
                child {node[yshift=-0.95cm,label=layer1](8_church_concept3){\includegraphics[width=20mm]{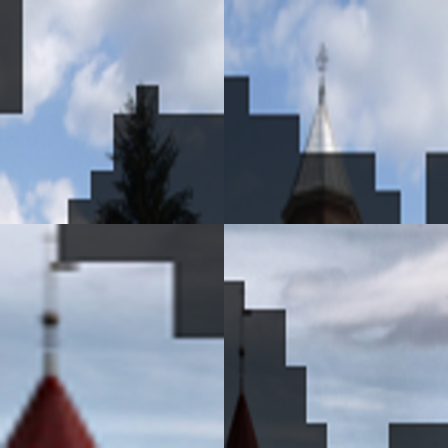}} 
                edge from parent[draw=none]}
                child {node[yshift=-0.95cm](8_church_concept2){\includegraphics[width=20mm]{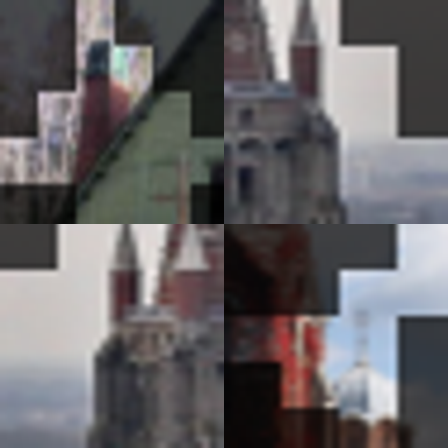}} 
                edge from parent[draw=none]}
                child {node[yshift=-0.95cm](8_church_concept4){\includegraphics[width=20mm]{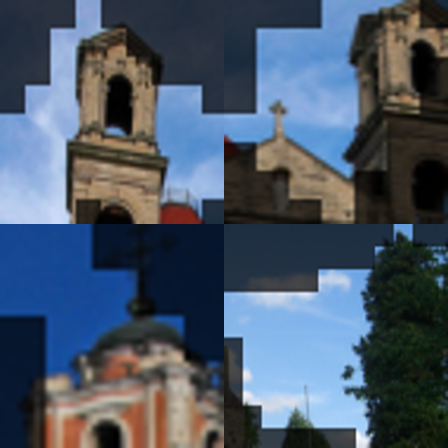}} 
                edge from parent[draw=none]}
                child {node[yshift=-0.95cm](8_church_concept13){\includegraphics[width=20mm]{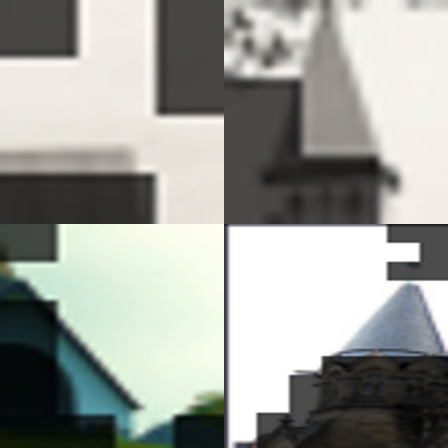}} 
                edge from parent[draw=none]}
                }
            child {node[](15_church_concept7){\includegraphics[width=20mm]{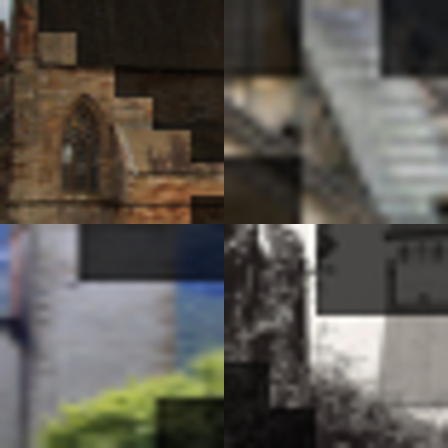}} 
            edge from parent[draw=none]}
            child {node[](15_church_concept6){\includegraphics[width=20mm]{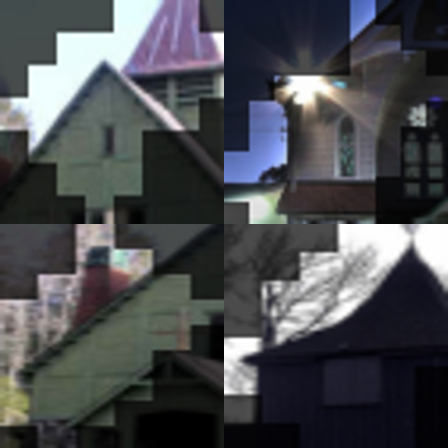}} 
            edge from parent[draw=none]}} 
        child {node[yshift=-0.93cm](22_church_concept1){\includegraphics[width=20mm]{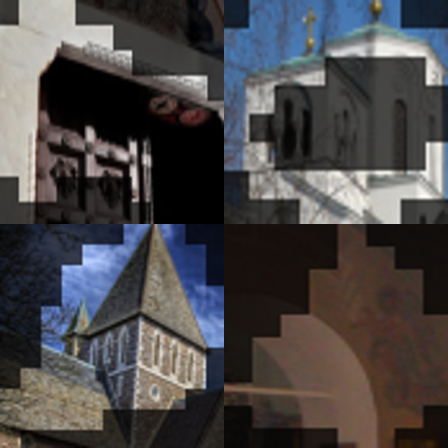}} 
        edge from parent[draw=none]} 
    }  
    child {node[](29_church_concept1){\includegraphics[width=20mm]{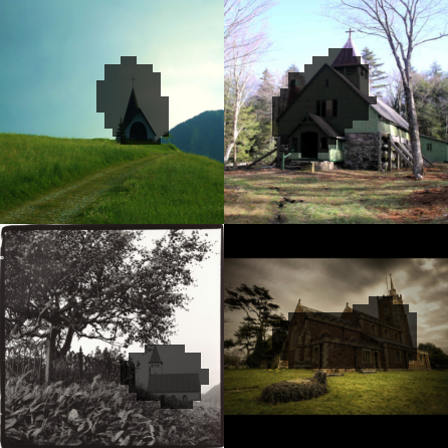}} 
    edge from parent[draw=none]} 
    child {node[](29_church_concept4){\includegraphics[width=20mm]{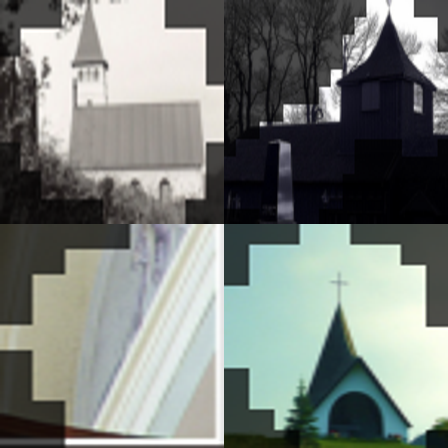}} edge from parent[draw=none]};
\begin{scope}[on background layer]
\draw[line width=0.2720588235294118mm,-,/pgfplots/color of colormap=(728)] ($ (8_church_concept13.east) - (0.2, 0) $) -- ( $ (15_church_concept7.west) + (0.25,0) $);
\draw[line width=0.1975mm,-,/pgfplots/color of colormap=(803)] ($ (8_church_concept13.east) - (0.2, 0) $) -- ( $ (15_church_concept6.west) + (0.25,0) $);
\draw[line width=0.3125mm,-,/pgfplots/color of colormap=(688)] ($ (15_church_concept7.east) - (0.2, 0) $) -- ( $ (22_church_concept1.west) + (0.25,0) $);
\draw[line width=0.365mm,-,/pgfplots/color of colormap=(635)] ($ (15_church_concept6.east) - (0.2, 0) $) -- ( $ (22_church_concept1.west) + (0.25,0) $);
\draw[line width=0.49230769230769234mm,-,/pgfplots/color of colormap=(508)] ($ (15_church_concept6.east) - (0.2, 0) $) -- ( $ (22_church_concept6.west) + (0.25,0) $);
\draw[line width=0.38400000000000006mm,-,/pgfplots/color of colormap=(616)] ($ (8_church_concept4.east) - (0.2, 0) $) -- ( $ (15_church_concept2.west) + (0.25,0) $);
\draw[line width=0.4647058823529412mm,-,/pgfplots/color of colormap=(536)] ($ (8_church_concept4.east) - (0.2, 0) $) -- ( $ (15_church_concept7.west) + (0.25,0) $);
\draw[line width=0.3299999999999999mm,-,/pgfplots/color of colormap=(671)] ($ (15_church_concept2.east) - (0.2, 0) $) -- ( $ (22_church_concept1.west) + (0.25,0) $);
\draw[line width=0.43846153846153857mm,-,/pgfplots/color of colormap=(562)] ($ (15_church_concept2.east) - (0.2, 0) $) -- ( $ (22_church_concept6.west) + (0.25,0) $);
\draw[line width=0.2647058823529412mm,-,/pgfplots/color of colormap=(736)] ($ (8_church_concept2.east) - (0.2, 0) $) -- ( $ (15_church_concept7.west) + (0.25,0) $);
\draw[line width=0.36374999999999996mm,-,/pgfplots/color of colormap=(637)] ($ (8_church_concept2.east) - (0.2, 0) $) -- ( $ (15_church_concept5.west) + (0.25,0) $);
\draw[line width=0.20749999999999996mm,-,/pgfplots/color of colormap=(793)] ($ (8_church_concept2.east) - (0.2, 0) $) -- ( $ (15_church_concept6.west) + (0.25,0) $);
\draw[line width=0.30000000000000004mm,-,/pgfplots/color of colormap=(700)] ($ (15_church_concept5.east) - (0.2, 0) $) -- ( $ (22_church_concept4.west) + (0.25,0) $);
\draw[line width=0.26999999999999996mm,-,/pgfplots/color of colormap=(731)] ($ (15_church_concept5.east) - (0.2, 0) $) -- ( $ (22_church_concept1.west) + (0.25,0) $);
\draw[line width=0.23676470588235293mm,-,/pgfplots/color of colormap=(764)] ($ (8_church_concept3.east) - (0.2, 0) $) -- ( $ (15_church_concept7.west) + (0.25,0) $);
\draw[line width=0.3575mm,-,/pgfplots/color of colormap=(643)] ($ (8_church_concept3.east) - (0.2, 0) $) -- ( $ (15_church_concept5.west) + (0.25,0) $);
\draw[line width=0.1825mm,-,/pgfplots/color of colormap=(818)] ($ (8_church_concept3.east) - (0.2, 0) $) -- ( $ (15_church_concept6.west) + (0.25,0) $);
\draw[line width=0.5454545454545454mm,-,/pgfplots/color of colormap=(455)] ($ (22_church_concept1.east) - (0.2, 0) $) -- ( $ (29_church_concept3.west) + (0.25,0) $);
\draw[line width=0.48235294117647065mm,-,/pgfplots/color of colormap=(518)] ($ (22_church_concept1.east) - (0.2, 0) $) -- ( $ (29_church_concept4.west) + (0.25,0) $);
\draw[line width=0.465mm,-,/pgfplots/color of colormap=(535)] ($ (22_church_concept1.east) - (0.2, 0) $) -- ( $ (29_church_concept2.west) + (0.25,0) $);
\draw[line width=0.43181818181818177mm,-,/pgfplots/color of colormap=(569)] ($ (22_church_concept6.east) - (0.2, 0) $) -- ( $ (29_church_concept3.west) + (0.25,0) $);
\draw[line width=0.33333333333333337mm,-,/pgfplots/color of colormap=(667)] ($ (22_church_concept6.east) - (0.2, 0) $) -- ( $ (29_church_concept1.west) + (0.25,0) $);
\draw[line width=0.3970588235294118mm,-,/pgfplots/color of colormap=(603)] ($ (22_church_concept6.east) - (0.2, 0) $) -- ( $ (29_church_concept4.west) + (0.25,0) $);
\draw[line width=0.505mm,-,/pgfplots/color of colormap=(495)] ($ (22_church_concept6.east) - (0.2, 0) $) -- ( $ (29_church_concept2.west) + (0.25,0) $);
\draw[line width=0.44999999999999984mm,-,/pgfplots/color of colormap=(551)] ($ (22_church_concept4.east) - (0.2, 0) $) -- ( $ (29_church_concept3.west) + (0.25,0) $);
\draw[line width=0.438235294117647mm,-,/pgfplots/color of colormap=(562)] ($ (22_church_concept4.east) - (0.2, 0) $) -- ( $ (29_church_concept4.west) + (0.25,0) $);
\draw[line width=0.4600000000000001mm,-,/pgfplots/color of colormap=(540)] ($ (22_church_concept4.east) - (0.2, 0) $) -- ( $ (29_church_concept2.west) + (0.25,0) $);
\draw[line width=0.9872340425531915mm,-,/pgfplots/color of colormap=(13)] ($ (29_church_concept3.east) - (0.2, 0) $) -- ( $ (class church.west) + (0.25,0) $);
\draw[line width=0.873404255319149mm,-,/pgfplots/color of colormap=(127)] ($ (29_church_concept4.east) - (0.2, 0) $) -- ( $ (class church.west) + (0.25,0) $);
\draw[line width=0.8085106382978724mm,-,/pgfplots/color of colormap=(192)] ($ (29_church_concept2.east) - (0.2, 0) $) -- ( $ (class church.west) + (0.25,0) $);
\draw[line width=0.971276595744681mm,-,/pgfplots/color of colormap=(29)] ($ (29_church_concept1.east) - (0.2, 0) $) -- ( $ (class church.west) + (0.25,0) $);

\end{scope}
    \end{tikzpicture}
    \end{center}
    \vspace{-0.5cm}
    \caption{A VCC for four layers of a VGG16~\cite{szegedy2015going} model targeting the class ``church''. Darker lines denote larger concept contributions.}
    \label{fig:vcc_vgg16}
\end{figure*}

%% file: VCCs/mobilenetv3_guineapig.tex
\begin{figure*}
    \begin{center}
    \begin{tikzpicture}[grow=left,
        every node/.style = {inner sep=0pt},
every label/.append style = {label distance=2pt, align = center},
         sibling distance = 6em,
           level 1/.style = {level distance=9em,anchor=east},
           level 2/.style = {level distance=8em,anchor=east},
           level 3/.style = {level distance=8em,anchor=east},
           level 4/.style = {level distance=8em,anchor=east},
           ]  
\node[anchor=south,label=Guinea Pig](class guinea_pig)
{\includegraphics[width=34mm]{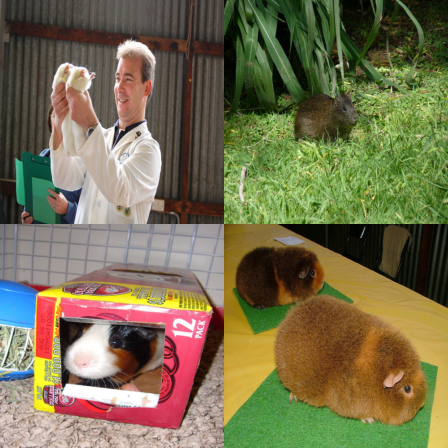}}
    child {node[label=layer4](6_guinea_pig_concept3){\includegraphics[width=20mm]{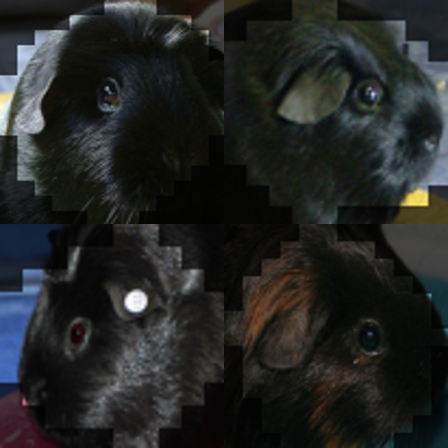}} edge from parent[draw=none]}
    child {node[](6_guinea_pig_concept2){\includegraphics[width=20mm]{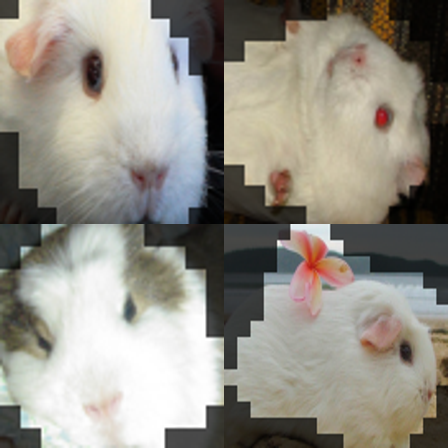}} edge from parent[draw=none]
        child {node[yshift=-0.95cm,label=layer3](4_guinea_pig_concept2){\includegraphics[width=20mm]{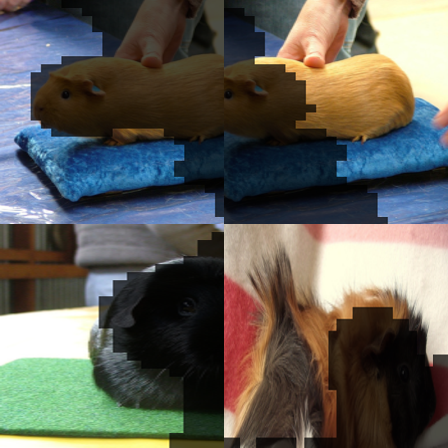}} 
        edge from parent[draw=none]}
        child {node[yshift=-0.95cm](4_guinea_pig_concept3) {\includegraphics[width=20mm]{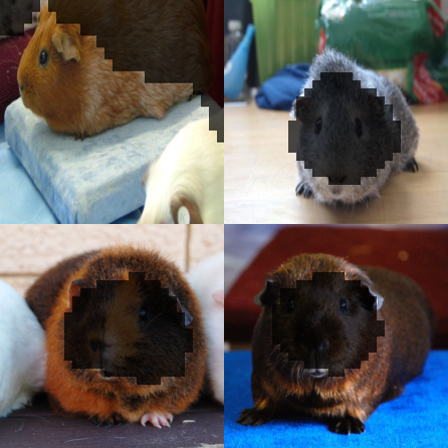}} 
        edge from parent[draw=none]
            child {node[yshift=-0.95cm,label=layer2](2_guinea_pig_concept3){\includegraphics[width=20mm]{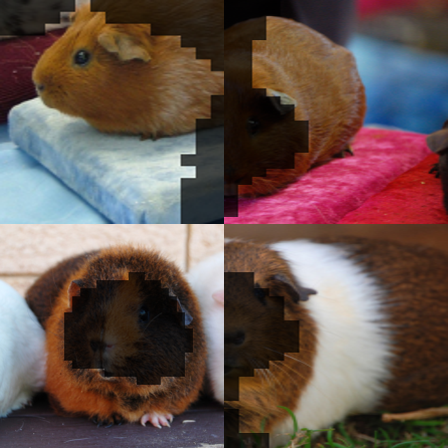}} 
            edge from parent[draw=none]}
            child {node[yshift=-0.95cm](2_guinea_pig_concept4){\includegraphics[width=20mm]{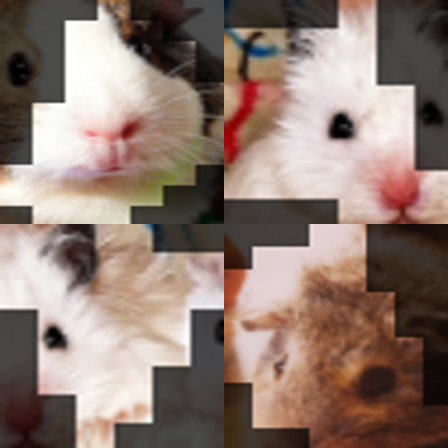}} 
            edge from parent[draw=none]
                child {node[yshift=-0.95cm,label=layer1](0_guinea_pig_concept6){\includegraphics[width=20mm]{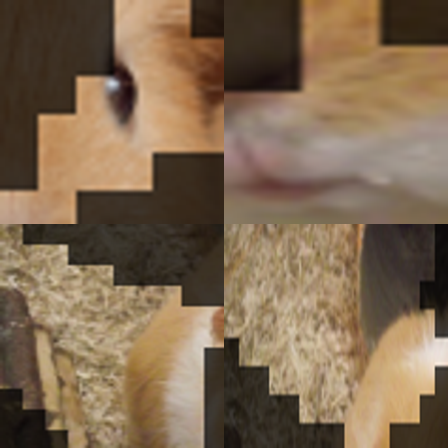}} 
                edge from parent[draw=none]}
                child {node[yshift=-0.95cm](0_guinea_pig_concept9){\includegraphics[width=20mm]{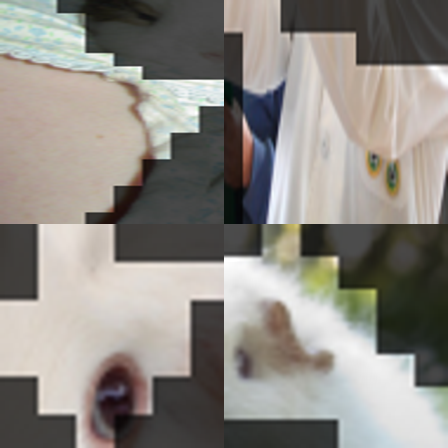}} 
                edge from parent[draw=none]}
                child {node[yshift=-0.95cm](0_guinea_pig_concept5){\includegraphics[width=20mm]{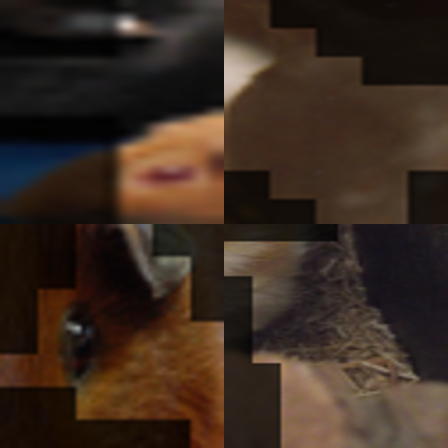}} 
                edge from parent[draw=none]}
                child {node[yshift=-0.95cm](0_guinea_pig_concept8){\includegraphics[width=20mm]{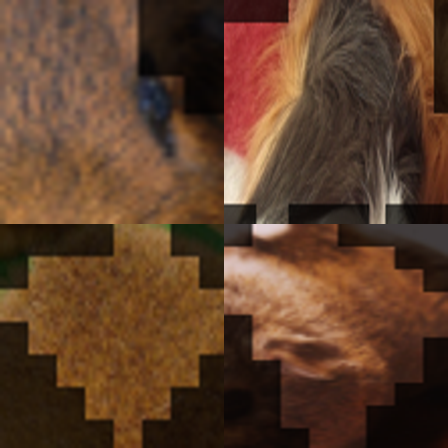}} 
                edge from parent[draw=none]}
                }
            child {node[yshift=-0.95cm](2_guinea_pig_concept8){\includegraphics[width=20mm]{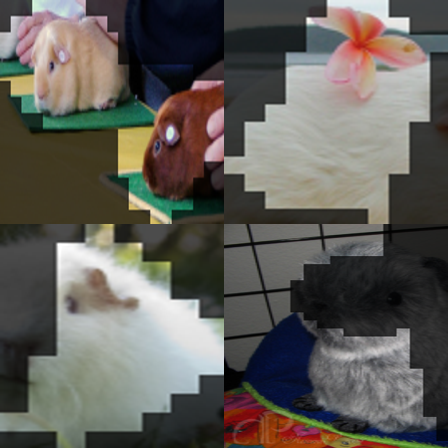}} 
            edge from parent[draw=none]}
            child {node[yshift=-0.95cm](2_guinea_pig_concept10){\includegraphics[width=20mm]{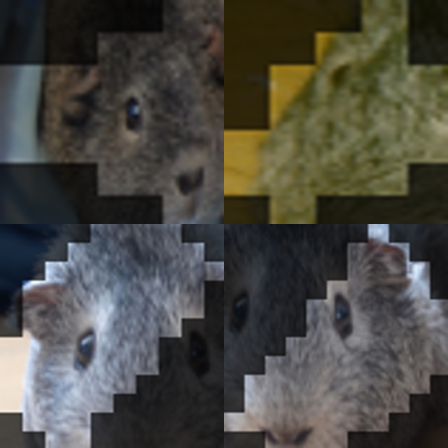}} 
            edge from parent[draw=none]}
            }
        child {node[yshift=-0.95cm](4_guinea_pig_concept5){\includegraphics[width=20mm]{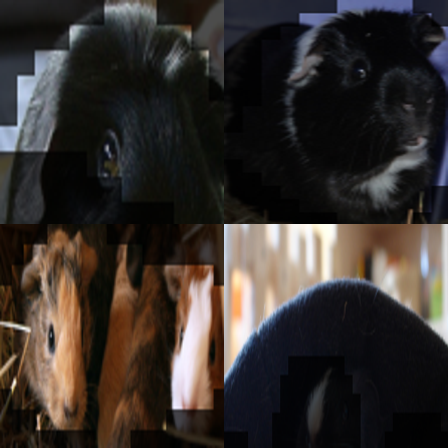}} 
        edge from parent[draw=none]} 
        child {node[yshift=-0.95cm](4_guinea_pig_concept6){\includegraphics[width=20mm]{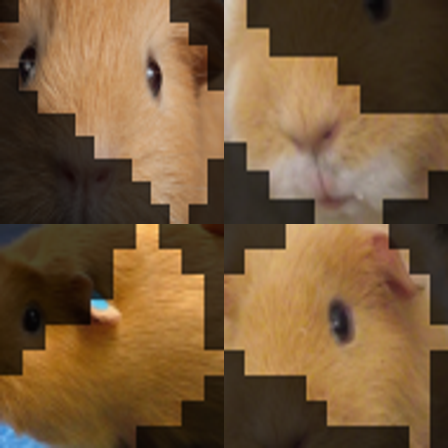}} 
        edge from parent[draw=none]} 
    }  
    child {node[](6_guinea_pig_concept1){\includegraphics[width=20mm]{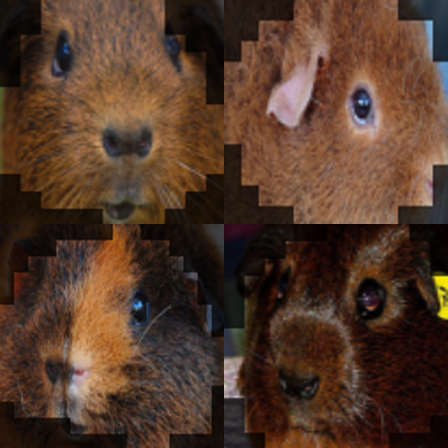}} 
    edge from parent[draw=none]} 
    child {node[](6_guinea_pig_concept4){\includegraphics[width=20mm]{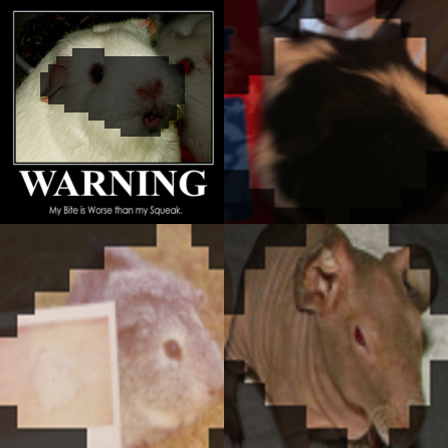}} edge from parent[draw=none]};
\begin{scope}[on background layer]
\draw[line width=0.4676470588235294mm,-,/pgfplots/color of colormap=(533)] ($ (0_guinea_pig_concept8.east) - (0.2, 0) $) -- ( $ (2_guinea_pig_concept10.west) + (0.25,0) $);
\draw[line width=0.4333333333333334mm,-,/pgfplots/color of colormap=(567)] ($ (0_guinea_pig_concept8.east) - (0.2, 0) $) -- ( $ (2_guinea_pig_concept8.west) + (0.25,0) $);
\draw[line width=0.4666666666666667mm,-,/pgfplots/color of colormap=(534)] ($ (0_guinea_pig_concept8.east) - (0.2, 0) $) -- ( $ (2_guinea_pig_concept4.west) + (0.25,0) $);
\draw[line width=0.4454545454545453mm,-,/pgfplots/color of colormap=(555)] ($ (2_guinea_pig_concept10.east) - (0.2, 0) $) -- ( $ (4_guinea_pig_concept2.west) + (0.25,0) $);
\draw[line width=0.4833333333333333mm,-,/pgfplots/color of colormap=(517)] ($ (2_guinea_pig_concept10.east) - (0.2, 0) $) -- ( $ (4_guinea_pig_concept6.west) + (0.25,0) $);
\draw[line width=0.45500000000000007mm,-,/pgfplots/color of colormap=(545)] ($ (2_guinea_pig_concept10.east) - (0.2, 0) $) -- ( $ (4_guinea_pig_concept5.west) + (0.25,0) $);
\draw[line width=0.35mm,-,/pgfplots/color of colormap=(650)] ($ (2_guinea_pig_concept10.east) - (0.2, 0) $) -- ( $ (4_guinea_pig_concept3.west) + (0.25,0) $);
\draw[line width=0.44499999999999995mm,-,/pgfplots/color of colormap=(556)] ($ (2_guinea_pig_concept8.east) - (0.2, 0) $) -- ( $ (4_guinea_pig_concept5.west) + (0.25,0) $);
\draw[line width=0.49615384615384617mm,-,/pgfplots/color of colormap=(504)] ($ (2_guinea_pig_concept8.east) - (0.2, 0) $) -- ( $ (4_guinea_pig_concept3.west) + (0.25,0) $);
\draw[line width=0.4681818181818181mm,-,/pgfplots/color of colormap=(532)] ($ (2_guinea_pig_concept4.east) - (0.2, 0) $) -- ( $ (4_guinea_pig_concept2.west) + (0.25,0) $);
\draw[line width=0.4230769230769231mm,-,/pgfplots/color of colormap=(577)] ($ (2_guinea_pig_concept4.east) - (0.2, 0) $) -- ( $ (4_guinea_pig_concept3.west) + (0.25,0) $);
\draw[line width=0.45mm,-,/pgfplots/color of colormap=(550)] ($ (0_guinea_pig_concept5.east) - (0.2, 0) $) -- ( $ (2_guinea_pig_concept10.west) + (0.25,0) $);
\draw[line width=0.4566666666666667mm,-,/pgfplots/color of colormap=(544)] ($ (0_guinea_pig_concept5.east) - (0.2, 0) $) -- ( $ (2_guinea_pig_concept8.west) + (0.25,0) $);
\draw[line width=0.4366666666666667mm,-,/pgfplots/color of colormap=(564)] ($ (0_guinea_pig_concept5.east) - (0.2, 0) $) -- ( $ (2_guinea_pig_concept4.west) + (0.25,0) $);
\draw[line width=0.4184210526315789mm,-,/pgfplots/color of colormap=(582)] ($ (0_guinea_pig_concept9.east) - (0.2, 0) $) -- ( $ (2_guinea_pig_concept3.west) + (0.25,0) $);
\draw[line width=0.5mm,-,/pgfplots/color of colormap=(500)] ($ (0_guinea_pig_concept9.east) - (0.2, 0) $) -- ( $ (2_guinea_pig_concept10.west) + (0.25,0) $);
\draw[line width=0.44666666666666666mm,-,/pgfplots/color of colormap=(554)] ($ (0_guinea_pig_concept9.east) - (0.2, 0) $) -- ( $ (2_guinea_pig_concept8.west) + (0.25,0) $);
\draw[line width=0.4333333333333334mm,-,/pgfplots/color of colormap=(567)] ($ (0_guinea_pig_concept9.east) - (0.2, 0) $) -- ( $ (2_guinea_pig_concept4.west) + (0.25,0) $);
\draw[line width=0.5000000000000001mm,-,/pgfplots/color of colormap=(500)] ($ (2_guinea_pig_concept3.east) - (0.2, 0) $) -- ( $ (4_guinea_pig_concept6.west) + (0.25,0) $);
\draw[line width=0.5899999999999999mm,-,/pgfplots/color of colormap=(411)] ($ (2_guinea_pig_concept3.east) - (0.2, 0) $) -- ( $ (4_guinea_pig_concept5.west) + (0.25,0) $);
\draw[line width=0.48461538461538456mm,-,/pgfplots/color of colormap=(516)] ($ (2_guinea_pig_concept3.east) - (0.2, 0) $) -- ( $ (4_guinea_pig_concept3.west) + (0.25,0) $);
\draw[line width=0.4647058823529412mm,-,/pgfplots/color of colormap=(536)] ($ (0_guinea_pig_concept6.east) - (0.2, 0) $) -- ( $ (2_guinea_pig_concept10.west) + (0.25,0) $);
\draw[line width=0.4666666666666667mm,-,/pgfplots/color of colormap=(534)] ($ (0_guinea_pig_concept6.east) - (0.2, 0) $) -- ( $ (2_guinea_pig_concept8.west) + (0.25,0) $);
\draw[line width=0.44333333333333336mm,-,/pgfplots/color of colormap=(557)] ($ (0_guinea_pig_concept6.east) - (0.2, 0) $) -- ( $ (2_guinea_pig_concept4.west) + (0.25,0) $);
\draw[line width=0.7649999999999999mm,-,/pgfplots/color of colormap=(236)] ($ (4_guinea_pig_concept6.east) - (0.2, 0) $) -- ( $ (6_guinea_pig_concept1.west) + (0.25,0) $);
\draw[line width=0.6555555555555553mm,-,/pgfplots/color of colormap=(345)] ($ (4_guinea_pig_concept6.east) - (0.2, 0) $) -- ( $ (6_guinea_pig_concept2.west) + (0.25,0) $);
\draw[line width=0.7250000000000002mm,-,/pgfplots/color of colormap=(275)] ($ (4_guinea_pig_concept6.east) - (0.2, 0) $) -- ( $ (6_guinea_pig_concept4.west) + (0.25,0) $);
\draw[line width=0.7100000000000001mm,-,/pgfplots/color of colormap=(290)] ($ (4_guinea_pig_concept5.east) - (0.2, 0) $) -- ( $ (6_guinea_pig_concept1.west) + (0.25,0) $);
\draw[line width=0.6249999999999999mm,-,/pgfplots/color of colormap=(376)] ($ (4_guinea_pig_concept5.east) - (0.2, 0) $) -- ( $ (6_guinea_pig_concept4.west) + (0.25,0) $);
\draw[line width=0.5125000000000001mm,-,/pgfplots/color of colormap=(488)] ($ (4_guinea_pig_concept3.east) - (0.2, 0) $) -- ( $ (6_guinea_pig_concept3.west) + (0.25,0) $);
\draw[line width=0.6549999999999998mm,-,/pgfplots/color of colormap=(346)] ($ (4_guinea_pig_concept2.east) - (0.2, 0) $) -- ( $ (6_guinea_pig_concept1.west) + (0.25,0) $);
\draw[line width=0.6892857142857143mm,-,/pgfplots/color of colormap=(311)] ($ (4_guinea_pig_concept2.east) - (0.2, 0) $) -- ( $ (6_guinea_pig_concept4.west) + (0.25,0) $);
\draw[line width=0.976mm,-,/pgfplots/color of colormap=(24)] ($ (6_guinea_pig_concept1.east) - (0.2, 0) $) -- ( $ (class guinea_pig.west) + (0.25,0) $);
\draw[line width=0.9280000000000002mm,-,/pgfplots/color of colormap=(72)] ($ (6_guinea_pig_concept4.east) - (0.2, 0) $) -- ( $ (class guinea_pig.west) + (0.25,0) $);
\draw[line width=0.9719999999999999mm,-,/pgfplots/color of colormap=(29)] ($ (6_guinea_pig_concept2.east) - (0.2, 0) $) -- ( $ (class guinea_pig.west) + (0.25,0) $);
\draw[line width=0.9730000000000001mm,-,/pgfplots/color of colormap=(27)] ($ (6_guinea_pig_concept3.east) - (0.2, 0) $) -- ( $ (class guinea_pig.west) + (0.25,0) $);

\end{scope}
    \end{tikzpicture}
    \end{center}
    \vspace{-0.5cm}
    \caption{A VCC for four layers of a MobileNetv3~\cite{howard2019searching} model targeting the class ``guinea pig''. Darker lines denote larger concept contributions.}
    \label{fig:vcc_mobilenetv3}
\end{figure*}

%% file: VCCs/mvit_crutch.tex
\begin{figure*}
    \begin{center}
    \begin{tikzpicture}[grow=left,
        every node/.style = {inner sep=0pt},
every label/.append style = {label distance=2pt, align = center},
         sibling distance = 6em,
           level 1/.style = {level distance=9em,anchor=east},
           level 2/.style = {level distance=8em,anchor=east},
           level 3/.style = {level distance=8em,anchor=east},
           level 4/.style = {level distance=8em,anchor=east},
           ]  
\node[anchor=south,label=Crutch](class crutch)
{\includegraphics[width=34mm]{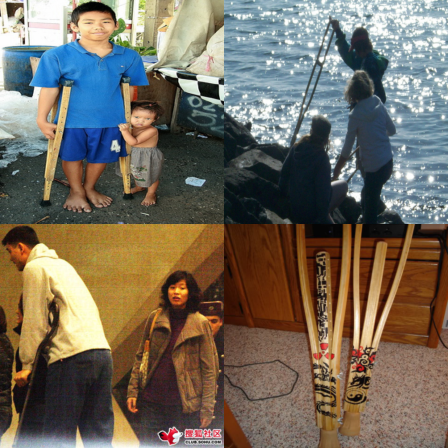}}
    child {node[label=layer4](15_crutch_concept2){\includegraphics[width=20mm]{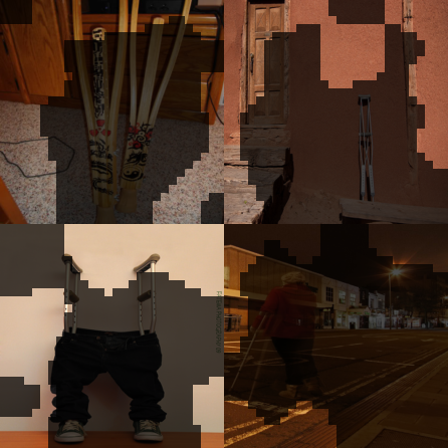}} edge from parent[draw=none]
        child {node[yshift=-0.95cm,label=layer3](9_crutch_concept1){\includegraphics[width=20mm]{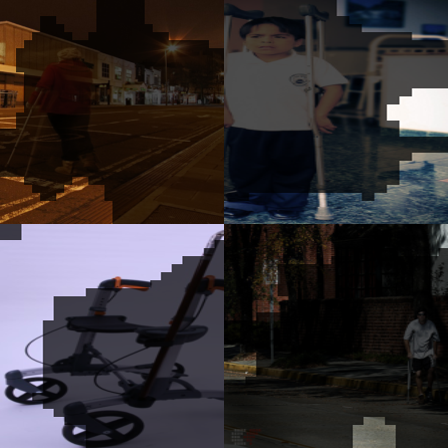}} 
        edge from parent[draw=none]}
        child {node[yshift=-0.95cm](9_crutch_concept3) {\includegraphics[width=20mm]{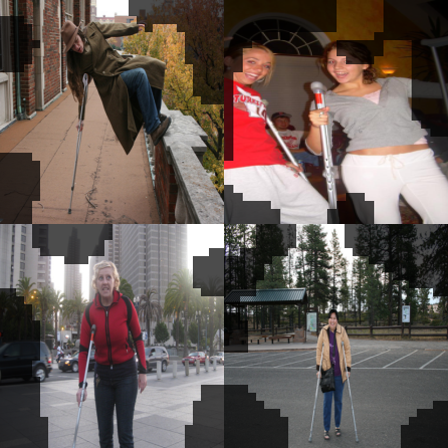}} 
        edge from parent[draw=none]
            child {node[yshift=-0.95cm,label=layer2](3_crutch_concept3){\includegraphics[width=20mm]{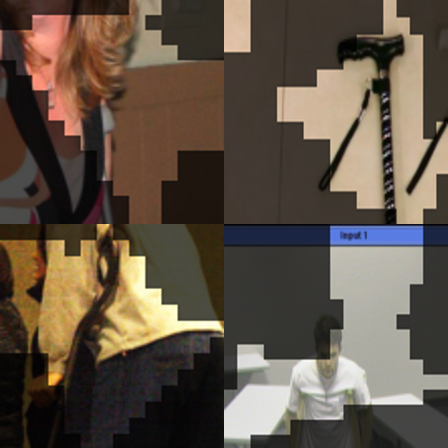}} 
            edge from parent[draw=none]}
            child {node[yshift=-0.95cm](3_crutch_concept6){\includegraphics[width=20mm]{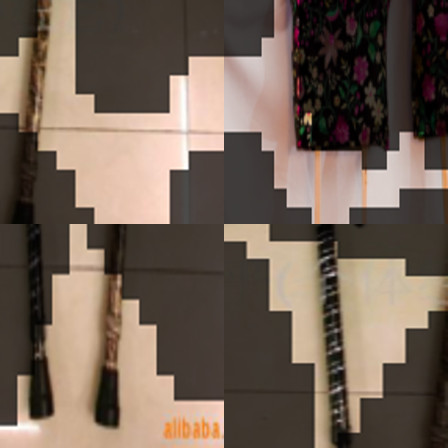}} 
            edge from parent[draw=none]
                child {node[yshift=-0.95cm,label=layer1](1_crutch_concept1){\includegraphics[width=20mm]{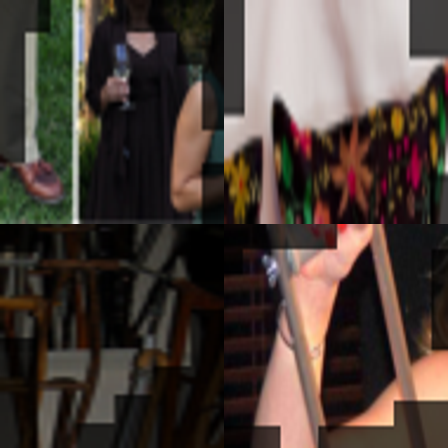}} 
                edge from parent[draw=none]}
                child {node[yshift=-0.95cm](1_crutch_concept13){\includegraphics[width=20mm]{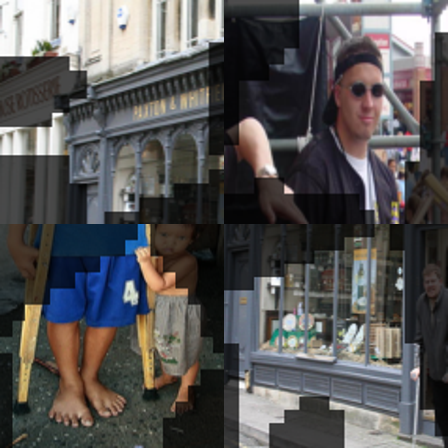}} 
                edge from parent[draw=none]}
                }
            child {node[yshift=-0.95cm](3_crutch_concept8){\includegraphics[width=20mm]{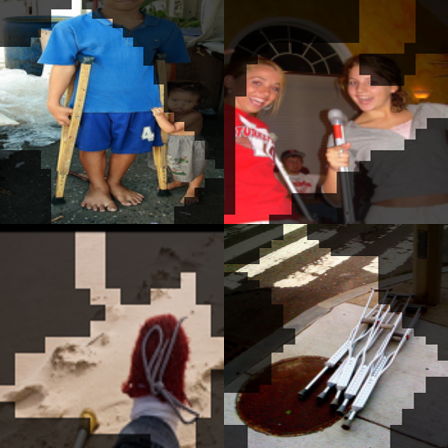}} 
            edge from parent[draw=none]}
            child {node[yshift=-0.95cm](3_crutch_concept7){\includegraphics[width=20mm]{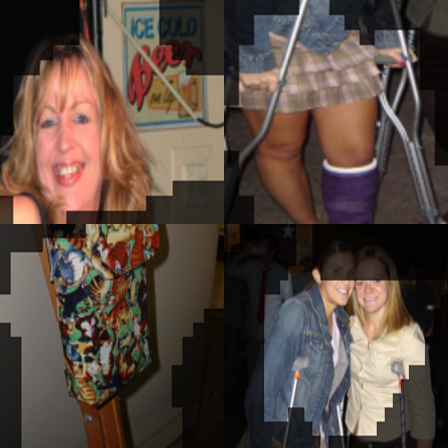}} 
            edge from parent[draw=none]}
            }
        child {node[yshift=-0.95cm,](9_crutch_concept5){\includegraphics[width=20mm]{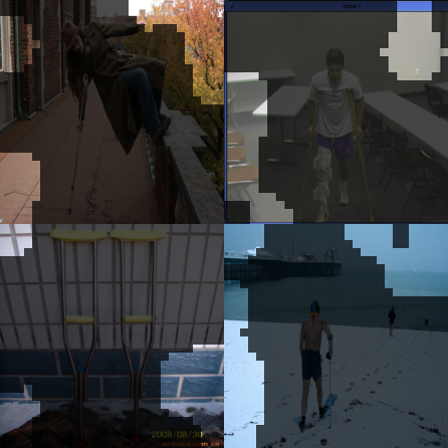}} 
        edge from parent[draw=none]} 
        child {node[yshift=-0.95cm,](9_crutch_concept6){\includegraphics[width=20mm]{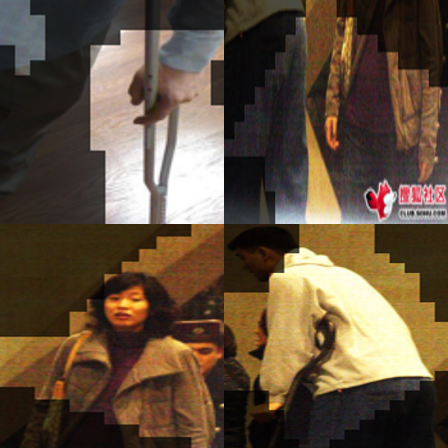}} 
        edge from parent[draw=none]} 
    }  
    child {node[](15_crutch_concept1){\includegraphics[width=20mm]{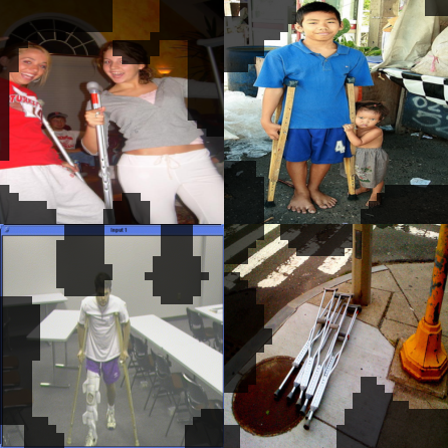}} edge from parent[draw=none]};
\begin{scope}[on background layer]
\draw[line width=0.5519999999999998mm,-,/pgfplots/color of colormap=(449)] ($ (1_crutch_concept13.east) - (0.2, 0) $) -- ( $ (3_crutch_concept3.west) + (0.25,0) $);
\draw[line width=0.45mm,-,/pgfplots/color of colormap=(550)] ($ (3_crutch_concept3.east) - (0.2, 0) $) -- ( $ (9_crutch_concept3.west) + (0.25,0) $);
\draw[line width=0.6199999999999999mm,-,/pgfplots/color of colormap=(381)] ($ (1_crutch_concept1.east) - (0.2, 0) $) -- ( $ (3_crutch_concept7.west) + (0.25,0) $);
\draw[line width=0.5240000000000001mm,-,/pgfplots/color of colormap=(476)] ($ (1_crutch_concept1.east) - (0.2, 0) $) -- ( $ (3_crutch_concept3.west) + (0.25,0) $);
\draw[line width=0.4423076923076924mm,-,/pgfplots/color of colormap=(558)] ($ (3_crutch_concept7.east) - (0.2, 0) $) -- ( $ (9_crutch_concept3.west) + (0.25,0) $);
\draw[line width=0.35666666666666674mm,-,/pgfplots/color of colormap=(644)] ($ (3_crutch_concept7.east) - (0.2, 0) $) -- ( $ (9_crutch_concept5.west) + (0.25,0) $);
\draw[line width=0.2833333333333333mm,-,/pgfplots/color of colormap=(717)] ($ (3_crutch_concept7.east) - (0.2, 0) $) -- ( $ (9_crutch_concept1.west) + (0.25,0) $);
\draw[line width=0.38125mm,-,/pgfplots/color of colormap=(619)] ($ (3_crutch_concept7.east) - (0.2, 0) $) -- ( $ (9_crutch_concept6.west) + (0.25,0) $);
\draw[line width=0.565mm,-,/pgfplots/color of colormap=(435)] ($ (9_crutch_concept3.east) - (0.2, 0) $) -- ( $ (15_crutch_concept1.west) + (0.25,0) $);
\draw[line width=0.495mm,-,/pgfplots/color of colormap=(505)] ($ (9_crutch_concept5.east) - (0.2, 0) $) -- ( $ (15_crutch_concept1.west) + (0.25,0) $);
\draw[line width=0.40555555555555556mm,-,/pgfplots/color of colormap=(595)] ($ (9_crutch_concept5.east) - (0.2, 0) $) -- ( $ (15_crutch_concept2.west) + (0.25,0) $);
\draw[line width=0.515mm,-,/pgfplots/color of colormap=(485)] ($ (9_crutch_concept6.east) - (0.2, 0) $) -- ( $ (15_crutch_concept1.west) + (0.25,0) $);
\draw[line width=0.4166666666666667mm,-,/pgfplots/color of colormap=(584)] ($ (9_crutch_concept6.east) - (0.2, 0) $) -- ( $ (15_crutch_concept2.west) + (0.25,0) $);
\draw[line width=0.4384615384615385mm,-,/pgfplots/color of colormap=(562)] ($ (3_crutch_concept8.east) - (0.2, 0) $) -- ( $ (9_crutch_concept3.west) + (0.25,0) $);
\draw[line width=0.3766666666666667mm,-,/pgfplots/color of colormap=(624)] ($ (3_crutch_concept8.east) - (0.2, 0) $) -- ( $ (9_crutch_concept5.west) + (0.25,0) $);
\draw[line width=0.32857142857142857mm,-,/pgfplots/color of colormap=(672)] ($ (3_crutch_concept8.east) - (0.2, 0) $) -- ( $ (9_crutch_concept1.west) + (0.25,0) $);
\draw[line width=0.428125mm,-,/pgfplots/color of colormap=(572)] ($ (3_crutch_concept8.east) - (0.2, 0) $) -- ( $ (9_crutch_concept6.west) + (0.25,0) $);
\draw[line width=0.4153846153846154mm,-,/pgfplots/color of colormap=(585)] ($ (3_crutch_concept6.east) - (0.2, 0) $) -- ( $ (9_crutch_concept3.west) + (0.25,0) $);
\draw[line width=0.3357142857142857mm,-,/pgfplots/color of colormap=(665)] ($ (3_crutch_concept6.east) - (0.2, 0) $) -- ( $ (9_crutch_concept1.west) + (0.25,0) $);
\draw[line width=0.403125mm,-,/pgfplots/color of colormap=(597)] ($ (3_crutch_concept6.east) - (0.2, 0) $) -- ( $ (9_crutch_concept6.west) + (0.25,0) $);
\draw[line width=1.0mm,-,/pgfplots/color of colormap=(0)] ($ (15_crutch_concept1.east) - (0.2, 0) $) -- ( $ (class crutch.west) + (0.25,0) $);
\draw[line width=0.3375mm,-,/pgfplots/color of colormap=(663)] ($ (15_crutch_concept2.east) - (0.2, 0) $) -- ( $ (class crutch.west) + (0.25,0) $);
\end{scope}
    \end{tikzpicture}
    \end{center}
    \vspace{-0.5cm}
    \caption{A VCC for four layers of a MViT~\cite{fan2021multiscale} model targeting the class ``crutch''. Darker lines denote larger concept contributions.}
    \label{fig:vcc_mvit}
\end{figure*}

%% file: VCCs/vitb_snowmobile.tex
\begin{figure*}
    \begin{center}
    \begin{tikzpicture}[grow=left,
        every node/.style = {inner sep=0pt},
every label/.append style = {label distance=2pt, align = center},
         sibling distance = 6em,
           level 1/.style = {level distance=9em,anchor=east},
           level 2/.style = {level distance=8em,anchor=east},
           level 3/.style = {level distance=8em,anchor=east},
           level 4/.style = {level distance=8em,anchor=east},
           ]  
\node[anchor=south,label=Snowmobile](class snowmobile)
{\includegraphics[width=34mm]{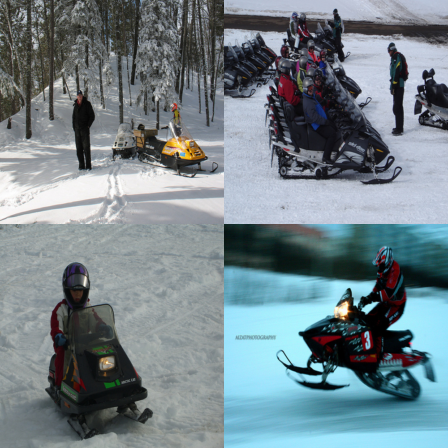}}
    child {node[label=10](10_snowmobile_concept3){\includegraphics[width=20mm]{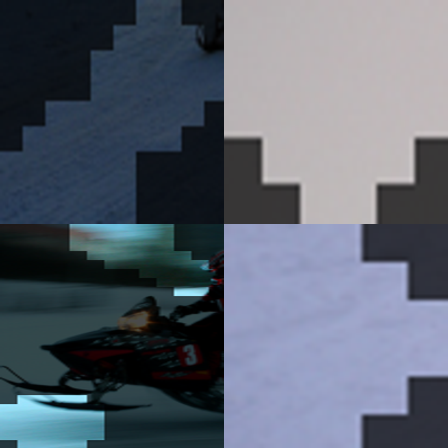}} edge from parent[draw=none]}
    child {node[](10_snowmobile_concept1){\includegraphics[width=20mm]{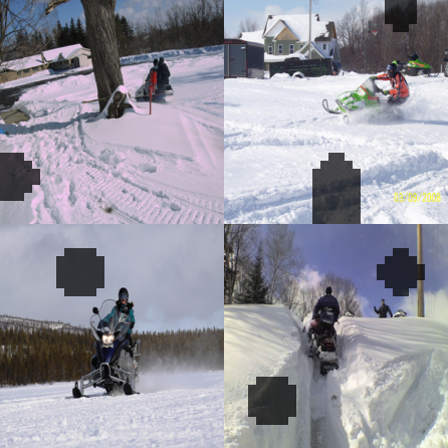}} edge from parent[draw=none]
        child {node[label=8](8_snowmobile_concept1){\includegraphics[width=20mm]{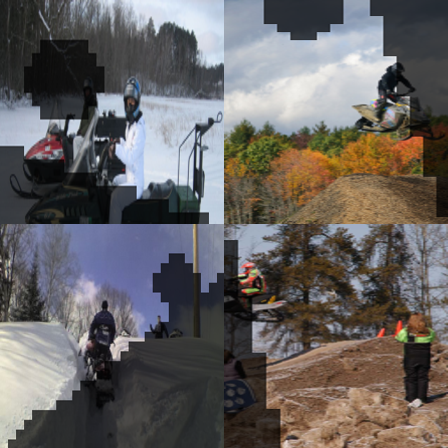}} 
        edge from parent[draw=none]}
        child {node[](8_snowmobile_concept3) {\includegraphics[width=20mm]{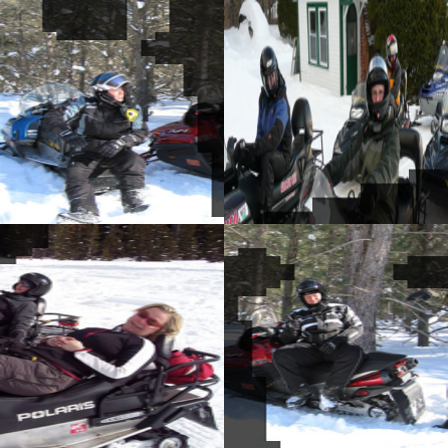}} 
        edge from parent[draw=none]
            child {node[label=5](5_snowmobile_concept3){\includegraphics[width=20mm]{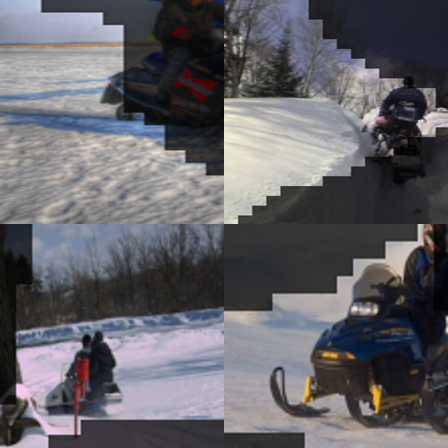}} 
            edge from parent[draw=none]}
            child {node[](5_snowmobile_concept4){\includegraphics[width=20mm]{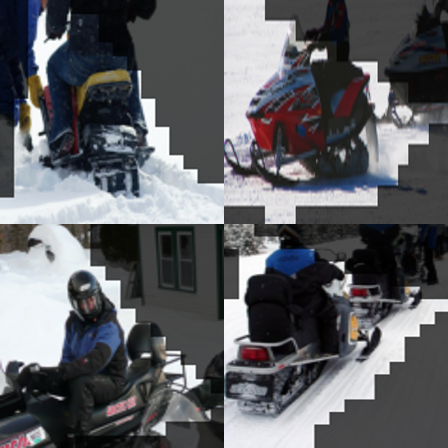}} 
            edge from parent[draw=none]
                child {node[label=2](2_snowmobile_concept8){\includegraphics[width=20mm]{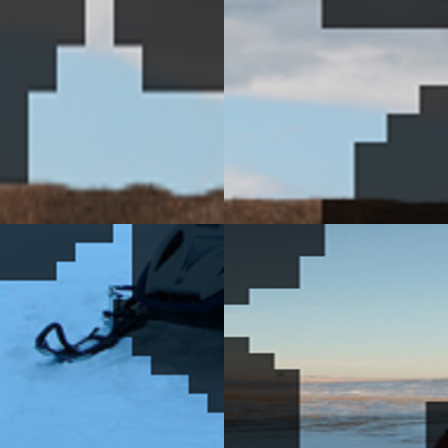}} 
                edge from parent[draw=none]}
                child {node[](2_snowmobile_concept7){\includegraphics[width=20mm]{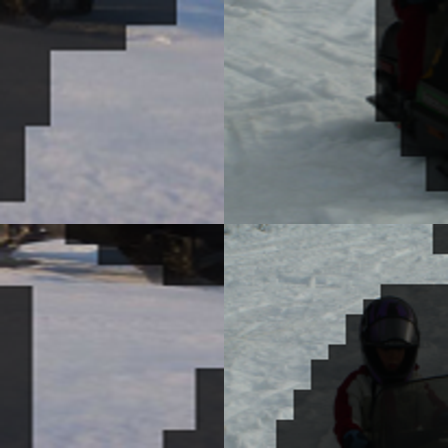}} 
                edge from parent[draw=none]}
                child {node[](2_snowmobile_concept12){\includegraphics[width=20mm]{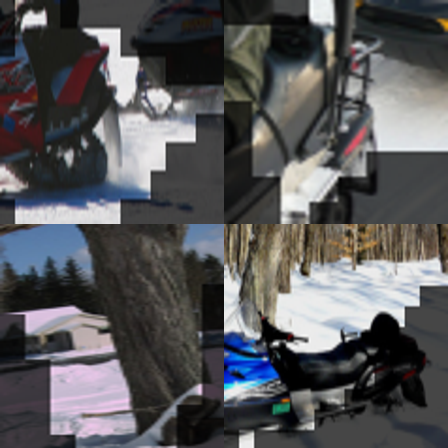}} 
                edge from parent[draw=none]}
                child {node[](2_snowmobile_concept3){\includegraphics[width=20mm]{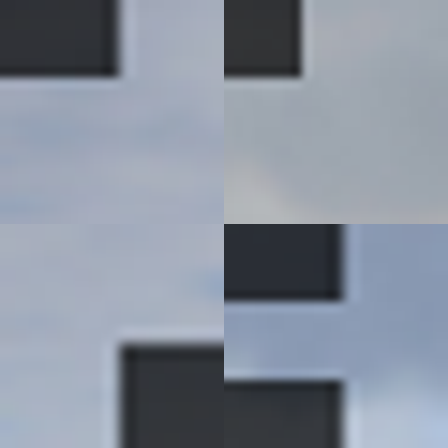}} 
                edge from parent[draw=none]}
                }
            child {node[label=5](5_snowmobile_concept6){\includegraphics[width=20mm]{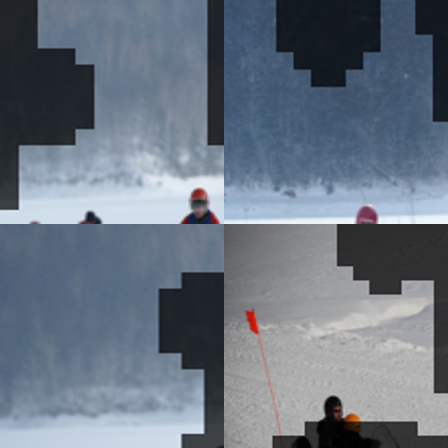}} 
            edge from parent[draw=none]}
            }
        child {node[](8_snowmobile_concept6){\includegraphics[width=20mm]{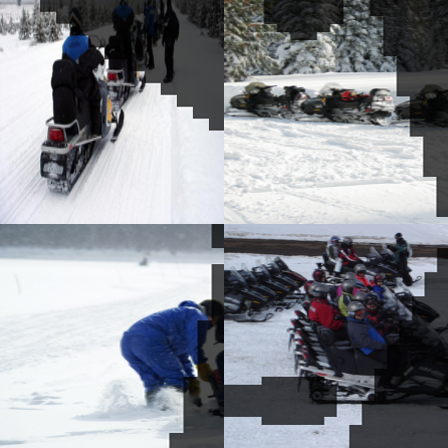}} 
        edge from parent[draw=none]} 
        }  
    child {node[](10_snowmobile_concept2){\includegraphics[width=20mm]{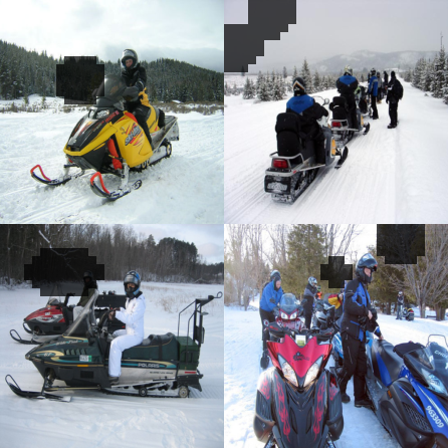}} edge from parent[draw=none]};
\begin{scope}[on background layer]
\draw[line width=0.2954545454545455mm,-,/pgfplots/color of colormap=(705)] ($ (2_snowmobile_concept3.east) - (0.2, 0) $) -- ( $ (5_snowmobile_concept6.west) + (0.25,0) $);
\draw[line width=0.4045454545454545mm,-,/pgfplots/color of colormap=(596)] ($ (2_snowmobile_concept3.east) - (0.2, 0) $) -- ( $ (5_snowmobile_concept4.west) + (0.25,0) $);
\draw[line width=0.5692307692307692mm,-,/pgfplots/color of colormap=(431)] ($ (5_snowmobile_concept6.east) - (0.2, 0) $) -- ( $ (8_snowmobile_concept1.west) + (0.25,0) $);
\draw[line width=0.5590909090909091mm,-,/pgfplots/color of colormap=(441)] ($ (5_snowmobile_concept6.east) - (0.2, 0) $) -- ( $ (8_snowmobile_concept6.west) + (0.25,0) $);
\draw[line width=0.5590909090909092mm,-,/pgfplots/color of colormap=(441)] ($ (5_snowmobile_concept4.east) - (0.2, 0) $) -- ( $ (8_snowmobile_concept6.west) + (0.25,0) $);
\draw[line width=0.5416666666666667mm,-,/pgfplots/color of colormap=(459)] ($ (5_snowmobile_concept4.east) - (0.2, 0) $) -- ( $ (8_snowmobile_concept3.west) + (0.25,0) $);
\draw[line width=0.36363636363636365mm,-,/pgfplots/color of colormap=(637)] ($ (2_snowmobile_concept12.east) - (0.2, 0) $) -- ( $ (5_snowmobile_concept6.west) + (0.25,0) $);
\draw[line width=0.4113636363636363mm,-,/pgfplots/color of colormap=(589)] ($ (2_snowmobile_concept12.east) - (0.2, 0) $) -- ( $ (5_snowmobile_concept4.west) + (0.25,0) $);
\draw[line width=0.3464285714285714mm,-,/pgfplots/color of colormap=(654)] ($ (2_snowmobile_concept12.east) - (0.2, 0) $) -- ( $ (5_snowmobile_concept3.west) + (0.25,0) $);
\draw[line width=0.6269230769230768mm,-,/pgfplots/color of colormap=(374)] ($ (5_snowmobile_concept3.east) - (0.2, 0) $) -- ( $ (8_snowmobile_concept1.west) + (0.25,0) $);
\draw[line width=0.3545454545454546mm,-,/pgfplots/color of colormap=(646)] ($ (2_snowmobile_concept7.east) - (0.2, 0) $) -- ( $ (5_snowmobile_concept6.west) + (0.25,0) $);
\draw[line width=0.4545454545454545mm,-,/pgfplots/color of colormap=(546)] ($ (2_snowmobile_concept7.east) - (0.2, 0) $) -- ( $ (5_snowmobile_concept4.west) + (0.25,0) $);
\draw[line width=0.3113636363636364mm,-,/pgfplots/color of colormap=(689)] ($ (2_snowmobile_concept8.east) - (0.2, 0) $) -- ( $ (5_snowmobile_concept6.west) + (0.25,0) $);
\draw[line width=0.43977272727272726mm,-,/pgfplots/color of colormap=(561)] ($ (2_snowmobile_concept8.east) - (0.2, 0) $) -- ( $ (5_snowmobile_concept4.west) + (0.25,0) $);
\draw[line width=0.4482142857142857mm,-,/pgfplots/color of colormap=(552)] ($ (2_snowmobile_concept8.east) - (0.2, 0) $) -- ( $ (5_snowmobile_concept3.west) + (0.25,0) $);
\draw[line width=0.3555555555555555mm,-,/pgfplots/color of colormap=(645)] ($ (8_snowmobile_concept1.east) - (0.2, 0) $) -- ( $ (10_snowmobile_concept2.west) + (0.25,0) $);
\draw[line width=0.3777777777777779mm,-,/pgfplots/color of colormap=(623)] ($ (8_snowmobile_concept1.east) - (0.2, 0) $) -- ( $ (10_snowmobile_concept1.west) + (0.25,0) $);
\draw[line width=0.40499999999999997mm,-,/pgfplots/color of colormap=(596)] ($ (8_snowmobile_concept1.east) - (0.2, 0) $) -- ( $ (10_snowmobile_concept3.west) + (0.25,0) $);
\draw[line width=0.4333333333333334mm,-,/pgfplots/color of colormap=(567)] ($ (8_snowmobile_concept6.east) - (0.2, 0) $) -- ( $ (10_snowmobile_concept1.west) + (0.25,0) $);
\draw[line width=0.68mm,-,/pgfplots/color of colormap=(320)] ($ (8_snowmobile_concept6.east) - (0.2, 0) $) -- ( $ (10_snowmobile_concept3.west) + (0.25,0) $);
\draw[line width=0.4055555555555556mm,-,/pgfplots/color of colormap=(595)] ($ (8_snowmobile_concept3.east) - (0.2, 0) $) -- ( $ (10_snowmobile_concept1.west) + (0.25,0) $);
\draw[line width=0.983mm,-,/pgfplots/color of colormap=(17)] ($ (10_snowmobile_concept2.east) - (0.2, 0) $) -- ( $ (class snowmobile.west) + (0.25,0) $);
\draw[line width=0.9100000000000001mm,-,/pgfplots/color of colormap=(90)] ($ (10_snowmobile_concept1.east) - (0.2, 0) $) -- ( $ (class snowmobile.west) + (0.25,0) $);
\draw[line width=0.49400000000000005mm,-,/pgfplots/color of colormap=(506)] ($ (10_snowmobile_concept3.east) - (0.2, 0) $) -- ( $ (class snowmobile.west) + (0.25,0) $);

\end{scope}
    \end{tikzpicture}
    \end{center}
    \vspace{-0.5cm}
    \caption{A VCC for four layers of a ViT~\cite{dosovitskiy2020image} model targeting the class ``snowmobile''. Darker lines denote larger concept contributions.}
    \label{fig:vcc_vitb}
\end{figure*}

%% file: VCCs/r18_indigobunting.tex
\begin{figure*}
    \begin{center}
    \begin{tikzpicture}[grow=left,
        every node/.style = {inner sep=0pt},
every label/.append style = {label distance=2pt, align = center},
         sibling distance = 6em,
           level 1/.style = {level distance=9em,anchor=east},
           level 2/.style = {level distance=8em,anchor=east},
           level 3/.style = {level distance=8em,anchor=east},
           level 4/.style = {level distance=8em,anchor=east},
           ]  
\node[anchor=south,label=Indigo Bunting](class_Indigo_Bunting)
{\includegraphics[width=34mm]{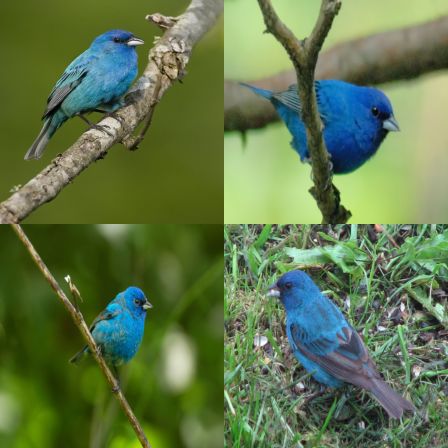}}
    child {node[label=stage4](stage4_Indigo_Bunting_concept3){\includegraphics[width=20mm]{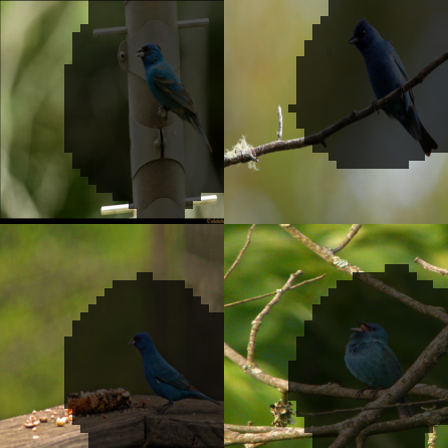}} edge from parent[draw=none]}
    child {node[](stage4_Indigo_Bunting_concept2){\includegraphics[width=20mm]{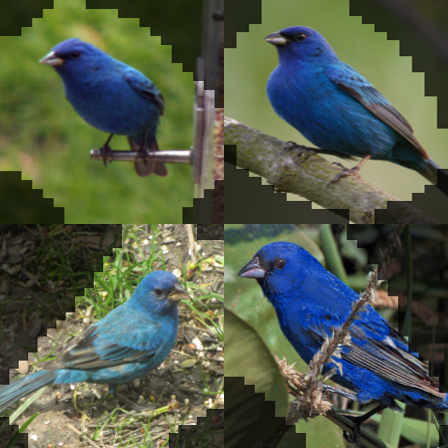}} edge from parent[draw=none]
        child {node[yshift=-0.95cm,label=stage3](stage3_Indigo_Bunting_concept1){\includegraphics[width=20mm]{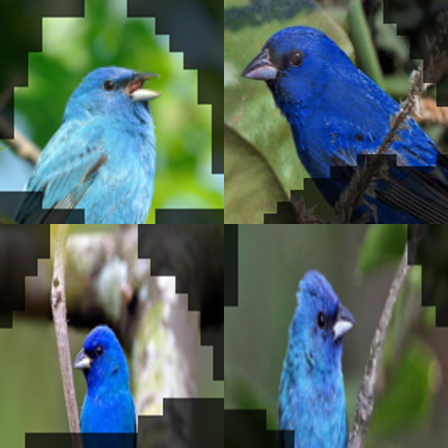}} 
        edge from parent[draw=none]}
        child {node[yshift=-0.95cm](stage3_Indigo_Bunting_concept4){\includegraphics[width=20mm]{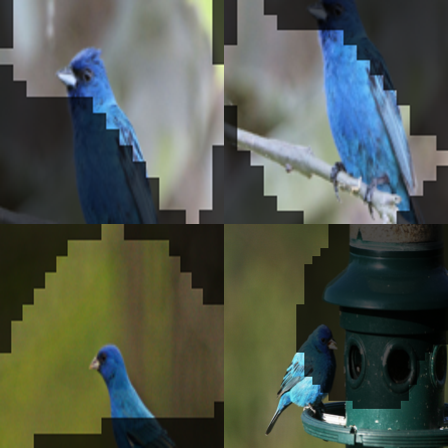}} 
        edge from parent[draw=none]}
        child {node[yshift=-0.95cm](stage3_Indigo_Bunting_concept5) {\includegraphics[width=20mm]{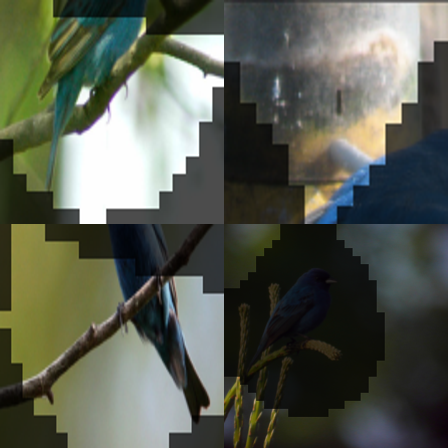}} 
        edge from parent[draw=none]
            child {node[label=stage2](stage2_Indigo_Bunting_concept2){\includegraphics[width=20mm]{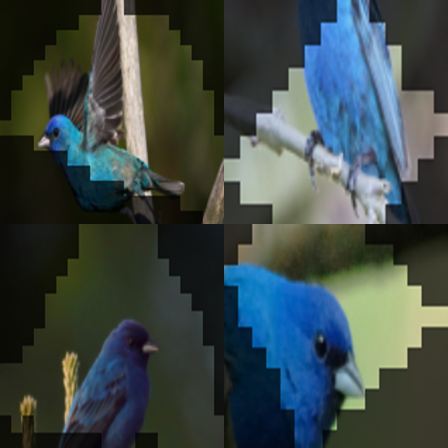}} 
            edge from parent[draw=none]}
            child {node[](stage2_Indigo_Bunting_concept4){\includegraphics[width=20mm]{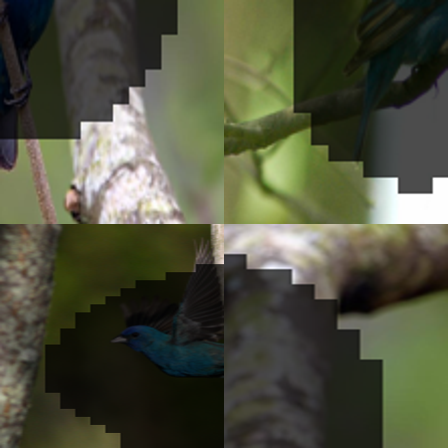}} 
            edge from parent[draw=none]
                child {node[yshift=-0.95cm,label=stage1](stage1_Indigo_Bunting_concept1){\includegraphics[width=20mm]{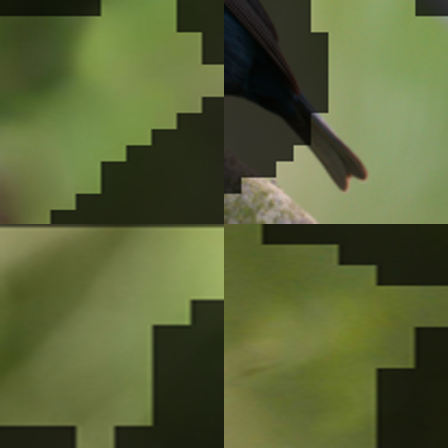}} 
                edge from parent[draw=none]}
                child {node[yshift=-0.95cm](stage1_Indigo_Bunting_concept7){\includegraphics[width=20mm]{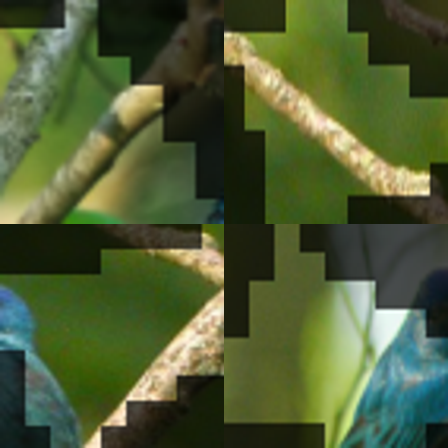}} 
                edge from parent[draw=none]}
                child {node[yshift=-0.95cm](stage1_Indigo_Bunting_concept4){\includegraphics[width=20mm]{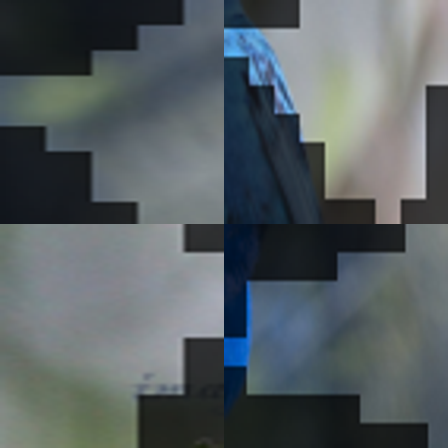}} 
                edge from parent[draw=none]}
                child {node[yshift=-0.95cm](stage1_Indigo_Bunting_concept3){\includegraphics[width=20mm]{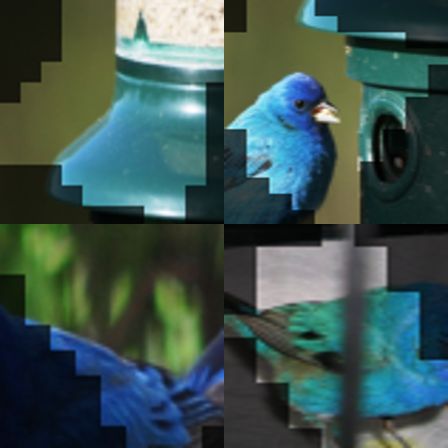}} 
                edge from parent[draw=none]}
                child {node[yshift=-0.95cm](stage1_Indigo_Bunting_concept2){\includegraphics[width=20mm]{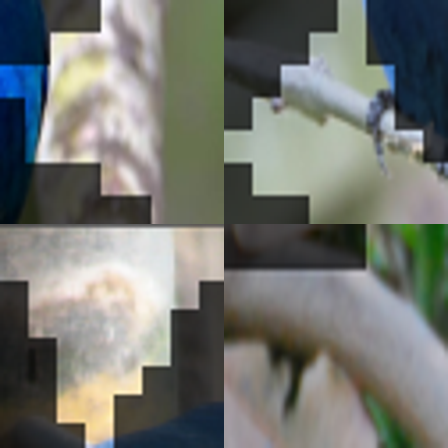}} 
                edge from parent[draw=none]}
                }
            child {node[](stage2_Indigo_Bunting_concept3){\includegraphics[width=20mm]{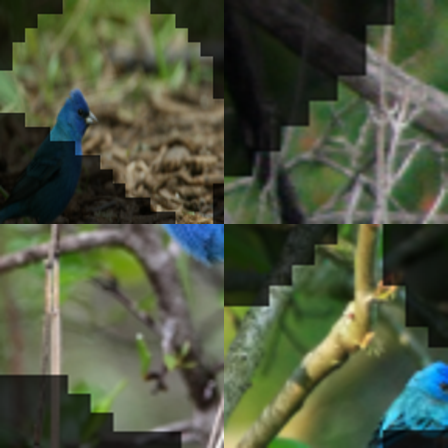}} 
            edge from parent[draw=none]}
            child {node[](stage2_Indigo_Bunting_concept1){\includegraphics[width=20mm]{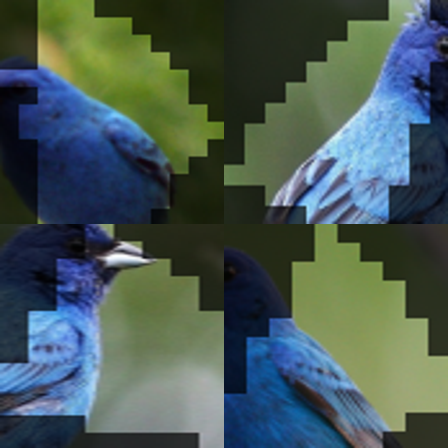}} 
            edge from parent[draw=none]}
            }
        child {node[yshift=-0.95cm](stage3_Indigo_Bunting_concept3){\includegraphics[width=20mm]{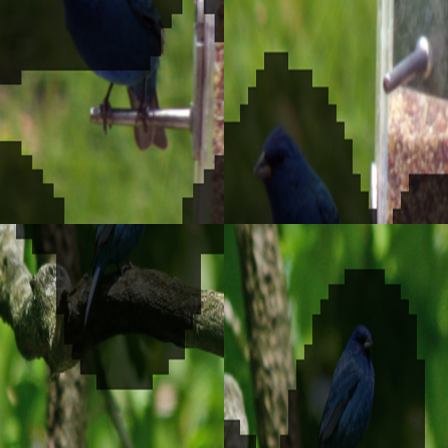}} 
        edge from parent[draw=none]}
        child {node[yshift=-0.95cm](stage3_Indigo_Bunting_concept6){\includegraphics[width=20mm]{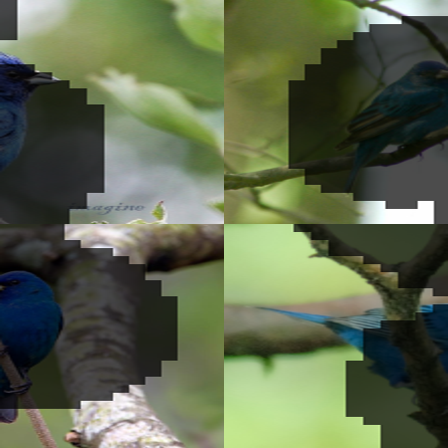}} 
        edge from parent[draw=none]} 
    }  
    child {node[](stage4_Indigo_Bunting_concept1){\includegraphics[width=20mm]{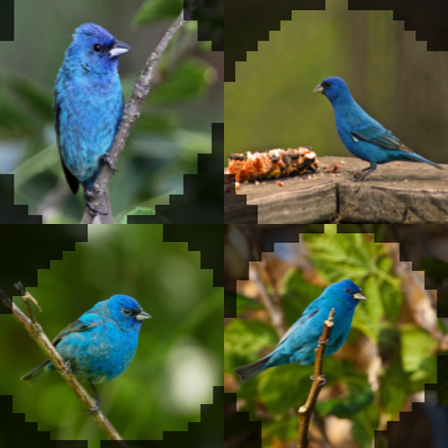}} 
    edge from parent[draw=none]} 
    child {node[](stage4_Indigo_Bunting_concept4){\includegraphics[width=20mm]{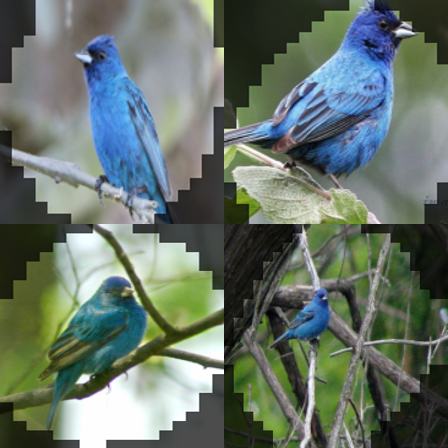}} edge from parent[draw=none]};
\begin{scope}[on background layer]
\draw[line width=0.30625mm,-,/pgfplots/color of colormap=(694)] ($ (stage1_Indigo_Bunting_concept4.east) - (0.2, 0) $) -- ( $ (stage2_Indigo_Bunting_concept4.west) + (0.25,0) $);
\draw[line width=0.46904761904761905mm,-,/pgfplots/color of colormap=(531)] ($ (stage1_Indigo_Bunting_concept4.east) - (0.2, 0) $) -- ( $ (stage2_Indigo_Bunting_concept1.west) + (0.25,0) $);
\draw[line width=0.33125mm,-,/pgfplots/color of colormap=(669)] ($ (stage2_Indigo_Bunting_concept4.east) - (0.2, 0) $) -- ( $ (stage3_Indigo_Bunting_concept1.west) + (0.25,0) $);
\draw[line width=0.6125mm,-,/pgfplots/color of colormap=(388)] ($ (stage2_Indigo_Bunting_concept1.east) - (0.2, 0) $) -- ( $ (stage3_Indigo_Bunting_concept3.west) + (0.25,0) $);
\draw[line width=0.5499999999999999mm,-,/pgfplots/color of colormap=(451)] ($ (stage1_Indigo_Bunting_concept2.east) - (0.2, 0) $) -- ( $ (stage2_Indigo_Bunting_concept3.west) + (0.25,0) $);
\draw[line width=0.346875mm,-,/pgfplots/color of colormap=(654)] ($ (stage1_Indigo_Bunting_concept2.east) - (0.2, 0) $) -- ( $ (stage2_Indigo_Bunting_concept4.west) + (0.25,0) $);
\draw[line width=0.5119047619047619mm,-,/pgfplots/color of colormap=(489)] ($ (stage1_Indigo_Bunting_concept2.east) - (0.2, 0) $) -- ( $ (stage2_Indigo_Bunting_concept1.west) + (0.25,0) $);
\draw[line width=0.2772727272727272mm,-,/pgfplots/color of colormap=(723)] ($ (stage2_Indigo_Bunting_concept3.east) - (0.2, 0) $) -- ( $ (stage3_Indigo_Bunting_concept5.west) + (0.25,0) $);
\draw[line width=0.3125mm,-,/pgfplots/color of colormap=(688)] ($ (stage2_Indigo_Bunting_concept3.east) - (0.2, 0) $) -- ( $ (stage3_Indigo_Bunting_concept1.west) + (0.25,0) $);
\draw[line width=0.340625mm,-,/pgfplots/color of colormap=(660)] ($ (stage2_Indigo_Bunting_concept3.east) - (0.2, 0) $) -- ( $ (stage3_Indigo_Bunting_concept3.west) + (0.25,0) $);
\draw[line width=0.3692307692307693mm,-,/pgfplots/color of colormap=(631)] ($ (stage2_Indigo_Bunting_concept3.east) - (0.2, 0) $) -- ( $ (stage3_Indigo_Bunting_concept4.west) + (0.25,0) $);
\draw[line width=0.384375mm,-,/pgfplots/color of colormap=(616)] ($ (stage1_Indigo_Bunting_concept1.east) - (0.2, 0) $) -- ( $ (stage2_Indigo_Bunting_concept4.west) + (0.25,0) $);
\draw[line width=0.5071428571428571mm,-,/pgfplots/color of colormap=(493)] ($ (stage1_Indigo_Bunting_concept1.east) - (0.2, 0) $) -- ( $ (stage2_Indigo_Bunting_concept1.west) + (0.25,0) $);
\draw[line width=0.4892857142857143mm,-,/pgfplots/color of colormap=(511)] ($ (stage1_Indigo_Bunting_concept7.east) - (0.2, 0) $) -- ( $ (stage2_Indigo_Bunting_concept3.west) + (0.25,0) $);
\draw[line width=0.384375mm,-,/pgfplots/color of colormap=(616)] ($ (stage1_Indigo_Bunting_concept7.east) - (0.2, 0) $) -- ( $ (stage2_Indigo_Bunting_concept4.west) + (0.25,0) $);
\draw[line width=0.5452380952380951mm,-,/pgfplots/color of colormap=(455)] ($ (stage1_Indigo_Bunting_concept7.east) - (0.2, 0) $) -- ( $ (stage2_Indigo_Bunting_concept1.west) + (0.25,0) $);
\draw[line width=0.5214285714285714mm,-,/pgfplots/color of colormap=(479)] ($ (stage1_Indigo_Bunting_concept3.east) - (0.2, 0) $) -- ( $ (stage2_Indigo_Bunting_concept3.west) + (0.25,0) $);
\draw[line width=0.378125mm,-,/pgfplots/color of colormap=(622)] ($ (stage1_Indigo_Bunting_concept3.east) - (0.2, 0) $) -- ( $ (stage2_Indigo_Bunting_concept4.west) + (0.25,0) $);
\draw[line width=0.5404761904761904mm,-,/pgfplots/color of colormap=(460)] ($ (stage1_Indigo_Bunting_concept3.east) - (0.2, 0) $) -- ( $ (stage2_Indigo_Bunting_concept1.west) + (0.25,0) $);
\draw[line width=0.755mm,-,/pgfplots/color of colormap=(245)] ($ (stage3_Indigo_Bunting_concept5.east) - (0.2, 0) $) -- ( $ (stage4_Indigo_Bunting_concept4.west) + (0.25,0) $);
\draw[line width=0.6666666666666665mm,-,/pgfplots/color of colormap=(334)] ($ (stage3_Indigo_Bunting_concept5.east) - (0.2, 0) $) -- ( $ (stage4_Indigo_Bunting_concept1.west) + (0.25,0) $);
\draw[line width=0.6733333333333333mm,-,/pgfplots/color of colormap=(327)] ($ (stage3_Indigo_Bunting_concept5.east) - (0.2, 0) $) -- ( $ (stage4_Indigo_Bunting_concept2.west) + (0.25,0) $);
\draw[line width=0.9099999999999999mm,-,/pgfplots/color of colormap=(91)] ($ (stage3_Indigo_Bunting_concept1.east) - (0.2, 0) $) -- ( $ (stage4_Indigo_Bunting_concept4.west) + (0.25,0) $);
\draw[line width=0.7333333333333332mm,-,/pgfplots/color of colormap=(267)] ($ (stage3_Indigo_Bunting_concept1.east) - (0.2, 0) $) -- ( $ (stage4_Indigo_Bunting_concept1.west) + (0.25,0) $);
\draw[line width=0.27083333333333326mm,-,/pgfplots/color of colormap=(730)] ($ (stage3_Indigo_Bunting_concept1.east) - (0.2, 0) $) -- ( $ (stage4_Indigo_Bunting_concept3.west) + (0.25,0) $);
\draw[line width=0.7299999999999998mm,-,/pgfplots/color of colormap=(271)] ($ (stage3_Indigo_Bunting_concept1.east) - (0.2, 0) $) -- ( $ (stage4_Indigo_Bunting_concept2.west) + (0.25,0) $);
\draw[line width=0.6949999999999997mm,-,/pgfplots/color of colormap=(306)] ($ (stage3_Indigo_Bunting_concept3.east) - (0.2, 0) $) -- ( $ (stage4_Indigo_Bunting_concept4.west) + (0.25,0) $);
\draw[line width=0.5619047619047619mm,-,/pgfplots/color of colormap=(439)] ($ (stage3_Indigo_Bunting_concept3.east) - (0.2, 0) $) -- ( $ (stage4_Indigo_Bunting_concept1.west) + (0.25,0) $);
\draw[line width=0.3083333333333333mm,-,/pgfplots/color of colormap=(692)] ($ (stage3_Indigo_Bunting_concept3.east) - (0.2, 0) $) -- ( $ (stage4_Indigo_Bunting_concept3.west) + (0.25,0) $);
\draw[line width=0.75mm,-,/pgfplots/color of colormap=(250)] ($ (stage3_Indigo_Bunting_concept3.east) - (0.2, 0) $) -- ( $ (stage4_Indigo_Bunting_concept2.west) + (0.25,0) $);
\draw[line width=0.8450000000000001mm,-,/pgfplots/color of colormap=(155)] ($ (stage3_Indigo_Bunting_concept4.east) - (0.2, 0) $) -- ( $ (stage4_Indigo_Bunting_concept4.west) + (0.25,0) $);
\draw[line width=0.7476190476190476mm,-,/pgfplots/color of colormap=(253)] ($ (stage3_Indigo_Bunting_concept4.east) - (0.2, 0) $) -- ( $ (stage4_Indigo_Bunting_concept1.west) + (0.25,0) $);
\draw[line width=0.21666666666666665mm,-,/pgfplots/color of colormap=(784)] ($ (stage3_Indigo_Bunting_concept4.east) - (0.2, 0) $) -- ( $ (stage4_Indigo_Bunting_concept3.west) + (0.25,0) $);
\draw[line width=0.7899999999999999mm,-,/pgfplots/color of colormap=(211)] ($ (stage3_Indigo_Bunting_concept4.east) - (0.2, 0) $) -- ( $ (stage4_Indigo_Bunting_concept2.west) + (0.25,0) $);
\draw[line width=0.2681818181818182mm,-,/pgfplots/color of colormap=(732)] ($ (stage2_Indigo_Bunting_concept2.east) - (0.2, 0) $) -- ( $ (stage3_Indigo_Bunting_concept5.west) + (0.25,0) $);
\draw[line width=0.31875mm,-,/pgfplots/color of colormap=(682)] ($ (stage2_Indigo_Bunting_concept2.east) - (0.2, 0) $) -- ( $ (stage3_Indigo_Bunting_concept1.west) + (0.25,0) $);
\draw[line width=1.0mm,-,/pgfplots/color of colormap=(0)] ($ (stage4_Indigo_Bunting_concept4.east) - (0.2, 0) $) -- ( $ (class_Indigo_Bunting.west) + (0.25,0) $);
\draw[line width=1.0mm,-,/pgfplots/color of colormap=(0)] ($ (stage4_Indigo_Bunting_concept1.east) - (0.2, 0) $) -- ( $ (class_Indigo_Bunting.west) + (0.25,0) $);
\draw[line width=0.85mm,-,/pgfplots/color of colormap=(150)] ($ (stage4_Indigo_Bunting_concept2.east) - (0.2, 0) $) -- ( $ (class_Indigo_Bunting.west) + (0.25,0) $);
\draw[line width=0.95mm,-,/pgfplots/color of colormap=(50)] ($ (stage4_Indigo_Bunting_concept3.east) - (0.2, 0) $) -- ( $ (class_Indigo_Bunting.west) + (0.25,0) $);
\draw[line width=0.7049999999999998mm,-,/pgfplots/color of colormap=(296)] ($ (stage3_Indigo_Bunting_concept6.east) - (0.2, 0) $) -- ( $ (stage4_Indigo_Bunting_concept4.west) + (0.25,0) $);
\draw[line width=0.5571428571428572mm,-,/pgfplots/color of colormap=(443)] ($ (stage3_Indigo_Bunting_concept6.east) - (0.2, 0) $) -- ( $ (stage4_Indigo_Bunting_concept1.west) + (0.25,0) $);
\draw[line width=0.2583333333333333mm,-,/pgfplots/color of colormap=(742)] ($ (stage3_Indigo_Bunting_concept6.east) - (0.2, 0) $) -- ( $ (stage4_Indigo_Bunting_concept3.west) + (0.25,0) $);
\draw[line width=0.6533333333333333mm,-,/pgfplots/color of colormap=(347)] ($ (stage3_Indigo_Bunting_concept6.east) - (0.2, 0) $) -- ( $ (stage4_Indigo_Bunting_concept2.west) + (0.25,0) $);
\end{scope}
    \end{tikzpicture}
    \end{center}
    \vspace{-0.5cm}
    \caption{A VCC for four layers of a ResNet18~\cite{he2016deep} model trained on the finegrained CUB~\cite{WahCUB_200_2011} dataset, targeting the class ``indigo bunting''. Darker lines denote larger concept contributions.}
    \label{fig:vcc_r50_cub}
\end{figure*}